\documentclass[10pt,journal,compsoc]{IEEEtran}

\ifCLASSOPTIONcompsoc
  \usepackage[nocompress]{cite}
\else
  \usepackage{cite}
\fi

\ifCLASSINFOpdf
\else
\fi

\usepackage{url}            
\usepackage{booktabs}       
\usepackage{amsfonts}       
\usepackage{nicefrac}       
\usepackage{microtype}      
\usepackage{xcolor}         
\renewcommand{\textcolor}[2]{#2}

\usepackage{graphicx}
\usepackage{amsmath,amssymb}
\usepackage{verbatim}
\usepackage{multirow}
\usepackage{colortbl}
\usepackage{marvosym}
\usepackage{makecell}
\usepackage{subcaption}
\usepackage{bm}
\usepackage{wrapfig}
\usepackage{caption}
\usepackage{algorithm}
\usepackage{algorithmic}
\usepackage[marginal]{footmisc}

\usepackage{makecell}
\usepackage{array}
\usepackage{multirow}
\usepackage{ragged2e}
\usepackage{booktabs}

\usepackage{bbding}

\usepackage{rotating}
\usepackage{tabularray}

\usepackage{hyperref}

\begin{document}
\title{Towards Collaborative Autonomous Driving: Simulation Platform and End-to-End System}

\author{Genjia Liu, Yue Hu, Chenxin Xu, Weibo Mao, Junhao Ge, Zhengxiang Huang, \\ Yifan Lu, Yinda Xu, Junkai Xia, Yafei Wang, Siheng Chen
\IEEEcompsocitemizethanks{
\IEEEcompsocthanksitem Genjia Liu, Yue Hu, Chenxin Xu, Weibo Mao, Junhao Ge, Zhengxiang Huang, Yifan Lu, Yinda Xu, Junkai Xia, Siheng Chen are with the Cooperative Medianet Innovation Center (CMIC) at
Shanghai Jiao Tong University, Shanghai, China. 
E-mail: LGJ1zed, 18671129361, xcxwakaka, kirino.mao, cancaries, huangzhengxiang, yifan\_lu, yinda\_xu, xiajunkai, sihengc@sjtu.edu.cn.
\IEEEcompsocthanksitem Yafei Wang is with the School of Mechanical Engineering, Shanghai Jiao Tong University, E-mail: wyfjlu@sjtu.edu.cn.
\IEEEcompsocthanksitem Corresponding author is Siheng Chen.
}
}


\IEEEtitleabstractindextext{%
\begin{abstract}

Vehicle-to-everything-aided autonomous driving (V2X-AD) has a huge potential to provide a safer driving solution. Despite extensive research in transportation and communication to support V2X-AD,  the actual utilization of these infrastructures and communication resources in enhancing driving performances remains largely unexplored. This highlights the necessity of collaborative autonomous driving; that is, a machine learning approach that optimizes the information sharing strategy to improve the driving performance of each vehicle. This effort necessitates two key foundations: a platform capable of generating data to facilitate the training and testing of V2X-AD, and a comprehensive system that integrates full driving-related functionalities with mechanisms for information sharing. 
From the platform perspective, we present \textbf{V2Xverse}, a comprehensive simulation platform for collaborative autonomous driving.
This platform provides a complete pipeline for collaborative driving: multi-agent driving dataset generation scheme, codebase for deploying full-stack collaborative driving systems, closed-loop driving performance evaluation with scenario customization. 
From the system perspective, we introduce \textbf{CoDriving}, a novel end-to-end collaborative driving system that properly integrates V2X communication over the entire autonomous pipeline, promoting driving with shared perceptual information.
The core idea is a novel driving-oriented communication strategy, that is, selectively complementing the driving-critical regions in single-view using sparse yet informative perceptual cues. Leveraging this strategy, CoDriving improves driving performance while optimizing communication efficiency. We make comprehensive benchmarks with V2Xverse, analyzing both modular performance and closed-loop driving performance.
Experimental results show that CoDriving: i) significantly improves the driving score by 62.49\% and drastically reduces the pedestrian collision rate by 53.50\% compared to the SOTA end-to-end driving method, and ii) achieves sustaining driving performance superiority over dynamic constraint communication conditions.
Our code is available at \href{https://github.com/CollaborativePerception/V2Xverse}{https://github.com/CollaborativePerception/V2Xverse}.


\end{abstract}

}

\maketitle

\IEEEdisplaynontitleabstractindextext

\IEEEpeerreviewmaketitle










\vspace{-3mm}
\IEEEraisesectionheading{\section{Introduction}\label{sec:introduction}}

\IEEEPARstart{V}ehicle-to-everything-communication-aided autonomous driving (V2X-AD) targets to improve driving performance by enabling vehicles and their surroundings, such as roadside units, and pedestrians equipped with smart devices, to exchange complementary information. V2X-AD can significantly address the inherent limitations of single-agent autonomous driving~\cite{TransFuser,TransFuser+,InterFuser,LAV,TCP,NEAT}, like restricted visibility, the unpredictability of other road users, and the development of secure paths. Through the exchange of data, V2X-AD equips individual autonomous vehicles with an enriched perception of their surroundings~\cite{liu2020when2com,LiuWho2com:ICRA20,WangV2vnet:ECCV20,XuOPV2V:ICRA22,HuWhere2comm:NeurIPS22 ,HuCollaboration:CVPR23}, enabling them to see beyond obstructions and promptly identify smaller, fast-moving entities. \textcolor{blue}{This enhanced environmental awareness enables more accurate forecasting and effective route planning, leading to faster responses in emergencies. This capability significantly contributes to accident prevention, promoting safer and more reliable autonomous driving~\cite{Accident1,WangDeepAccident:ArXiv23}}.


To enable V2X-AD, previous works have made efforts from various aspects.
From the transportation perspective, intelligent infrastructures equipped with roadside units~\cite{ITS1,ITS2,ITS3,ITS4,ITS5,zimmer2024tumtraf} have been strategically deployed along roads. 
\textcolor{blue}{These units provide complementary viewpoints to aid vehicles in perceiving the driving environment.}
From the communication perspective, customized V2X communication standards and protocols~\cite{V2X1,V2X2,V2X3,V2X4,comm1,comm2,comm4,comm5} have emerged to facilitate reliable and real-time data exchange among vehicles and infrastructures.
\textcolor{blue}{Supported by intelligent roadside units and advanced communication protocols, the development of V2X-AD is grounded on a robust foundation.}
Recently, the nascent field of collaborative perception has emerged, aiming to address challenges in multi-agent systems from a perceptual standpoint. This approach focuses on enhancing the perceptual capabilities of each agent by facilitating the exchange of complementary perceptual information among them. Specifically, in the context of V2X-AD, the agents are conceptualized as vehicles and roadside units integrated within infrastructure systems. These entities collaborate by sharing perceptual data, enabling each vehicle to accurately identify all surrounding foreground objects. 
To achieve this, a bunch of collaborative perception methods have been proposed to address a serious of critical challenges, including the trade-off between perception performance and communication costs~\cite{HuCollaboration:CVPR23,LiuWho2com:ICRA20,liu2020when2com,WangV2vnet:ECCV20,LiLearning:NeurIPS21,Li_2021_RAL,HuWhere2comm:NeurIPS22,LuRobust:ICRA23,LiMultiRobot:CoRL22,GaoRegularized:RCS20,XuOPV2V:ICRA22,xu2022v2xvit,YuDAIRV2X:CVPR22,XuCoBEVT:CoRL22,hu2024pragmatic}, robustness to pose error~\cite{LuRobust:ICRA23}, communication latency~\cite{LeiLatency:ECCV22,wei-cobevflow-2023}, and the heterogeneous issue~\cite{lu-heal-2023}.

%



Despite the encouraging progress on collaborative perception, a fundamental inherent limitation remains rarely explored, that is, these works focus solely on optimizing module-level perception capability, \textcolor{blue}{it is still unknown how exactly system-level driving capability can be enhanced.}
To fill this gap, we aim to expand collaborative perception to cover holistic driving capabilities, beginning with perception and extending through essential modules such as planning and control.
We define this effort as \textbf{collaborative autonomous driving}. 
Distinct from the concept of V2X-AD, a comprehensive engineering solution that includes infrastructure setup, communication systems, and system optimization among other aspects, collaborative autonomous driving focuses on a machine learning strategy to improve each agent's system-level driving performance through information sharing among multiple agents.
To this end, we focus on two essential components in developing collaborative autonomous driving: the closed-loop driving platform and the end-to-end driving system.
First, a comprehensive platform that \textcolor{blue}{encompasses the full collaborative autonomous driving process, providing training data and supporting driving system evaluation.} While a real-world platform is a direct approach, its high cost and safety concerns necessitate the development of a simulation platform as a practical alternative.
Second, we need an end-to-end system that integrates information sharing with related driving functions. Such a system allows for a thorough examination of how information sharing benefits the driving capabilities.

From the platform perspective, this paper presents \textbf{V2Xverse}, a comprehensive simulation platform for collaborative autonomous driving.
The key feature of our proposed V2Xverse is to enable both offline benchmark generation for driving-related subtasks and online closed-loop evaluation for driving performances in diverse scenarios, comprehensively supporting the development of collaborative autonomous driving systems.
To create V2X-AD scenarios, V2Xverse outfits multiple vehicles with full driving capabilities and strategically places intelligent infrastructures. This setup offers complementary viewpoints for vehicle perception and facilitates communication between vehicles and infrastructures.
To support the subsequent developments of collaborative autonomous driving methods, V2Xverse provides a full set of driving signals and annotations for system training, and it also provides diverse safety-critical scenarios for closed-loop driving evaluation, covering a convincing test length in virtual towns.
V2Xverse offers three distinct advantages compared to existing platforms.
First, V2Xverse promotes multi-agent simulation while the previous driving platforms~\cite{carlaleaderboard,Nocrash,TransFuser,shao2023reasonnet} only support single-agent driving simulation.
Second, V2Xverse promotes full driving functions simulation while previous collaborative perception platforms~\cite{XuOPV2V:ICRA22,LiV2XSim:RAL22} only support functions related to the perception module.
Third, V2Xverse supports comprehensive V2X-AD scenarios, including diverse sensor equipment, model integration, and flexible scenario customization; see a summary in Table~\ref{table:platform comparison}. Through these features, V2Xverse enables the development of diverse collaborative autonomous driving systems.

From the system perspective, we develop \textbf{CoDriving}, a novel end-to-end collaborative autonomous driving system that leverages perceptual information sharing to improve driving performance.
Contrasting with previous V2X communication strategies in collaborative perception~\cite{LiuWho2com:ICRA20,liu2020when2com,HuWhere2comm:NeurIPS22}, which focus on optimizing perception abilities, our system introduces a novel~\emph{driving-oriented communication strategy}. 
The core idea is to optimize communication content based on the feedback from the driving plan. This mechanism allows each intelligent vehicle to identify driving-critical areas and selectively request collaborative messages to enhance perceptual information in these areas.
To achieve this, we develop CoDriving with two functionalities: i) end-to-end autonomous driving, offering full driving capabilities; ii) driving-oriented collaboration, which leverages a novel driving request map that assigns higher scores to the spatial regions near the planned driving waypoints. This request map enables the selection of sparse perceptual features in the driving-critical regions, which are then employed to improve individual driving capabilities.
CoDriving offers two distinct advantages.
First, CoDriving leverages information sharing to upgrade the whole driving system, improving perception, planning, and driving performance, while previous collaborative perception approaches~\cite{HuWhere2comm:NeurIPS22,HuCollaboration:CVPR23} solely optimize the perception performance.
Second, CoDriving promotes adaptability to different communication conditions and demonstrates generalizability to different modalities, while the previous collaborative driving method~\cite{CuiCoopernaut:CVPR22} can only work at a predefined ample communication bandwidth and cater to LiDAR input.


%

To validate the effectiveness of our proposed V2Xverse platform and CoDriving system, we undertake three key evaluations. First, we conduct system-level evaluations in terms of perception, planning, and closed-loop driving performance, which benchmarks earlier collaborative perception methods by integrating them into the whole driving system, showcasing the adaptability and extensibility of our V2Xverse platform.
\textcolor{blue}{Second, we perform an additional evaluation of perception performance, which compares our CoDriving against previous collaborative perception methods using established collaborative perception benchmarks, including DAIR-V2X~\cite{YuDAIRV2X:CVPR22}, V2V4Real~\cite{XuV2V4Real:CVPR23}, OPV2V~\cite{XuOPV2V:ICRA22}, V2XSIM2.0~\cite{LiV2XSim:RAL22}, and TUMTraf-V2X~\cite{zimmer2024tumtraf}.} 
Third, we conduct robustness assessments, involving four types of practical issues: communication bandwidth limitation, communication latency, pose error, and effectiveness under heterogeneous setting. 
These evaluations provide a comprehensive understanding of the strengths and advancements offered by our V2Xverse platform and CoDriving system.
Comprehensive experimental results show that CoDriving improves the driving score by 62.49\% and drastically reduces the pedestrian collision rate by 53.50\% compared to the SOTA single-agent end-to-end driving method.

To sum up, our contributions are:

$\bullet$ We propose V2Xverse, a comprehensive V2X-aided autonomous driving simulation platform. This platform enables the development of collaborative driving systems by supporting offline benchmark generation for driving-related subtasks and online closed-loop driving performance evaluation in diverse scenarios.

$\bullet$ We propose CoDriving, a novel end-to-end 
collaborative autonomous driving system, which improves driving performance by sharing driving-critical information.

$\bullet$ We conduct comprehensive experiments and validate that: i) V2X communication-enabled information sharing significantly outperforms single-agent end-to-end autonomous driving systems, and ii) CoDriving achieves superior performance-bandwidth trade-off in system-level evaluations.

The rest of the paper is organized as follows: in Section~\ref{sec:related work}, we review existing works related to V2X-AD. In Section~\ref{sec:platform}, we introduce our simulation platform V2Xverse, from platform construction to benchmark generation.
In Section~\ref{sec:system}, we introduce our collaborative driving system CoDriving, including the end-to-end driving pipeline and driving-oriented collaboration strategy.
\textcolor{blue}{In Section~\ref{sec:Benchmark_perception}, we present an offline evaluation on existing widely-used collaborative perception benchmarks, validating the effectiveness of CoDriving compared to previous collaborative perception methods in terms of perception capability.
In Section~\ref{sec:Benchmark_driving}, we present system-level evaluations in V2Xverse, validating the effectiveness of CoDriving
in terms of perception, waypoints planning and closed-loop driving capacities.}
Finally, we draw the conclusion of this paper in Section~\ref{sec:conclusion}.

\begin{table*}[t]
\centering
\caption{\textcolor{blue}{Comparison of V2Xverse with existing CARLA-based closed-loop driving platforms/benchmarks.}}
\vspace{-2mm}
\begin{tabular}{c|l|cccc|ccc}
\toprule
\multicolumn{2}{c|}{\multirow{2}{*}{\begin{tabular}[c]{@{}c@{}} \\ \textit{\textbf{CARLA-based Benchmarks}}\end{tabular}}} & \multicolumn{4}{c}{\textbf{Single Agent}} & \multicolumn{3}{c}{\textbf{Multiple Agents}} \\ \cmidrule{3-9} 
\multicolumn{2}{c|}{} & \begin{tabular}[c]{@{}c@{}}\textbf{Leadboard}\\ \textbf{v1}\cite{carlaleaderboard}\end{tabular} & \begin{tabular}[c]{@{}c@{}}\textbf{NoCrash}\\ \cite{Nocrash}\end{tabular} & \textbf{Longest6}\cite{TransFuser+} & \textbf{DOS}\cite{shao2023reasonnet} & \begin{tabular}[c]{@{}c@{}}\textbf{OpenCDA}\\ \cite{XuOpenCDA:ITSC2021}\end{tabular} & \begin{tabular}[c]{@{}c@{}}\textbf{AutoCastSim}\\ \cite{CuiCoopernaut:CVPR22}\end{tabular} & \begin{tabular}[c]{@{}c@{}}\textbf{V2Xverse}\\ \textbf{(Ours)}\end{tabular} \\ \midrule
\multirow{5}{*}{\textbf{Traffic}} & \cellcolor[HTML]{EFEFEF}\textbf{Test Routes} & \cellcolor[HTML]{EFEFEF}100 & \cellcolor[HTML]{EFEFEF}25 & \cellcolor[HTML]{EFEFEF}36 & \cellcolor[HTML]{EFEFEF}100 & \cellcolor[HTML]{EFEFEF}1 & \cellcolor[HTML]{EFEFEF}3 & \cellcolor[HTML]{EFEFEF}67 \\
 & \textbf{Average Length} & 1.7km & N/A & 1.5km & N/A & 2.8km & N/A & 412m \\
 & \cellcolor[HTML]{EFEFEF}\textbf{Scenario Types} & \cellcolor[HTML]{EFEFEF} 21 & \cellcolor[HTML]{EFEFEF}3 & \cellcolor[HTML]{EFEFEF}7 & \cellcolor[HTML]{EFEFEF}4 & \cellcolor[HTML]{EFEFEF}2 & \cellcolor[HTML]{EFEFEF}3 & \cellcolor[HTML]{EFEFEF}24 \\
 & \textbf{Vehicles} & \checkmark & \checkmark & \checkmark & \checkmark & \checkmark & \checkmark & \checkmark \\
 & \cellcolor[HTML]{EFEFEF}\textbf{Pedestrians} & \cellcolor[HTML]{EFEFEF}\checkmark & \cellcolor[HTML]{EFEFEF}\checkmark & \cellcolor[HTML]{EFEFEF}\checkmark & \cellcolor[HTML]{EFEFEF}\checkmark & \cellcolor[HTML]{EFEFEF}-- & \cellcolor[HTML]{EFEFEF}-- & \cellcolor[HTML]{EFEFEF}\checkmark \\ \midrule
\multirow{4}{*}{\textbf{\begin{tabular}[c]{@{}c@{}}Intelligent\\ \\ Agents\end{tabular}}} & \textbf{Number} & 1 & 1 & 1 & 1 &  $\geq$2 &  $\geq$2 & $\geq$2  \\
 & \cellcolor[HTML]{EFEFEF}\textbf{Roadside Units} & \cellcolor[HTML]{EFEFEF}-- & \cellcolor[HTML]{EFEFEF}-- & \cellcolor[HTML]{EFEFEF}-- & \cellcolor[HTML]{EFEFEF}-- & \cellcolor[HTML]{EFEFEF}-- & \cellcolor[HTML]{EFEFEF}-- & \cellcolor[HTML]{EFEFEF}\checkmark \\
 & \textbf{Lidar} & \checkmark & -- & \checkmark & \checkmark & \checkmark & \checkmark & \checkmark \\
 & \cellcolor[HTML]{EFEFEF}\textbf{Camera} & \cellcolor[HTML]{EFEFEF}\checkmark & \cellcolor[HTML]{EFEFEF}\checkmark & \cellcolor[HTML]{EFEFEF}\checkmark & \cellcolor[HTML]{EFEFEF}\checkmark & \cellcolor[HTML]{EFEFEF}\checkmark & \cellcolor[HTML]{EFEFEF}-- & \cellcolor[HTML]{EFEFEF}\checkmark \\ \midrule
\multirow{2}{*}{\textbf{Communication}} & \textbf{V2V} & -- & -- & -- & -- & \checkmark & \checkmark & \checkmark \\
 & \cellcolor[HTML]{EFEFEF}\textbf{V2I} & \cellcolor[HTML]{EFEFEF}-- & \cellcolor[HTML]{EFEFEF}-- & \cellcolor[HTML]{EFEFEF}-- & \cellcolor[HTML]{EFEFEF}-- & \cellcolor[HTML]{EFEFEF}\checkmark & \cellcolor[HTML]{EFEFEF}-- & \cellcolor[HTML]{EFEFEF}\checkmark \\ \midrule
\multirow{2}{*}{\textbf{\begin{tabular}[c]{@{}c@{}}\textcolor{blue}{Robustness}\\ \textcolor{blue}{Evaluation}\end{tabular}}} & \textbf{Latency} & -- & -- & -- & -- & -- & -- & \checkmark \\
 & \cellcolor[HTML]{EFEFEF}\textbf{Pose Error} & \cellcolor[HTML]{EFEFEF}-- & \cellcolor[HTML]{EFEFEF}-- & \cellcolor[HTML]{EFEFEF}-- & \cellcolor[HTML]{EFEFEF}-- & \cellcolor[HTML]{EFEFEF}-- & \cellcolor[HTML]{EFEFEF}-- & \cellcolor[HTML]{EFEFEF}\checkmark \\ \midrule
\multirow{3}{*}{\textbf{\begin{tabular}[c]{@{}c@{}}Online Driving\\ \\ Performance\end{tabular}}} & \textbf{Completion} & \checkmark & \checkmark & \checkmark & \checkmark & \checkmark & \checkmark & \checkmark \\
 & \cellcolor[HTML]{EFEFEF}\textbf{Safety} & \cellcolor[HTML]{EFEFEF}\checkmark & \cellcolor[HTML]{EFEFEF}\checkmark & \cellcolor[HTML]{EFEFEF}\checkmark & \cellcolor[HTML]{EFEFEF}\checkmark & \cellcolor[HTML]{EFEFEF}\checkmark & \cellcolor[HTML]{EFEFEF}\checkmark & \cellcolor[HTML]{EFEFEF}\checkmark \\
 & \textbf{Efficiency} & -- & -- & -- & -- & \checkmark & \checkmark & \checkmark \\ \midrule
\multirow{2}{*}{\textbf{\begin{tabular}[c]{@{}c@{}}Modular\\ Performance\end{tabular}}} & \cellcolor[HTML]{EFEFEF}\textbf{Perception} & \cellcolor[HTML]{EFEFEF}-- & \cellcolor[HTML]{EFEFEF}-- & \cellcolor[HTML]{EFEFEF}-- & \cellcolor[HTML]{EFEFEF}-- & \cellcolor[HTML]{EFEFEF}\checkmark & \cellcolor[HTML]{EFEFEF}-- & \cellcolor[HTML]{EFEFEF}\checkmark \\
 & \textbf{Planning} & -- & -- & -- & -- & -- & -- & \checkmark \\
 \bottomrule
 
\end{tabular}

\vspace{-3mm}
\label{table:platform comparison}
\end{table*}
\vspace{-3mm}
\section{Related Work}
\label{sec:related work}

\noindent
\textbf{End-to-end autonomous driving.} 
Recently, learning-based end-to-end autonomous driving has emerged as an active topic, which directly maps environment information into control signals thus conceptually avoiding the cascading error of complex modular design. Recent works mainly fall into two categories: reinforcement learning (RL) and imitation learning (IL). Reinforcement learning for end-to-end autonomous driving involves training an autonomous vehicle to navigate and control itself by interacting with the environment, it offers the potential for systems that can adapt to diverse and complex driving scenarios.
\cite{latent_RL, GRAID} first map the vision environment into the latent embedding space with auxiliary semantic supervision, and then train the RL agent using the latent representation. WOR \cite{chen2021learning} builds a RL agent based on a forward model that assumes the world on rails, and then conducts policy distillation to obtain the final agent. Roach \cite{Roach} utilizes privileged environmental information to train a RL expert agent that maps bird’s-eye view (BEV) images to continuous actions. Imitation learning targets to clone the behavior of an expert agent by fitting the recorded driving data. Mainstream methods mainly improve driving performance from four aspects, ~\cite{NEAT,TransFuser,UniAD,InterFuser,shao2023reasonnet} leverage transformer architecture to learn better representations, ~\cite{MP3,UniAD,LAV,TCP,InterFuser} involve auxiliary tasks to assist in driving task learning,  ~\cite{TransFuser,TransFuser+,InterFuser} integrate representations from multi-modalities to exploit their complementary advantages for autonomous driving, and \cite{LBC,TCP} adopt policy distillation strategy.
CIL \cite{CIL} and CILRS \cite{CILRS} stand as early baselines for IL. CIL \cite{CIL} introduces a conditional architecture that uses different branches for different driving commands, and CILRS \cite{CILRS} extends the CIL model with deeper network and an additional speed prediction head. LBC \cite{LBC} applies a two-stage imitation learning. A cheating model with access to privileged information first clones the behavior of the expert, which is then imitated by a final model using sensor data. \cite{MP3} builds scene representations with a predicted online map and occupancy field, which is then followed by a motion planning module.
TCP \cite{TCP} conducts predictions for both future trajectory and control signal, where the trajectory branch provides guidance for the control prediction, and an intermediate distillation is conducted for better behavior clone. LAV \cite{LAV} augments the dataset by learning driving strategy from all vehicles around in addition to the ego vehicle. DriveAdapter \cite{DriveAdapter} explores to directly utilize the powerful RL teacher model for planning by adopting it to the predicted BEV segmentation. 
\textcolor{blue}{UniAD \cite{UniAD} introduces a planning-oriented autonomous driving framework, unifying
tracking, mapping, motion prediction, occupancy prediction, and planning in a single system,
where the key component is a query-based design to connect all modules.}

However, single-agent driving systems \cite{TCP,UniAD,TransFuser+,LAV,Roach,InterFuser,chen2021learning} inevitably encounter limitations in distant or occluded areas which may lead to catastrophic failures. Several approaches have emerged to tackle this issue. For instance, ReasonNet \cite{shao2023reasonnet} predicts the future evolution of scenes and infers potential occupancy by modeling the interaction between objects and the environment, thereby enhancing perception performance under occlusion. However, it fails to predict the motion of objects that have never appeared in the field of view, as well as invisible objects that do not interact with the environment. Coopernaut~\cite{CuiCoopernaut:CVPR22} introduces a visually-cooperative driving system, where the vehicle aggregates voxel representations from collaborators to improve driving decisions. Coopernaut takes a step towards end-to-end collaborative autonomous driving, but it can only handle a fixed communication bandwidth, and its strongly coupled framework limits the extension in other AD functionalities.
To overcome these limitations, we design an end-to-end full-featured collaborative autonomous driving system that ensures adaptability to any communication bandwidth.

\noindent
\textbf{V2X communication.}
V2X communication serves as the fundamental basis of V2X-AD, as it enables information sharing among agents. There are two kinds of mainstream V2X communication technologies: dedicated short-range communication (DSRC) and cellular-based communication. 
DSRC~\cite{comm1} achieves low end-to-end latency in a short communication range. Cellular communication~\cite{comm2,comm4,comm5} supports high mobility communication. 
Through the achieved progress in V2X communication, the communication bandwidth in a realistic communication system is always constrained. To better make use of the precise communication resources, we design a strategy to transmit compact driving features.

\noindent
\textbf{Collaborative perception.}
Collaborative perception~\cite{LiuWho2com:ICRA20,liu2020when2com,WangV2vnet:ECCV20,XuOPV2V:ICRA22,LiLearning:NeurIPS21,Li_2021_RAL,xu2022v2xvit,YuDAIRV2X:CVPR22,XuCoBEVT:CoRL22,HuWhere2comm:NeurIPS22,LuRobust:ICRA23,LiMultiRobot:CoRL22,GaoRegularized:RCS20,HuCollaboration:CVPR23,hu2024pragmatic} is an emerging application of V2X-communication-aided systems, which promotes the crucial perception module of autonomous driving through complementary perceptual information sharing.
Several high-quality platforms have emerged~\cite{Li_2021_RAL,XuOPV2V:ICRA22,YuDAIRV2X:CVPR22,HuWhere2comm:NeurIPS22}, which simulate collaborative perception scenarios and provide diverse perception annotations to aid in the development of collaborative perception systems. 
Collaborative perception systems have made remarkable progress, with CoCa3D~\cite{HuCollaboration:CVPR23} achieving nearly complete perception. 
However, collaborative perception mainly focused on improving perception performance, failed to achieve the ultimate driving goal of V2X-AD due to the lack of other necessary driving functions. In this work, we fill this gap by introducing a novel end-to-end collaborative driving system, achieving complete V2X-AD functionalities.


\noindent
\textbf{CARLA-based autonomous driving benchmark}.
CARLA-based autonomous driving benchmarks \cite{carlaleaderboard,Nocrash,TransFuser+,shao2023reasonnet,XuOpenCDA:ITSC2021,CuiCoopernaut:CVPR22} involve the creation of comprehensive test scenarios based on the NHTSA pre-crash typology and performance metrics within the CARLA simulator, providing a standardized environment for assessing the capabilities of autonomous systems. NoCrash \cite{Nocrash} serves as an early benchmark, it collects training data from CARLA Town01 under 4 specific weather conditions and performs generalization testing with other towns and weather conditions. Meanwhile, the CARLA simulator provides official evaluation leaderboards \cite{carlaleaderboard} (v1, v2). Leaderboard v1 involves 100 secret test routes with an average length of 1.7 km. Leaderboard v2 is released in 2023, it is a more challenging version equipped with updated maps and scenarios. Longest6 Benchmark \cite{TransFuser+} reduces and balances the test routes by choosing the 6 longest routes from each of the 6 available towns. DOS Benchmark \cite{shao2023reasonnet} is dedicated to validating driving performance in collision scenarios, involving 100 cases of 4 specifically designed scenarios. In addition to the aforementioned extensively discussed single-agent benchmarks, \cite{XuOpenCDA:ITSC2021,CuiCoopernaut:CVPR22} explore the feasibility of multi-agent driving evaluation. OpenCDA \cite{XuOpenCDA:ITSC2021} serves as a prototype cooperative driving automation platform that offers platooning scenario for benchmark testing. AutoCastSim \cite{CuiCoopernaut:CVPR22} also supports multi-agent autonomous driving simulation and offers three traffic scenarios with challenging vision occlusion. However, existing multi-agent benchmarks are limited to a small number of scenes and routes, and lack discussion on system-level performance of end-to-end driving pipelines, as well as the impact of practical issues on driving performance. In this work, V2Xverse fills this gap, enabling both offline/online evaluation of collaborative driving systems in various driving scenarios.
\begin{figure*}[!t]
\centering
\includegraphics[width=0.96\linewidth]{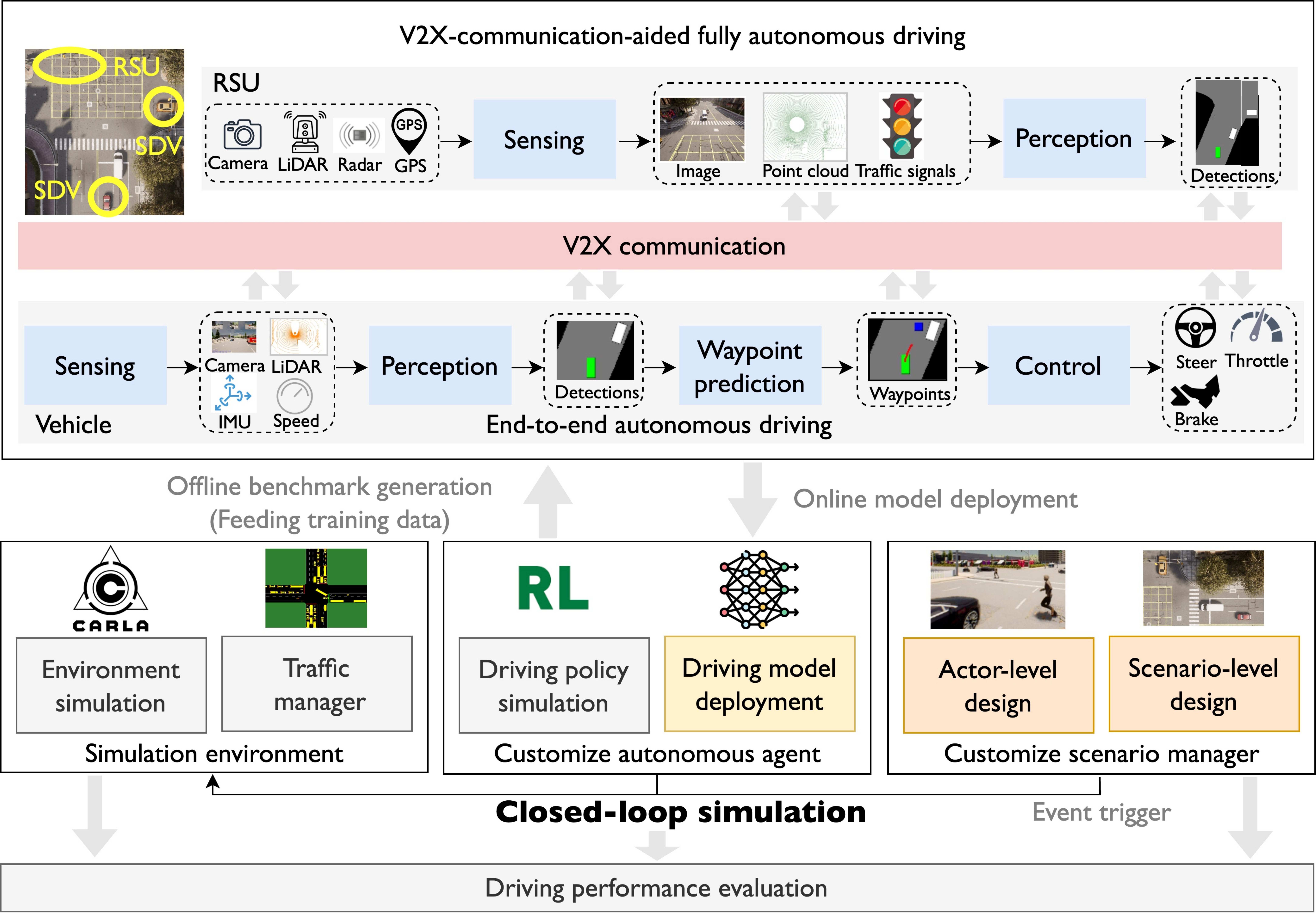}
\caption{\small Platform overview. V2Xverse simulates the complete V2X-AD driving pipeline, incorporating various driving functionalities and delivering extensive driving annotations. It facilitates both the offline benchmark generation and online closed-loop driving performance evaluation.}
\label{fig:platform}
\vspace{-4mm}
\end{figure*}

\section{V2Xverse: V2X-Aided Fully Autonomous Driving Simulation Platform}
\label{sec:platform}

This section presents V2Xverse, a comprehensive simulation platform for V2X-communication-aided autonomous driving, which supports both data generation as well as closed-loop evaluation for collaborative driving systems. Notably, closed-loop driving requires vehicles to drive continuously with real-time feedback from the environment, and closed-loop evaluation is critical in the assessment of driving systems since i) it more accurately simulates the dynamic conditions of real-world driving, thus addressing the limitations of open-loop evaluations that only document static scenarios and may not effectively highlight the strengths of various driving systems; and ii) it avoids the high expenses and safety risks associated with real-world tests, enabling quick prototyping and testing. This approach promotes the quick iteration of ideas and offers affordable access to a wide range of driving scenarios.

We introduce the platform from four aspects: platform construction, offline benchmark generation, online closed-loop evaluation, and comparisons with previous platforms.

\vspace{-3mm}
\subsection{Platform construction}
\label{sec:Platform construction}

\textcolor{blue}{V2Xverse is built based on CARLA simulator~\cite{DosovitskiyCARLA:CoRL2017}, it extends capabilities by supporting the deployment of closed-loop V2X-aided autonomous driving in challenging scenarios.
To evaluate autonomous driving proficiency, we consider the navigation task in realistic safety-critical traffic scenarios and deploy the complete pipeline for collaborative driving.}

\noindent
\textbf{Navigation task in safety-critical scenarios.} 
The navigation task requires autonomous vehicles to complete the predefined routes while dealing with high densities of dynamic interactive agents. The driving scenes include static road layouts, static objects and dynamic traffic agents. To assess driving safety more comprehensively, we customize varying safety-critical and challenging scenarios. From the traffic perspective, we turn off traffic lights to create more dynamic and interactive driving scenarios. From the agent perspective, we set up some crazy pedestrians that ignore traffic rules and surrounding vehicles. We also include more large vehicles and roadside obstacles for diverse simulations. Additionally, we design several trigger-based scenarios extended from CARLA Autonomous Driving Leaderboard setting~\cite{carlaleaderboard}, e.g. pedestrians suddenly appearing from occluded regions to cross the road, unexpected agents from occluded buildings, and unprotected turns at intersections, see Figure~\ref{Fig:scenario}. 
\textcolor{blue}{See Appendix~\ref{sec_A1} for more details in scenario generation.}

\begin{figure*}[!t]
  \centering
    \begin{subfigure}{0.30\linewidth}
    \includegraphics[width=1.0\linewidth]{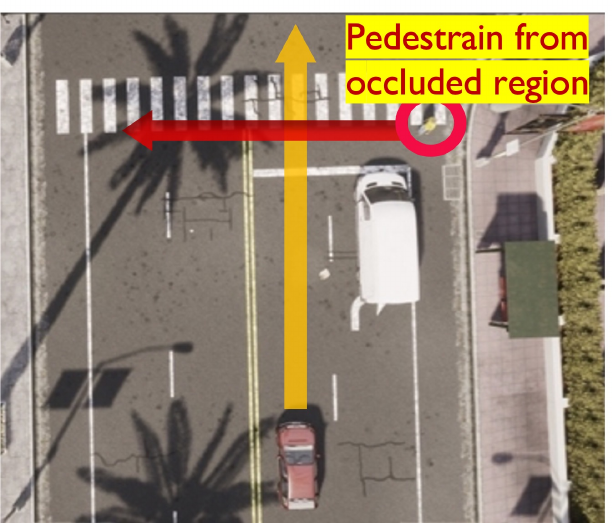}
    \label{fig:scenario_ped}
  \end{subfigure}
  \begin{subfigure}{0.304\linewidth}
    \includegraphics[width=1.0\linewidth]{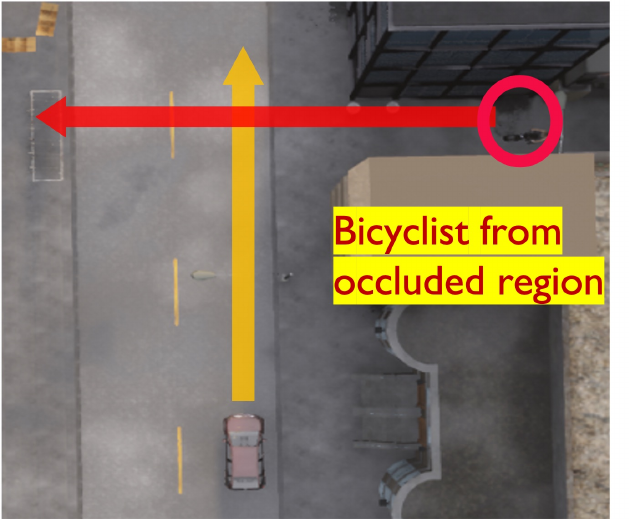}
    \label{fig:scenario_bicy}
  \end{subfigure}
  \begin{subfigure}{0.302\linewidth}
    \includegraphics[width=1.0\linewidth]{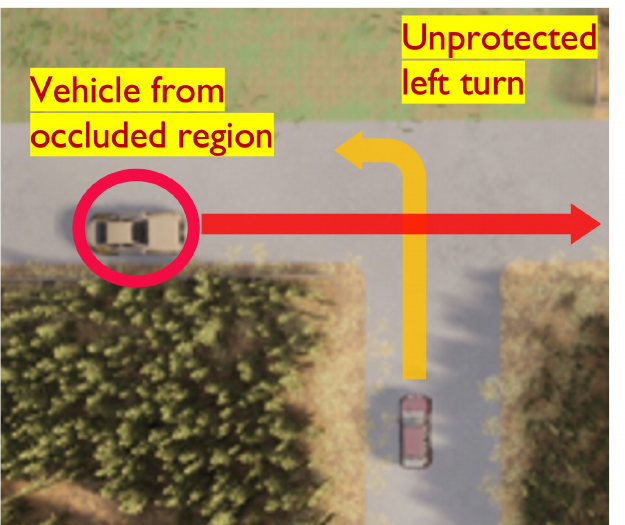}
    \label{fig:scenario_veh}
  \end{subfigure}
  \vspace{-2mm}
  \caption{\small Safety-critical scenarios caused by occlusion, including crazy pedestrians from occluded vehicles, unexpected bicyclists from occluded buildings, and vehicles rapidly coming from occluded road infrastructures.}
  \label{Fig:scenario}
\end{figure*}

\noindent
\textbf{Deployment of V2X collaboration.} 
To simulate V2X situations, we deploy multiple collaborative vehicles and roadside units (RSU) within effective collaboration distance, and build real-time communication between them. The vehicles start from adjacent positions and follow the same route. Meanwhile, RSUs are strategically positioned on the roadside alongside vehicles to ensure a comprehensive view of the traffic conditions, and we ensure that each vehicle has at least one roadside unit for valid collaboration at each moment.
Vehicles are equipped with complete driving pipeline, including localization, perception, planning, and control, while RSUs are solely equipped with a perception module.
Both vehicles and RSUs are equipped with a communication module and multi-modality sensors including four RGB cameras and a LiDAR.
All the driving-related signals, including pose, sensor data, waypoints, and speed, can be exchanged among agents through communication. We demonstrate the deployment of V2X collaboration in 
Appendix~\ref{sec:appendix data descrip}.


\vspace{-3mm}
\subsection{Offline benchmark generation}

V2Xverse platform enables offline benchmark generation by providing all the necessary driving annotations, which support the full-stack training of collaborative driving systems.
We apply an automatic labeling algorithm to generate data, where multiple autonomous vehicles concurrently drive following a widely-adopted expert policy~\cite{TransFuser,TransFuser+}, and their driving behaviors serve as ground-truth labels for imitation learning~\cite{LBC}.
\textcolor{blue}{The benchmark we provide is fully annotated, 
including 288k synchronous images from four views, 72k LiDAR point clouds, 482k 2D/3D bounding boxes of vehicles, pedestrians, way-points, speeds, steering angles and brake signals, see Figure \ref{fig:platform}. Statistically, we split the dataset samples: 26322 for training, 5502 for validating, 4306 for testing, spanning 8 towns and 108 routes with a frame rate of 5 FPS.} See Appendix~\ref{sec:appendix data descrip} for more details in data description.

\vspace{-3mm}
\subsection{Online closed-loop evaluation}
\label{sec:evaluation}
Unlike offline evaluations~\cite{Caesar2020nuScenesAM,UniAD,YuDAIRV2X:CVPR22,XuOPV2V:ICRA22,XuV2V4Real:CVPR23,LiV2XSim:RAL22}, which lack environmental interaction and adaptive evolution, online driving evaluation offers real-time interactive insights into system performance, allowing for the identification of safety risks and algorithm improvement in realistic environment~\cite{carlaleaderboard,Nocrash}. V2Xverse enables online closed-loop evaluation by supporting system deployment and driving performance evaluation.
We deploy the collaborative autonomous driving system on vehicles and RSUs in CARLA. The collaborative system transforms sensor observations and communicated messages into driving actions, and we assess the driving performance in the navigation task.
We set 67 evaluation routes that cover a wide range of driving situations, including three levels of speed and five levels of urgency. 
As we expect both safe and efficient driving, a bunch of metrics, including the driving score (DS), route completion ratio (RC), infraction score (IS), pedestrian/vehicle/layout collision rate, following~\cite{carlaleaderboard}, and mean speed are used for comprehensive driving performance evaluation.

\vspace{-3mm}
\subsection{Comparison with previous simulation platforms}
A vast array of significant works~\cite{carlaleaderboard,Nocrash,TransFuser+,shao2023reasonnet,XuOpenCDA:ITSC2021,CuiCoopernaut:CVPR22} have been proposed to build autonomous driving platforms and benchmarks based on CARLA. Compared with these platforms, our V2Xverse has the following advantages: 
i) V2Xverse supports communication aided multi-intelligent-agent simulation. In comparison to mainstream single-intelligent-agent driving benchmarks \cite{carlaleaderboard,Nocrash,TransFuser+,shao2023reasonnet}, V2Xverse extends functionality by supporting Vehicle-to-Vehicle (V2V) and Vehicle-to-Infrastructure (V2I) communication, and \textcolor{blue}{enables the evaluation of system robustness against practical challenges, such as communication latency and pose errors.
ii) V2Xverse provides various yet customizable safety-critical traffic scenarios (24 types), sufficient test routes (67), and realistic urban driving environment with pedestrians and vehicles, while existing multi-intelligent-agent simulation platforms \cite{XuOpenCDA:ITSC2021,CuiCoopernaut:CVPR22} caters to vehicle-only and limited scenarios and routes.
iii) V2Xverse enables the deployment and comprehensive evaluation for collaborative autonomous driving systems. V2Xverse provides various driving signals and annotations in V2X-AD situations, which support the V2X-based closed-loop driving task and modular tasks including both perception and planning; while~\cite{CuiCoopernaut:CVPR22} only caters to driving performance; see the detailed comparison with previous platforms in Table~\ref{table:platform comparison}. }

\begin{figure*}[!t]
\centering
\includegraphics[width=0.97\linewidth]{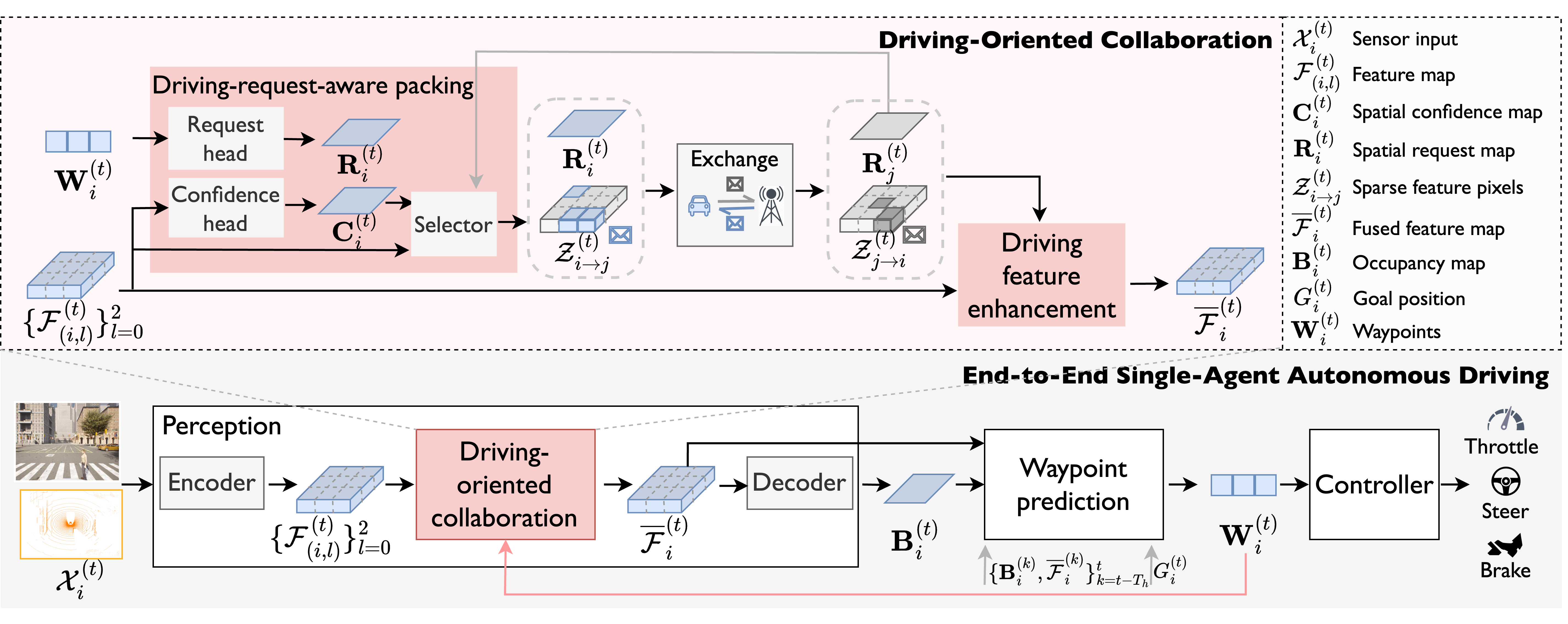}
\vspace{-1.5mm}
\caption{\small System overview. CoDriving comprises two
components: end-to-end single-agent autonomous driving, which transforms the sensor inputs into driving actions, and driving-oriented collaboration, which enhances the single-agent features by aggregating the driving-critical perceptual features shared through communication. The benefits propagate from the perception module to the entire driving pipeline, enhancing all driving signals.}
\label{fig:system}
\vspace{-2mm}
\end{figure*}

\vspace{-2mm}
\section{CoDriving: End-to-End Collaborative Autonomous Driving System}\label{sec:system}

This section introduces our proposed CoDriving, a novel end-to-end collaborative autonomous driving system, which promotes driving performance through information sharing.

\vspace{-3mm}
\subsection{Problem formulation}
\label{section_4_1}

Consider $N$ collaborative agents in the V2X scenario, where agents refer to vehicles and roadside units. We focus on the navigation task, where each vehicle is assigned a specific destination. Given the observations of the $N$ agents as $\{ \mathcal{X}_i \}_{i=1}^N$, the overall objective of collaborative autonomous driving is to maximize the driving performance of each vehicle by exchanging information among all vehicles and roadside units within a communication budget $B$; that is,
\begin{align}
\max_{\theta,\mathcal{P}}\sum_{i=1}^N  d\left(  \Phi_\theta\left(\mathcal{X}_i, \mathcal{D}_i,  \{\mathcal{P}_{j\to i}\}_{j=1}^N\right) \right),
\text{s.t.}\sum_{j\neq i}|\mathcal{P}_{j\to i}|\leq  B
\label{sec4:problem_formulation}
\end{align}
where $d(\cdot)$ is the driving performance metric, $\Phi_{\theta}(\cdot)$ is the end-to-end autonomous driving network with trainable parameter $\theta$, and $\mathcal{X}_i, \mathcal{D}_i$ are the observation and destination of the $i$th agent respectively, and $\mathcal{P}_{j\to i}$ is the message transmitted from the $j$th agent to the $i$th agent. Note that the driving performance of roadside units is evaluated as zero.

To optimize the trade-off between driving performance and communication cost, we present CoDriving, which comprises two components; see Figure~\ref{fig:system}. First, an end-to-end imitation learning-based autonomous driving is introduced in Section~\ref{subsec:end2end}, which offers full driving capabilities, including perception, waypoints planning, and driving control. Second, a novel driving-oriented collaboration is introduced in Section~\ref{subsec:collaboration}, which first leverages a driving-request-aware communication strategy to select sparse driving-critical perceptual information for sharing, and then leverages driving feature enhancement to boost the individual driving capabilities with the received collaborative messages.




\vspace{-2mm}
\subsection{End-to-end single-agent autonomous driving}
\label{subsec:end2end}
The end-to-end single-agent autonomous driving network learns to output driving actions based on inputs from different modalities. To achieve this, we integrate the driving-needed modular components in a unified system, including a perception module, a waypoint predictor, and a controller.
We use the bird's-eye-view (BEV) representation, as it provides a unified global coordinate system, avoiding complex coordinate transformations and better supporting cross-agent collaboration.




\textbf{Perception.} The perception module adapts to inputs from two modalities: RGB images and 3D point clouds, and detects surrounding objects with category and regressed bounding box. For the $i$th agent, given its input $\mathcal{X}_i^{(t)}$ at timestamp $t$, we leverage an encoder to extract BEV feature following~\cite{Lang2018PointPillarsFE, lu-heal-2023} 
; that is, $\Phi_{\rm enc}(\cdot)=\Phi_{\rm tcv}(\Phi_{\rm cv}(\cdot))$, where $\Phi_{\rm cv}$/$\Phi_{\rm tcv}$ represents convolution/transposed-convolution layers.
Specifically, the intermediate features from the 3rd/7th/12th block of $\Phi_{\rm cv}$ are denoted as 
$\{\mathcal{F}_{(i,l)}^{(t)} \in \mathbb{R}^{\frac{X}{2^l} \times \frac{Y}{2^l} \times 2^l D}\}_{l=0}^2 = \Phi_{\rm cv}(\mathcal{X}_i^{(t)})$.
Then, the final output of BEV encoder is obtained as $ \mathcal{F}_{i}^{(t)} = \Phi_{\rm scv}( \mathrm{concat}( \Phi_{\rm tcv}(\{\mathcal{F}_{(i,l)}^{(t)}\}_{l=0}^2))) \in \mathbb{R}^{X \times Y \times D}$, where $X,Y,D$ are its height, weight and channel, $\mathrm{concat}(\cdot)$ is feature concatenation, and $\Phi_{\rm scv}(\cdot)$ is a convolution block to shrink feature dimension. For image input, we additionally extract its 2D convolutional feature and transform it to BEV with a warping function~\cite{philion2020lift}, after which the encoder $\Phi_{\rm enc}(\cdot)$  extracts features in BEV.
For the 3D point cloud, we initialize the input by discretizing 3D points as a BEV map~\cite{Lang2018PointPillarsFE}.
In the collaboration phase in Section \ref{subsec:collaboration}, agents will share the critical regions within BEV features $\{\mathcal{F}_{(i,l)}^{(t)}\}_{l=0}^2$.

The detection decoder $\Phi_{\rm dec}(\cdot)$ comprises a classification head $\Phi_{\rm cls}(\cdot)$ and a regression head $\Phi_{\rm reg}(\cdot)$. Given the BEV feature $\mathcal{F}_i^{(t)}$, $\Phi_{\rm cls}(\cdot)$ generates the object probability heatmap in the view of the $i$th agent by $
     \mathbf{S}_i^{(t)}  =  \Phi_{\rm cls}(\mathcal{F}_i^{(t)}) \in \mathbb{R}^{X \times Y \times K}
$, where $K$ denotes the classes of objects. $\Phi_{\rm reg}(\cdot)$ generates a dense bounding box regression map by $
     \mathcal{O}_i^{(t)}  =  \Phi_{\rm reg}(\mathcal{F}_i^{(t)}) \in \mathbb{R}^{X \times Y \times 8K}
$, where each location of $\mathcal{O}_i^{(t)}$ represents the predicted K classes of 3D box $(x,y,z,h,w,l, \cos\alpha, \sin\alpha)$, denoting position residual, size, and angle.
We apply non-maximum suppression (NMS) to generate sparse 3D detections, and subsequently rasterize them into a binary BEV occupancy map $\mathbf{B}_i^{(t)}\in\{0,1\}^{X \times Y}$. The occupancy map $\mathbf{B}_i^{(t)}$ and BEV feature $\mathcal{F}_i^{(t)}$ then serve as input to the planning module. In this manner, the integration of structured occupancy data and detailed perceptual features offers complementary insights for waypoints planning.

\textbf{Waypoints planning.} The planning module takes a sequence of historical perceptual information and the future goal position as inputs, and outputs the trajectories that are progressing towards the goal, serving as a planner.
Given $T_h+1$ frames of historical BEV occupancy maps together with BEV features $\{\mathbf{B}_i^{(k)},\mathcal{F}_i^{(t)}\}_{k=t-T_{h}}^{t}$ and the goal position $G_i^{(t)}$ at timestamp $t$, the waypoints planning network $\Phi_{\rm way}(\cdot)$ generates the waypoints by
$
    \mathbf{W}_i^{(t)}  =  \Phi_{\rm way}(\{\mathbf{B}_i^{(k)},\mathcal{F}_i^{(t)}\}_{k=t-T_{h}}^{t},G_i^{(t)}) \in \mathbb{R}^{T_f\times 2},
$
where $\mathbf{W}_i^{(t)}$ represents the predicted trajectories of future $T_f$ timestamps, represented with $x, y$ coordinates. Here, we adopt MotionNet~\cite{Wu2020MotionNetJP} to extract spatial-temporal features from the sequence of BEV features and occupancy maps, and MLPs to encode goal position embedding.

\textbf{Controller.} The controller obtains executable driving actions from the predicted waypoints, including steer, throttle, and brake. Two PID controllers are utilized, one for lateral control and one for longitudinal control. The longitudinal controller uses the position vectors between consecutive waypoints, while the lateral controller considers the orientation. Our controller configuration follows ~\cite{LBC,TransFuser,TransFuser+}.

\vspace{-3mm}
\subsection{Driving-oriented collaboration}
\label{subsec:collaboration}

 
The end-to-end single-agent autonomous driving network unifies the driving-needed functions and achieves end-to-end autonomous driving. Driving-oriented collaboration leverages the collaboration capability of V2X communication to tackle the single agent's inevitable limited visibility issue through information sharing.

In this work, we present a novel driving-oriented collaboration scheme to optimize both driving performance and communication efficiency. This scheme includes i) driving-request-aware communication, we optimize communication efficiency by exchanging spatially sparse yet driving-critical BEV perceptual features by a driving request head; and ii) driving feature enhancement, we enhance the perceptual features of each agent with received messages. The enhanced BEV feature further serves to drive collaboration benefits throughout the entire system.




\textbf{Driving-request-aware communication.}
The driving-request-aware communication targets to pack the complementary perceptual information in driving-critical areas into a compact message. 
The core idea is to explore the spatial heterogeneity of driving requests and perceptual information.
The intuition is that: i) for driving requests, missing information at locations near the planned driving waypoints would cause catastrophic accidents, requesting and obtaining information at these locations could improve driving performance; and ii) for perceptual information, sending information at foreground areas helps recover the miss-detected objects due to the limited view, and background areas could be omitted to save precious bandwidth. Note that incorporating driving requests into communication is a novel design, diverging from previous communication strategies that specifically cater to the module-level perception utility. This new driving-request-aware communication allows for the optimization of valuable communication resources toward enhancing the ultimate system-level driving utility.

To achieve this, we enable a novel driving request map and a perceptual confidence map for each agent. 
The driving request map is implemented with a heatmap to highlight the regions around the planned driving route, where each element is negatively correlated with the distance to the waypoints predicted based on single agent's observation.
The perceptual confidence map is implemented with the object probability heatmap, where each element reflects the confidence of that spatial area containing objects.
Let $\mathbf{W}_i^{(t)}$ be the planned waypoints based on the single agent's observation, 
$\mathbf{S}_i^{(t)}$ be the object probability heatmap.
Then, the confidence map $\mathbf{C}_i^{(t)} \in \mathbb{R}^{X \times Y}$ is obtained with its elements given by
$    
\mathbf{C}_i^{(t)}(x,y)=\mathrm{max}(\mathbf{S}_i^{(t)}(x,y))
$, and the request map is given by 
$
    \mathbf{R}_i^{(t)}=\Phi_{\rm req}   (\mathbf{W}_i^{(t)}) 
$,
where $\Phi_{\rm req}(\cdot)$ is the request head, and we model the negative correlation with a Gaussian distribution with standard variance $\sigma$, which is widely used in the detection task to depict the importance decay pattern. The $(x,y)$th element of request map is  $\mathbf{R}_i^{(t)}(x, y) = {\rm exp}\left(-\frac{\left(x - W_x\right)^2 + \left(y - W_y\right)^2}{2\sigma^2}\right)$, where $(W_x, W_y)$ is the nearest waypoint. Confidence map $\mathbf{C}_i^{(t)}$ and request map $\mathbf{R}_i^{(t)}$ reflect foreground and waypoints probability from a bird's-eye view, and agents decide where to communicate based on these maps, offering spatially sparse yet driving-critical supportive features.

According to the driving request map and the perceptual confidence map, we further propose a BEV-based binary selection matrix to reflect where to communicate and then sample from BEV features $\{\mathcal{F}_{(i,l)}^{(t)}\}_{l=0}^2$. Let $\mathbf{M}^{(t)}_{j\rightarrow i} \in \{0, 1\}^{X \times Y}$ be a binary selection matrix to determine the spatial areas of messages sent from agent $j$ to $i$. Each element in $\mathbf{M}^{(t)}_{j\rightarrow i}$ determines whether the feature at the corresponding location should be selected to send. Thus, this sparse mask $\mathbf{M}^{(t)}_{j\rightarrow i}$ contributes to transmitting informative regions of the feature map, which are spatially sparse and critical for driving. We obtain the selection matrix $\mathbf{M}^{(t)}_{j\rightarrow i}$ by solving a constrained optimization problem conditioned on the $j$th agent's confidence map $\mathbf{C}_j^{(t)}$, the $i$th agent's request map $\mathbf{R}_i^{(t)}$ and the bandwidth limit $B$, as described in \eqref{sec4:problem_reformulate_2}
\begin{eqnarray}
\label{sec4:problem_reformulate_2}    
\max_{ \mathbf{M}_{j\to i}^{(t)} }  &&  \mathbf{M}_{j\to i}^{(t)} \odot \mathbf{C}_j^{(t)} \odot \mathbf{R}_i^{(t)}, 
\\  \nonumber
\mathrm{s.t.} &&  |\mathbf{M}_{j\to i}^{(t)}|\leq b,\mathbf{M}_{j\to i}^{(t)}\in\{0,1\}^{X\times Y},
\end{eqnarray}
where $b=B/\left(\Sigma_{l=0}^2 {2^{-l}D}\right)$, with its denominator derived from the ratio of total number of elements in $\{\mathcal{F}_{(i,l)}^{(t)}\}_{l=0}^2$ to the covered feature area. $\odot$ is element-wise multiplication. 

By introducing the selection matrix $\mathbf{M}_{j\to i}^{(t)}$,~\eqref{sec4:problem_reformulate_2} is actually a proxy of~\eqref{sec4:problem_formulation}. The objective of \eqref{sec4:problem_reformulate_2} ensures the priority of transmitting the perceptually informative features that are located within driving-critical areas. Fortunately, even with hard constraints and non-differentiability of binary variables, the optimization problem~\eqref{sec4:problem_reformulate_2} has a closed-form solution that satisfies the constraint. This solution is obtained by selecting those spatial regions whose corresponding elements rank top-$b$ in $\mathbf{C}_j^{(t)} \odot \mathbf{R}_i^{(t)}$. The detailed steps of selection function are: i) arrange the elements in the matrix $\mathbf{C}_j^{(t)} \odot \mathbf{R}_i^{(t)}$ in descending order; ii) given the communication budget constrain, decide the total number $b$ of communication regions; iii) set the spatial regions of $\mathbf{M}_{j\to i}^{(t)}$, where elements rank top-$b$ in $\mathbf{C}_j^{(t)} \odot \mathbf{R}_i^{(t)}$ as 1 and 0 verses.

The selected sparse feature map is then obtained as $\mathcal{Z}^{(t)}_{j\rightarrow i}= \{\mathcal{F}_{(j,l)}^{(t)} \odot \mathbf{M}^{(t,l)}_{j\rightarrow i}\}_{l=0}^2$, where $\mathbf{M}^{(t,0)}_{j\rightarrow i}=\mathbf{M}^{(t)}_{j\rightarrow i}$ and $\mathbf{M}^{(t,l)}_{j\rightarrow i}$ selects the regions of $\mathcal{F}_{(j,l)}^{(t)}$ with corresponding resolution. We obtain $\mathbf{M}^{(t,l)}_{j\rightarrow i}$ from $\mathbf{M}^{(t,l-1)}_{j\rightarrow i}$ by max-pooling to capture the same critical regions across different feature layers.
Overall, the message packs the selected feature map to provide supportive information and the driving request map to provide spatial priors to request driving-critical complementary information, this is, $\mathcal{P}^{(t)}_{i\rightarrow j}= (\mathcal{Z}^{(t)}_{i\rightarrow j},\mathbf{R}_i^{(t)})$. Only selected features and their indices are packed, significantly reducing communication costs while retaining sufficient information for perception and driving. 

It is worth noting that our perception and waypoints planning model are interconnected via the request map $\mathbf{R}_i^{(t)}$, which feeds the planning outcomes back to influence the front-end perception, establishing a feedback loop. As a result, the perception and planning modules collaboratively optimize the overarching driving objective, fostering synergy across AD system modules.



\textbf{Driving feature enhancement.} We enhance the feature of each agent by aggregating the received messages. Then, based on the augmented features, the driving-related signals across the entire system can be enhanced. Specifically, we utilize pixel-level Scaled Dot-Product Attention \cite{Attention_is_all_you_need,XuOPV2V:ICRA22} to fuse feature vectors from different agents within aligned locations. The non-parametric attention mechanism is computationally efficient in aggregating features, and it preserves robustness against disturbance by weighting feature vectors from each agent with scaled dot-product similarity.

At timestamp $t$, the $i$th agent receives messages $\{\mathcal{Z}^{(t)}_{j\rightarrow i}\}_{j\in\mathcal{N}_i}$ from other agents, where $\mathcal{N}_i$ denotes the set of collaborators whose detection range overlap with agent $i$. We also include the ego feature map in fusion and denote $\mathcal{Z}^{(t)}_{i\rightarrow i}=\{ \mathcal{F}_{(i,l)}^{(t)} \}_{l=0}^2$. The fused intermediate feature is obtained as 
$
\overline{\mathcal{F}}_{(i,l)}^{(t)}=\mathrm{softmax}\left(\mathcal{F}_{(i,l)}^{(t)} \mathcal{F}_{\mathcal{N}_i}^T/\sqrt{2^l D}\right)\mathcal{F}_{\mathcal{N}_i}
$, where $\mathcal{F}_{\mathcal{N}_i}=\left[\mathcal{F}_{(j,l)}^{(t)} \right]_{j\in\mathcal{N}_i}$ is the stack of neighbor features.
The collaborative perception feature is then obtained as 
$\overline{\mathcal{F}}_i^{(t)}= \Phi_{\rm scv}( \mathrm{concat}(\Phi_{\rm tcv}(\{\overline{\mathcal{F}}_{(i,l)}^{(t)}\}_{l=0}^2)))$, fulfilling the role of perception feature $\mathcal{F}_i^{(t)}=\Phi_{\rm enc}(\mathcal{X}_i^{(t)})$ in single-agent network in Section \ref{subsec:end2end}.

The collaborative perception feature $\overline{\mathcal{F}}_i^{(t)}$ propagates benefits across the entire system, enhancing all driving signals.  
Following the pipeline in Section \ref{subsec:end2end}, we similarly derive an object probability heatmap as $\overline{\mathbf{S}}_i^{(t)}=\Phi_{\rm cls}(\overline{\mathcal{F}}_i^{(t)})$ and a bounding box regression map as $\overline{\mathbf{O}}_i^{(t)}=\Phi_{\rm reg}(\overline{\mathcal{F}}_i^{(t)})$. Subsequently, the collaborative BEV occupancy map $\overline{\mathbf{B}}_i^{(t)}$ is generated using $\overline{\mathbf{S}}_i^{(t)}$ and $ \overline{\mathbf{O}}_i^{(t)}$ via NMS. During the waypoints planning phase, $\overline{\mathbf{B}}_i^{(t)},\overline{\mathcal{F}}_i^{(t)}$ substitute $\mathbf{B}_i^{(t)},\mathcal{F}_i^{(t)}$ in Section~\ref{subsec:end2end}, and are utilized to generate the improved waypoints via $\overline{\mathbf{W}}_i^{(t)}  =  \Phi_{\rm way}(\{\overline{\mathbf{B}}_i^{(k)},\overline{\mathcal{F}}_i^{(t)}\}_{k=t-T_{h}}^{t},G_i^{(t)})$. Then, the controller generates the final driving actions based on $\overline{\mathbf{W}}_i^{(t)}$.

To sum up, CoDriving first enhances single-agent detection capabilities by leveraging shared BEV features among collaborators. Specifically, these shared BEV features, drawn from various perspectives, synergize to create a more precise and complete representation of the driving-relevant environment.  Second, this enhanced BEV representation results in collaborative detections, which can better reflect the ground truth objects. Such a detailed grasp of the surroundings would simplify the perceiving of dynamics, facilitating the formulation of safer driving plans. Third, the enhanced detection capabilities inherent in collaborative autonomous driving yield more accurate occupancy maps. Fourth, the enhanced BEV representations and occupancy maps serve as critical inputs for the waypoints planning network. With these improved inputs, the network can then generate more precise waypoints plannings. Finally, this improvement directly contributes to safer driving maneuvers, especially in scenarios where objects are obscured in individual view.


\vspace{-4mm}
\subsection{Losses and training}
\label{subsec:losses}
To train the overall system, we supervise two tasks: 3D detection and waypoints planning. The overall loss is
\begin{equation*}
        L = L_{\rm cls}\left(\overline{\mathbf{C}}^{(t)}_i,\widehat{\mathbf{C}}^{(t)}_i \right) + L_{\rm reg}\left( \overline{\mathcal{O}}^{(t)}_i,\widehat{\mathcal{O}}^{(t)}_i \right) + L_{\rm way} \left(\overline{\mathbf{W}}_i^{(t)},\widehat{\mathbf{W}}_i^{(t)} \right)
\end{equation*}
where $\{ \widehat{\mathbf{C}}^{(t)}_i, \widehat{\mathcal{O}}^{(t)}_i \}$ and $\widehat{\mathbf{W}}_i^{(t)}$ represent ground-truth objects and waypoints for the $i$th agent.
$L_{\rm cls}+L_{\rm reg}$ and $L_{\rm way}$ are the centerpoint detection loss~\cite{zhou2019objects} and planning loss~\cite{Hu2020CollaborativeMP}. Specifically, we implement $L_{\rm cls}$ with pixel-wise Gaussian focal loss and $L_{\rm reg}, L_{\rm way}$ with weighted L1 loss.

\textcolor{blue}{The training pipeline consists of two stages. First, the perception module is trained from scratch with supervision from the detection loss. 
In the second stage, we freeze the perception module and train the planner for waypoint planning. During this phase, we randomly set the feature selection rate of the matrix $\mathbf{M}^{(t)}_{j\rightarrow i}$ within the range of 0 to 1, with the logarithm of the selection rate following a uniform distribution. This strategy simulates the uncertain communication bandwidth and helps the system adapt to varying bandwidth conditions by augmenting the transmitted features. In this way, the system is further optimized to fit the objective in~\eqref{sec4:problem_formulation}. The random selection rate is introduced in the later stages of training to ensure convergence.}

\vspace{-4mm}
\subsection{Discussion}
CoDriving has two distinct advantages. First, CoDriving breaks the limitation of single agent perception by sharing BEV features and comprehensively enhances the performance of the entire system, encompassing improvements in perception, planning, and driving behavior. Second, CoDriving achieves driving-oriented communication, it considers the driving request and specifically targets the driving-critical regions, which improves communication efficiency by transmitting spatially compact features.

Compared to existing collaborative perception methods~\cite{LiuWho2com:ICRA20,liu2020when2com,WangV2vnet:ECCV20,LiLearning:NeurIPS21,Li_2021_RAL,xu2022v2xvit,XuCoBEVT:CoRL22,HuWhere2comm:NeurIPS22,LuRobust:ICRA23,zimmer2024tumtraf}, CoDriving shows distinction in both aspects of system purpose and collaborative information selecting. First, CoDriving aims to extend beyond environmental perception, producing waypoints planning and driving control signals as outputs, 
while collaborative perception methods only focus on addressing perception tasks. Second, CoDriving optimizes the selection of collaborative information tailored for driving tasks.
This driving-centric collaboration is facilitated through a feedback mechanism, wherein outputs from the downstream planning module are utilized to highlight areas of driving requests, and these driving request maps are then relayed back to the front-end perception module. Such a process underscores the collaboration within a unified system. 
In comparison, collaborative perception methods optimize the selection of shared information tailored for perceptual tasks.

Compared to existing end-to-end autonomous driving systems~\cite{TCP,InterFuser,TransFuser+,LAV,chen2021learning,shao2023reasonnet,UniAD}, CoDriving enables feature sharing among multiple agents, enhancing the perception and planning ability of each participant. Especially, compared to Coopernaut~\cite{CuiCoopernaut:CVPR22}, our system outperforms in three aspects: i) CoDriving demonstrates adaptability to any communication bandwidth by solving the optimization problem~\eqref{sec4:problem_reformulate_2}, while Coopernaut relies on voxel pooling and can only handle a fixed communication bandwidth; ii) CoDriving adapts input modality of either RGB image or point clouds, while Coopernaut is designed for point clouds only; 
and iii) CoDriving facilitates driving-critical BEV feature sharing, enhancing interpretability, while Coopernaut communicates all the encoded point representations.

\vspace{-4mm}
\section{\textcolor{blue}{Perception} evaluation}
\label{sec:Benchmark_perception}

In this section, we compare the proposed CoDriving with previous state-of-the-art methods in collaborative perception using offline collaborative perception \textcolor{blue}{datasets~\cite{YuDAIRV2X:CVPR22,XuV2V4Real:CVPR23,XuOPV2V:ICRA22,LiV2XSim:RAL22,zimmer2024tumtraf}, covering both real-world and simulated scenarios. The evaluation includes the trade-off between perception performance and communication cost under both homogeneous and heterogeneous scenarios, and the robustness against two practical issues including communication latency and pose errors.} We begin by outlining the task setting, evaluation metrics, and details of the dataset used. Following these, the experimental results are presented. Note that CoDriving is a full-spectrum collaborative autonomous driving system, with capabilities in perception, planning, planning, and control. Here, we retain only the perception module of CoDriving and prune the subsequent modules, allowing it to be trained solely with annotations relevant to perception. 

\begin{figure*}[!t]
\centering
\begin{subfigure}{0.19\linewidth}
    \includegraphics[width=0.99\linewidth]{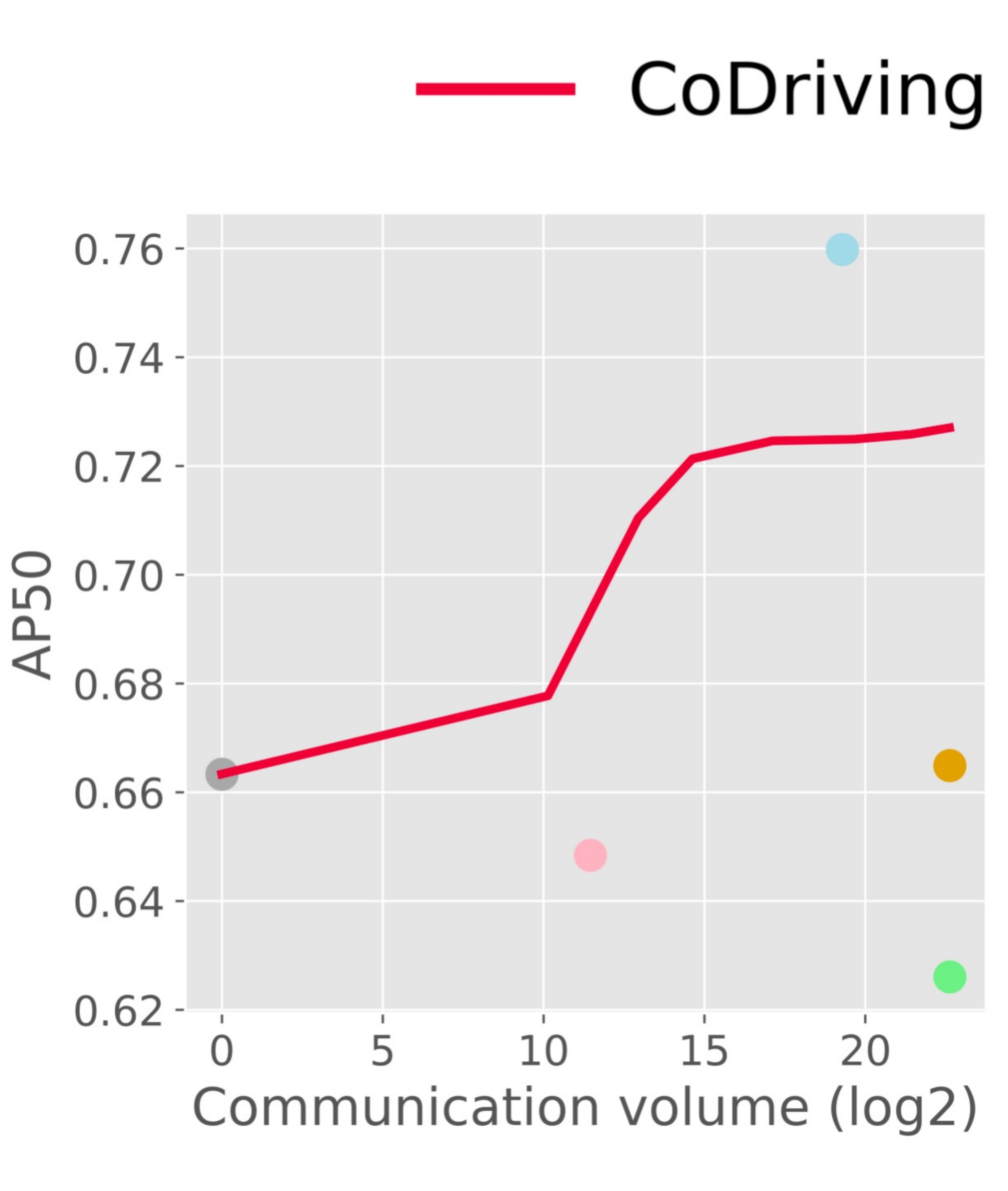}
    \vspace{-5mm}
  \caption{DAIR-V2X}
  \label{Fig:DAIRV2X_SOTAs}
  \end{subfigure}
\begin{subfigure}{0.19\linewidth}
    \vspace{0mm}
    \includegraphics[width=0.99\linewidth,height=1.165\linewidth]{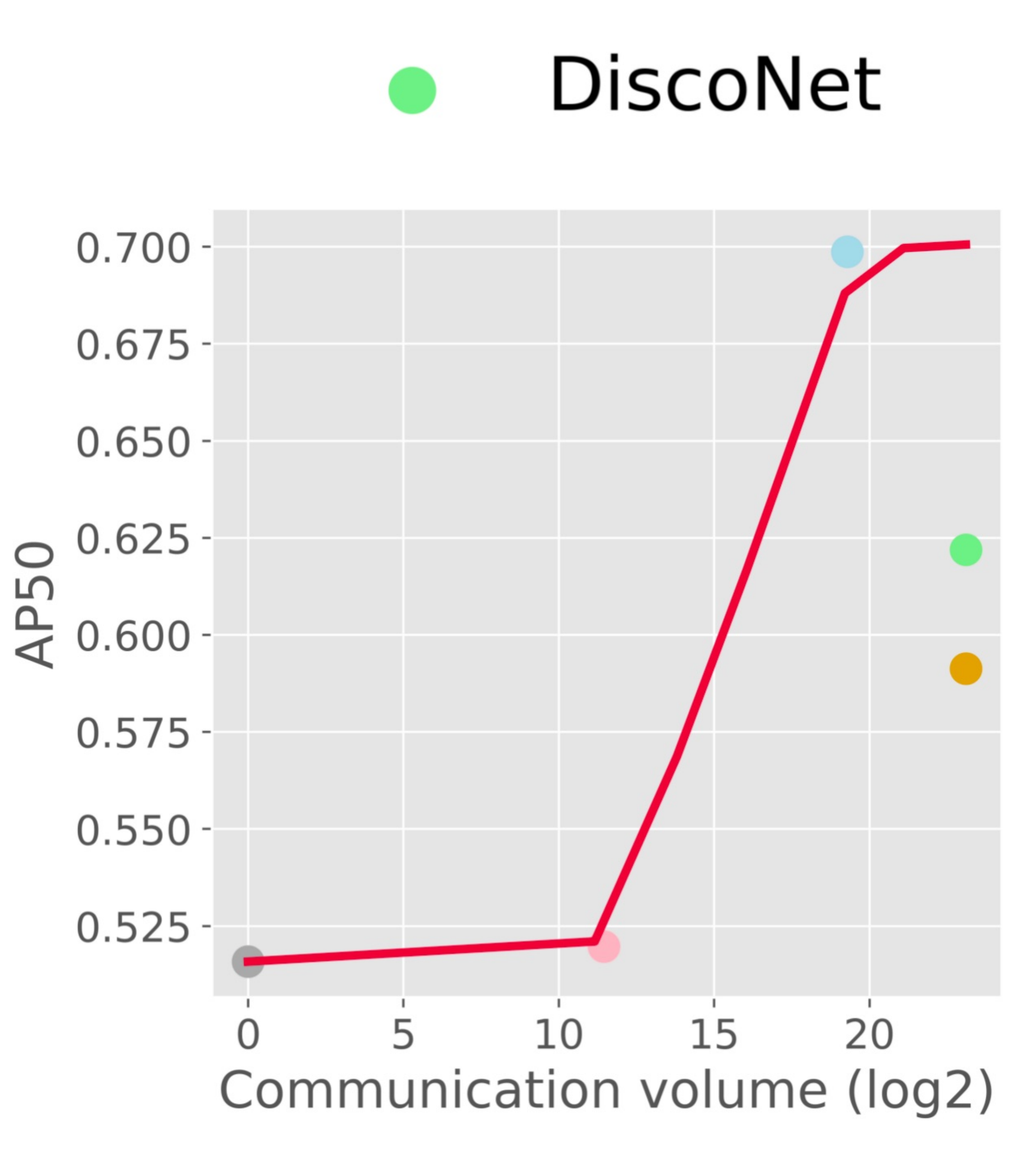}
    \vspace{-4.9mm}
  \caption{V2V4Real}
  \label{Fig:V2V4Real_SOTAs}
  \end{subfigure}
\begin{subfigure}{0.19\linewidth}
    \includegraphics[width=0.99\linewidth]{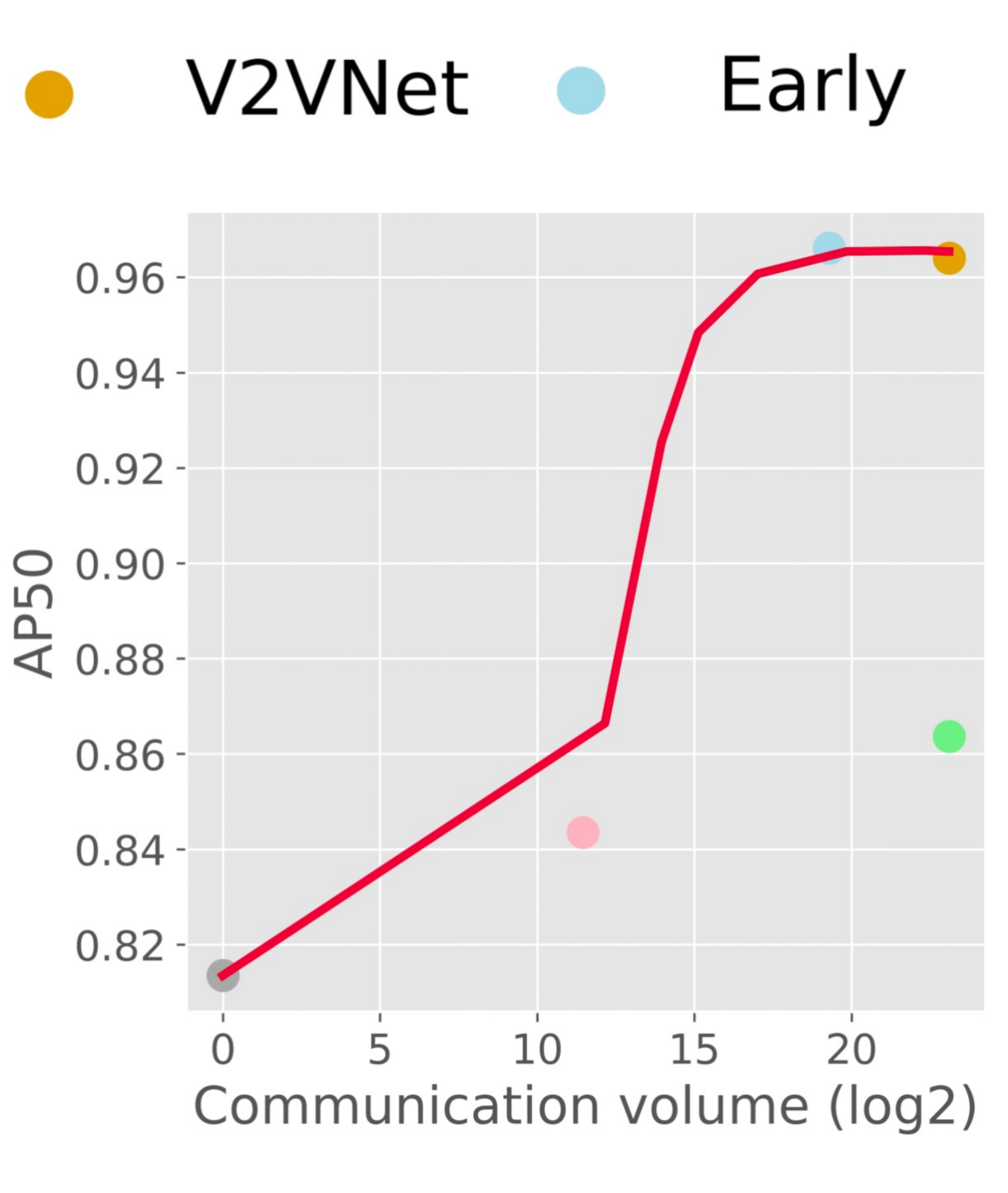}
    \vspace{-5mm}
  \caption{OPV2V}
  \label{Fig:OPV2V_SOTAs}
  \end{subfigure}
\begin{subfigure}{0.19\linewidth}
    \includegraphics[width=0.99\linewidth]{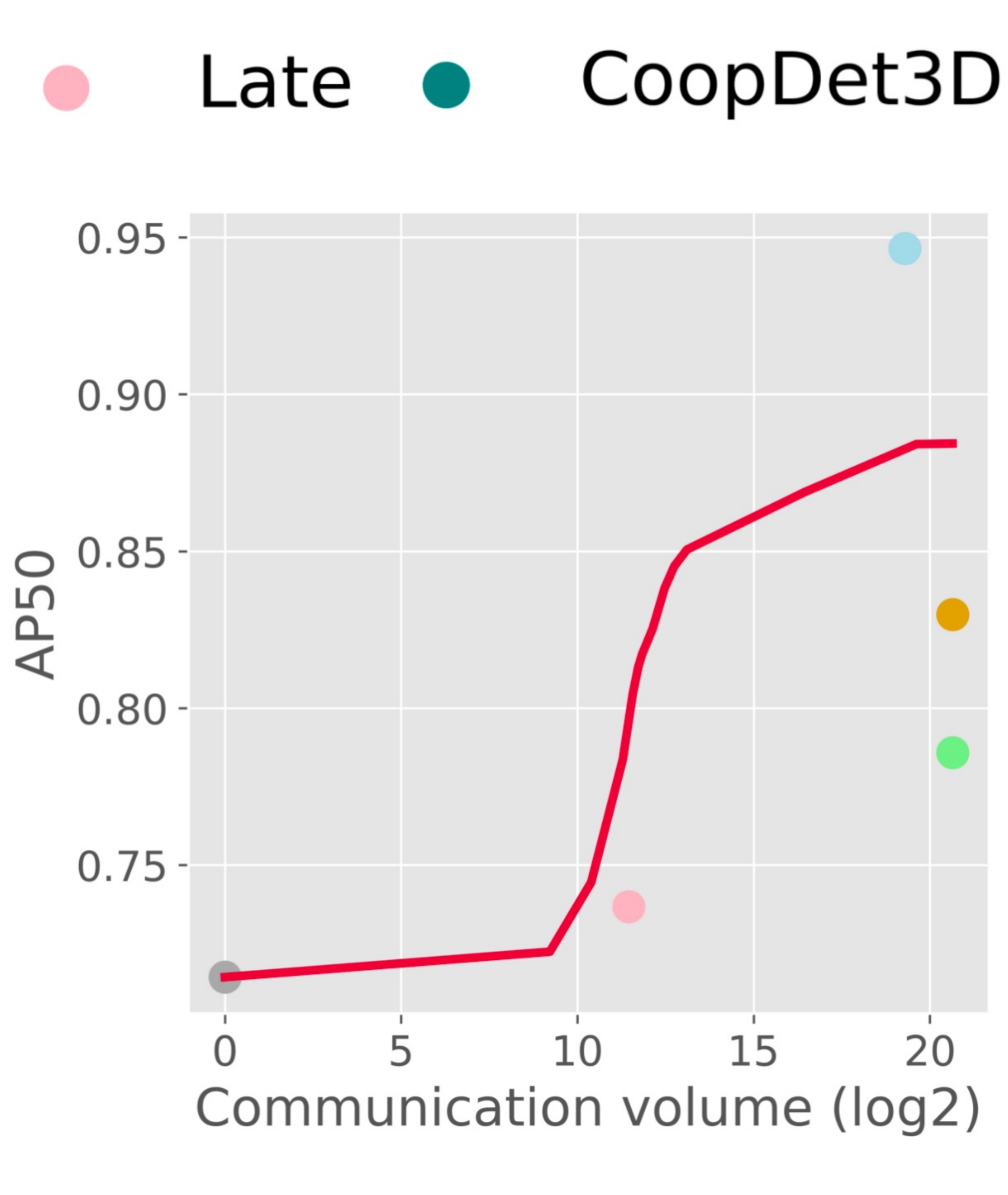}
    \vspace{-5mm}
  \caption{V2XSIM2.0}
  \label{Fig:V2XSIM_SOTAs}
  \end{subfigure}
\begin{subfigure}{0.19\linewidth}
    \includegraphics[width=0.99\linewidth]{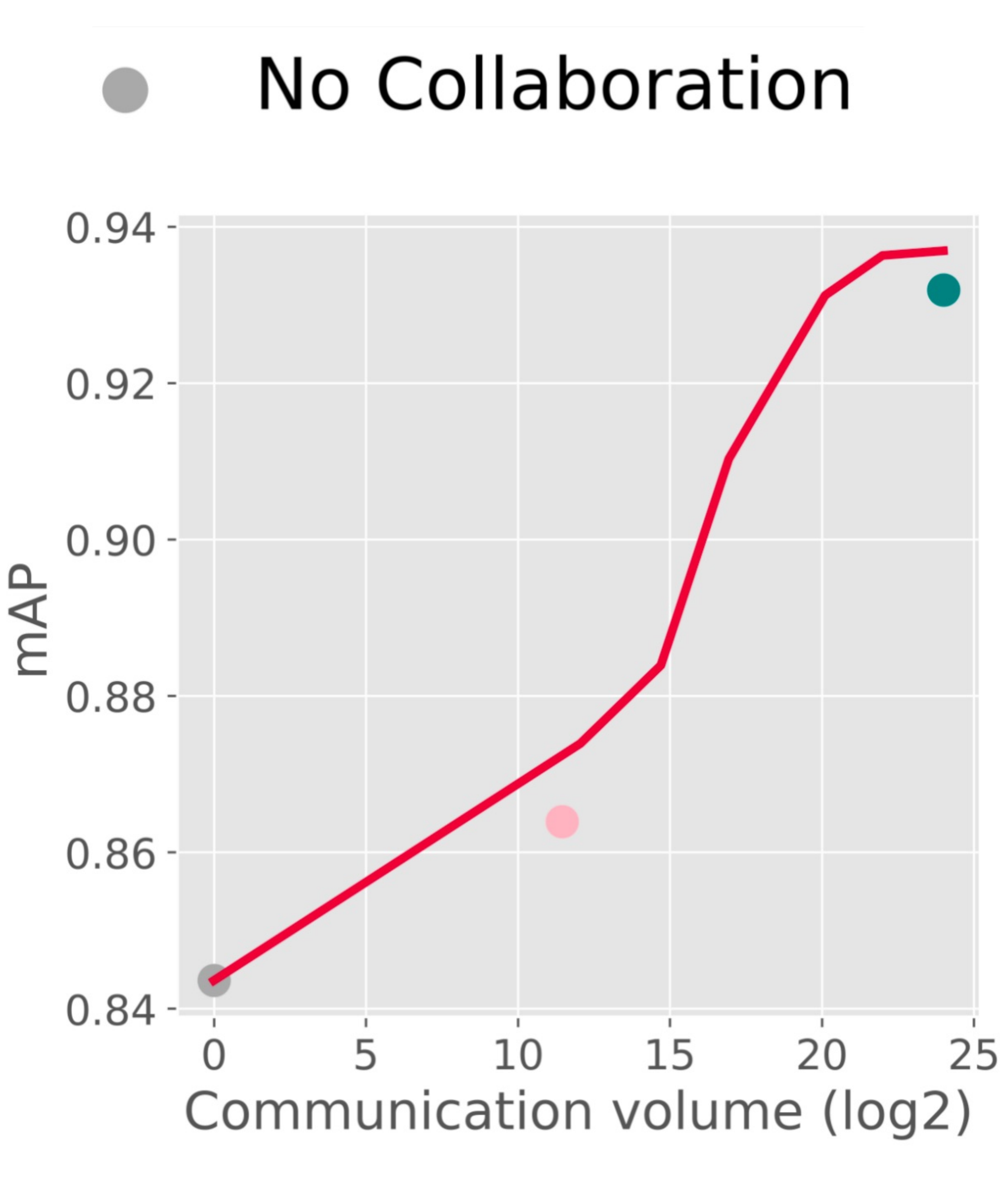}
    \vspace{-5mm}
  \caption{\textcolor{blue}{TUMTraf-V2X}}
  \label{Fig:TUMV2X_SOTAs}
  \end{subfigure}
\vspace{-2mm}
\caption{\textcolor{blue}{Benchmark CoDriving and previous collaborative perception methods on commonly used collaborative perception real-world and simulation datasets under homogeneous setting.}}
\vspace{-4mm}
\label{Fig:SOTAs}
\end{figure*}

\begin{figure*}[!t]
\centering
\begin{subfigure}{0.48\linewidth}
    \includegraphics[width=0.99\linewidth]{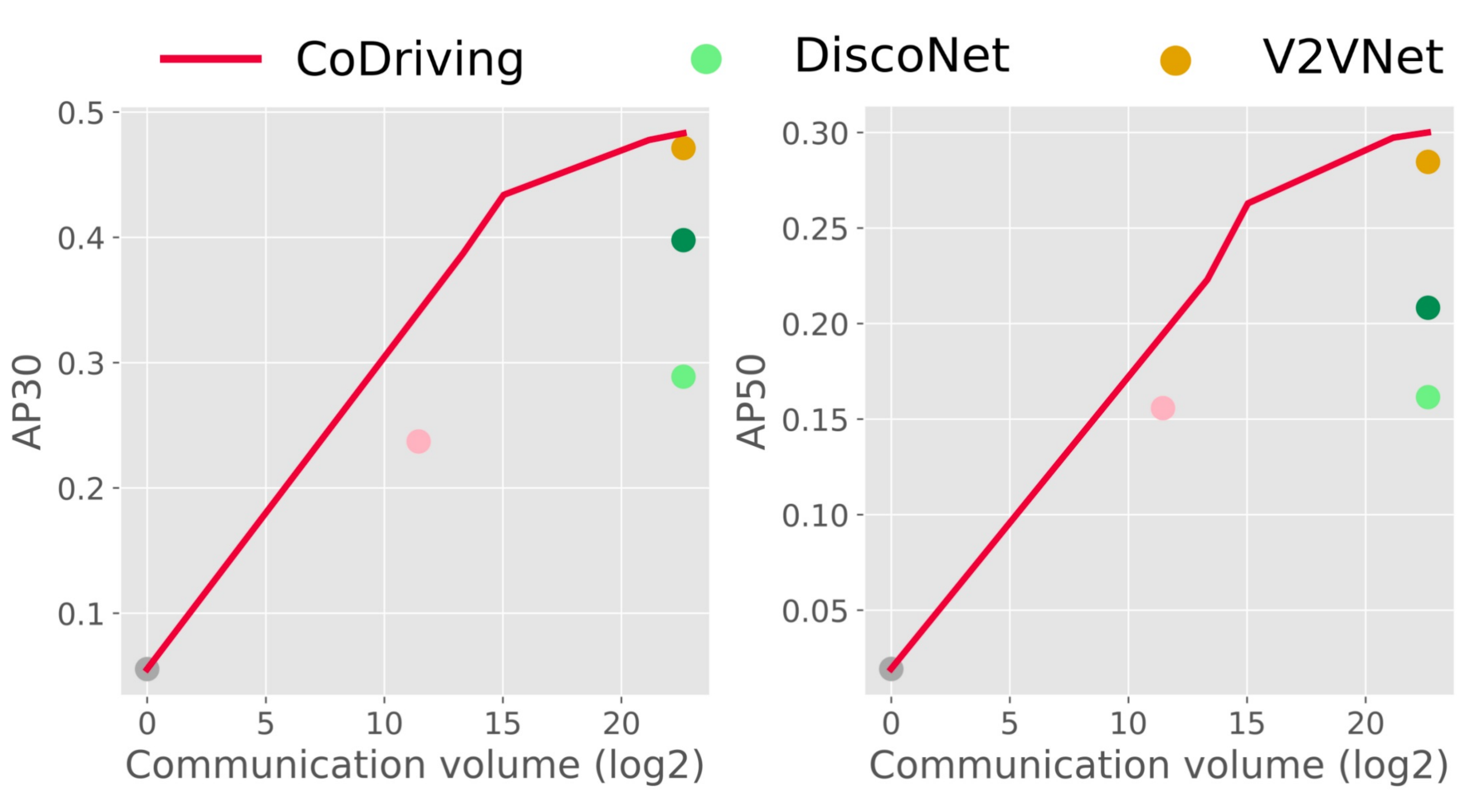}
    \vspace{-3mm}
  \caption{DAIR-V2X}
  \vspace{-0mm}
  \label{Fig:DAIRV2X_Hete}
  \end{subfigure}
\begin{subfigure}{0.48\linewidth}
    \includegraphics[width=0.99\linewidth]{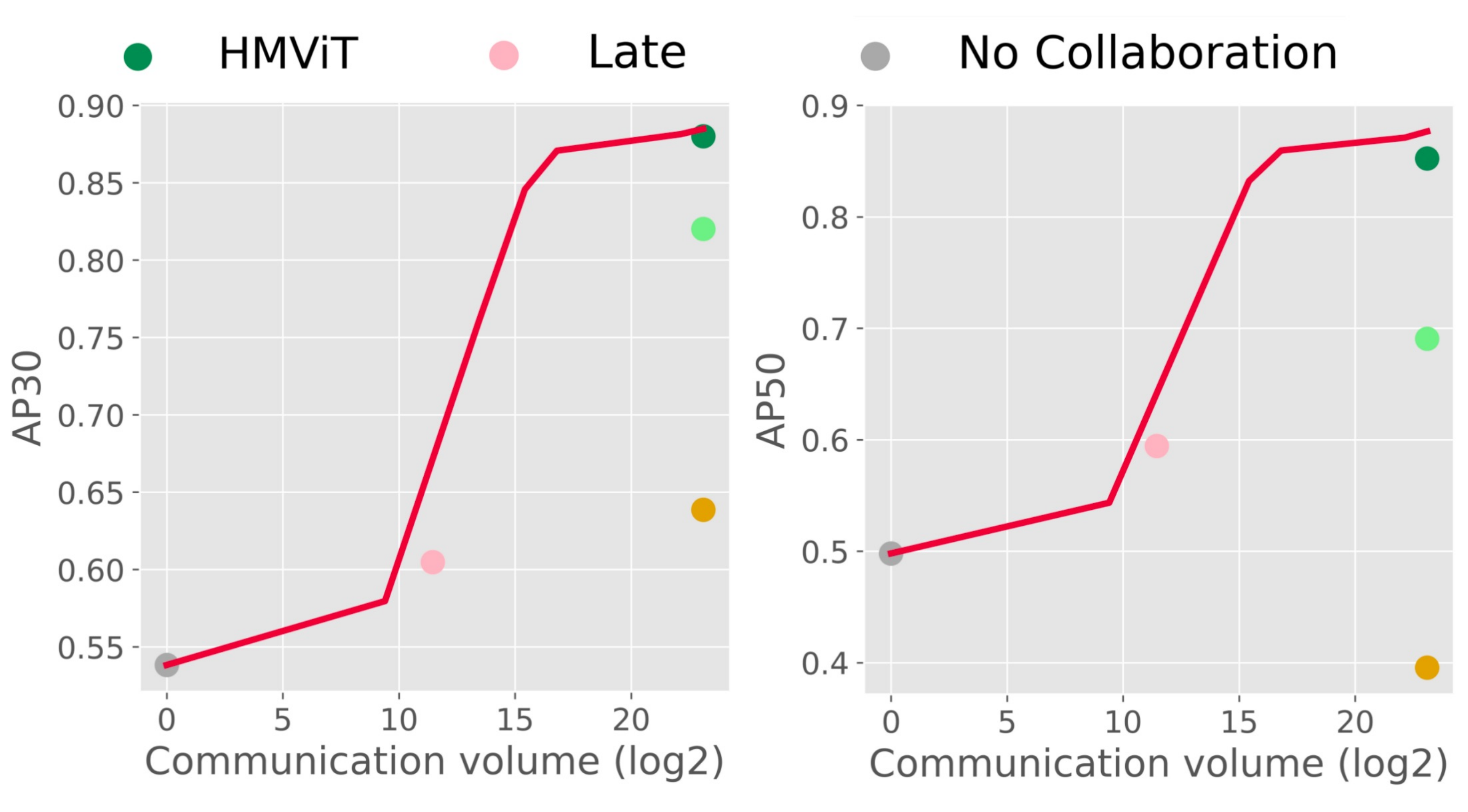}
    \vspace{-3mm}
  \caption{OPV2V}
  \vspace{-0mm}
  \label{Fig:OPV2V_Hete}
  \end{subfigure}
\vspace{-2mm}
\caption{Benchmark CoDriving and previous collaborative perception methods on commonly used collaborative perception real-world and simulation datasets under heterogeneous settings.}
\vspace{-5mm}
\label{Fig:Hete}
\end{figure*}

\vspace{-3mm}
\subsection{Tasks and metrics}
\label{subsec:det_metric}
We follow the common collaborative perception task setting, concentrating on collaborative 3D detection tasks. Here, each agent has a specific detection objective: to identify all objects within a predefined spatial area, based on its sensor data and collaborative messages enabled by communication.
To evaluate this task, we consider several assessment scenarios: i) the trade-off between performance and communication cost, evaluating collaborative detection capabilities under varying communication bandwidth constraints; ii) heterogeneous scenarios, where agents in the same collaborative scene are equipped with different types of sensors. Here, we randomly assign agents in the scene either LiDAR or camera, resulting in a balanced 1:1 ratio of agents across the different modalities; iii) communication latency issues, evaluating collaborative detection's performance when there are delays in the received collaborative messages; and iv) pose error issues, evaluating performance when the collaborative messages received are incorrectly localized.

\noindent\textbf{Detection performance.} Following the collaborative perception methods~\cite{HuWhere2comm:NeurIPS22,XuOPV2V:ICRA22,LiV2XSim:RAL22,zimmer2024tumtraf}, the detection results are evaluated by 1) Average Precision (AP) at Intersection-over-Union (IoU) thresholds of 0.30, 0.50. 2) Mean average precision (mAP) in BEV perspective, considering the BEV center distance.

\noindent\textbf{Communication cost.} Following the collaborative perception methods~\cite{HuWhere2comm:NeurIPS22,LiLearning:NeurIPS21}, the communication cost in the feature-based intermediate collaboration setting is calculated as ${\rm log}_2(H\times W \times ||\mathbf{M}||_1\times C\times 32/8)$, where $\mathbf{M}$ is the selection matrix representing the selected feature to be packed in the messages. Here, $32$ represents the float32 data type, and $8$ converts bits to bytes. 

\noindent\textbf{Communication latency and pose error settings.} For communication latency, we follow SyncNet~\cite{LeiLatency:ECCV22}, using latency varying from 0 ms to 500 ms. For pose error issues, following setting in CoAlign~\cite{LuRobust:ICRA23}, we use Gaussian noise with a mean of 0m/0$^\circ$ and standard deviations ranging from 0m/0$^\circ$ to 0.6m/0.6$^\circ$.

\vspace{-3mm}
\subsection{Datasets}
\vspace{-2mm}

We conduct comprehensive experiments on collaborative 3D object detection tasks using five commonly used datasets, including the real-world datasets, DAIR-V2X~\cite{YuDAIRV2X:CVPR22}, V2V4Real~\cite{XuV2V4Real:CVPR23}, TUMTraf-V2X~\cite{zimmer2024tumtraf}, and simulation datasets, OPV2V~\cite{XuOPV2V:ICRA22} and V2X-SIM2.0~\cite{LiV2XSim:RAL22}.


\noindent\textbf{DAIR-V2X.} DAIR-V2X~\cite{YuDAIRV2X:CVPR22} is a popular real-world collaborative perception dataset. Each scene contains two agents: a vehicle and a roadside unit. Each agent is equipped with a LiDAR and a camera. The perception range is 204.8m$\times$102.4m. 

\noindent\textbf{V2V4Real.} V2V4Real~\cite{XuV2V4Real:CVPR23} is a large real-world vehicle-to-vehicle collaborative perception dataset. It includes a total of $20$K frames of LiDAR point cloud captured by Velodyne VLP-32 sensor and $40$K frames of RGB images captured by two mono cameras (front and rear) with $240$K annotated 3D bounding boxes. The perception range is 280m$\times$80m. 

\noindent
\textcolor{blue}{\textbf{TUMTraf-V2X.} TUMTraf-V2X~\cite{zimmer2024tumtraf} is a real-world collaborative perception dataset. It includes sensor data from 4 roadside cameras, 1 roadside LiDAR, 1 vehicle camera, and 1 vehicle LiDAR. The perception range is 150m$\times$150m.}

\noindent\textbf{OPV2V.} OPV2V~\cite{XuOPV2V:ICRA22} is a vehicle-to-vehicle collaborative perception simulation dataset, co-simulated by OpenCDA~\cite{XuOpenCDA:ITSC2021} and CARLA~\cite{DosovitskiyCARLA:CoRL2017}. It includes $12$K frames of 3D point clouds and RGB images with $230$K annotated 3D boxes. The perception range is 280m$\times$80m. 

\noindent\textbf{V2X-SIM2.0.} V2X-SIM2.0~\cite{LiV2XSim:RAL22} is a vehicle-to-vehicle collaborative perception simulation dataset. Each scene contains a 20-second traffic flow at a certain intersection of three CARLA towns. It includes $47.2$k samples with $10$k frames of 3D point clouds and RGB images. The perception range is 100m$\times$80m. 


\begin{figure*}[!t]
\centering
\begin{subfigure}{0.48\linewidth}
\hspace{-1.5mm}
    \includegraphics[width=0.99\linewidth]{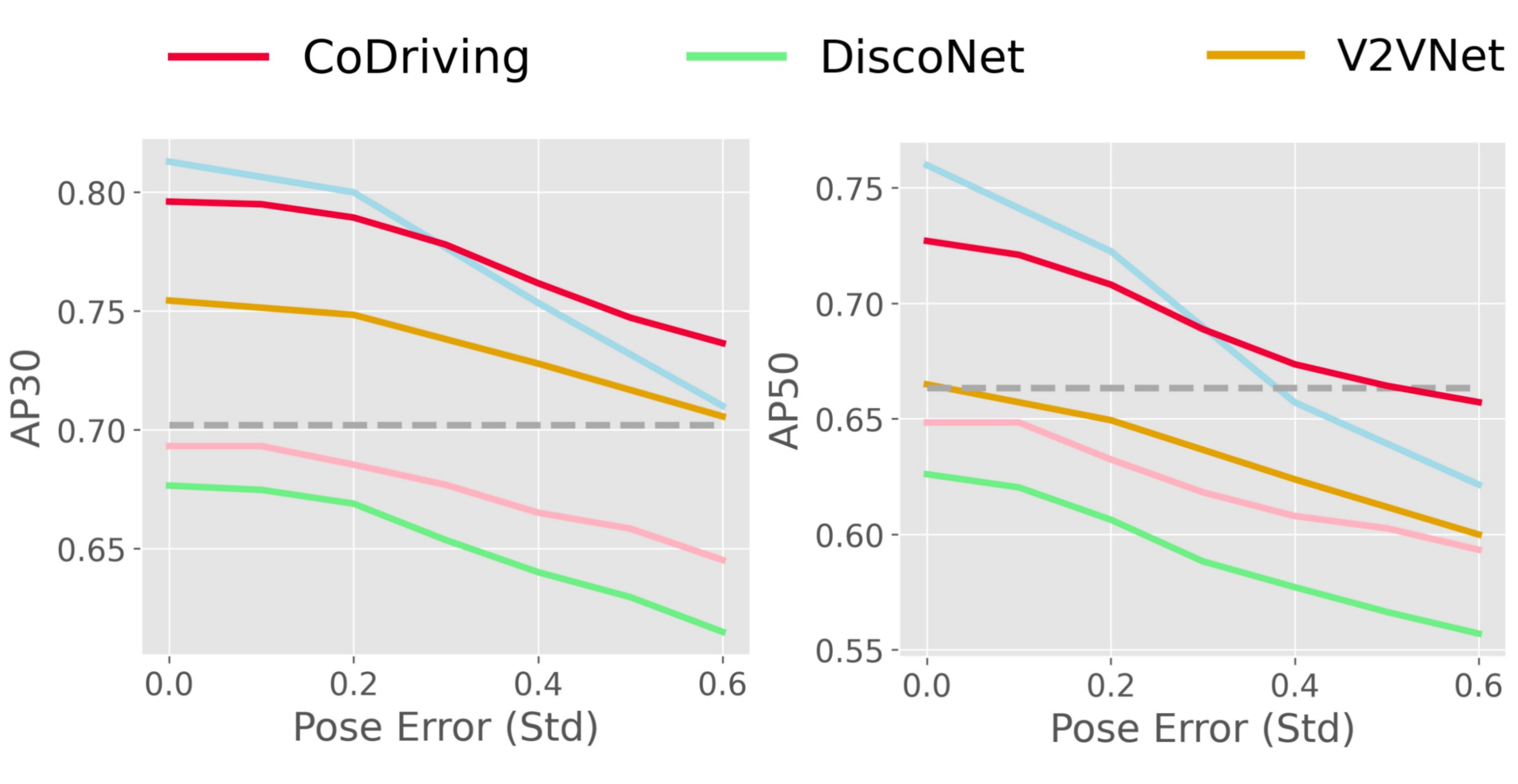}
    \vspace{-3mm}
  \caption{DAIR-V2X}
  \vspace{-0mm}
  \label{Fig:DAIRV2X_PE}
  \end{subfigure}
\begin{subfigure}{0.48\linewidth}
\hspace{-2mm}
    \raisebox{0.4mm}{\includegraphics[width=0.99\linewidth]{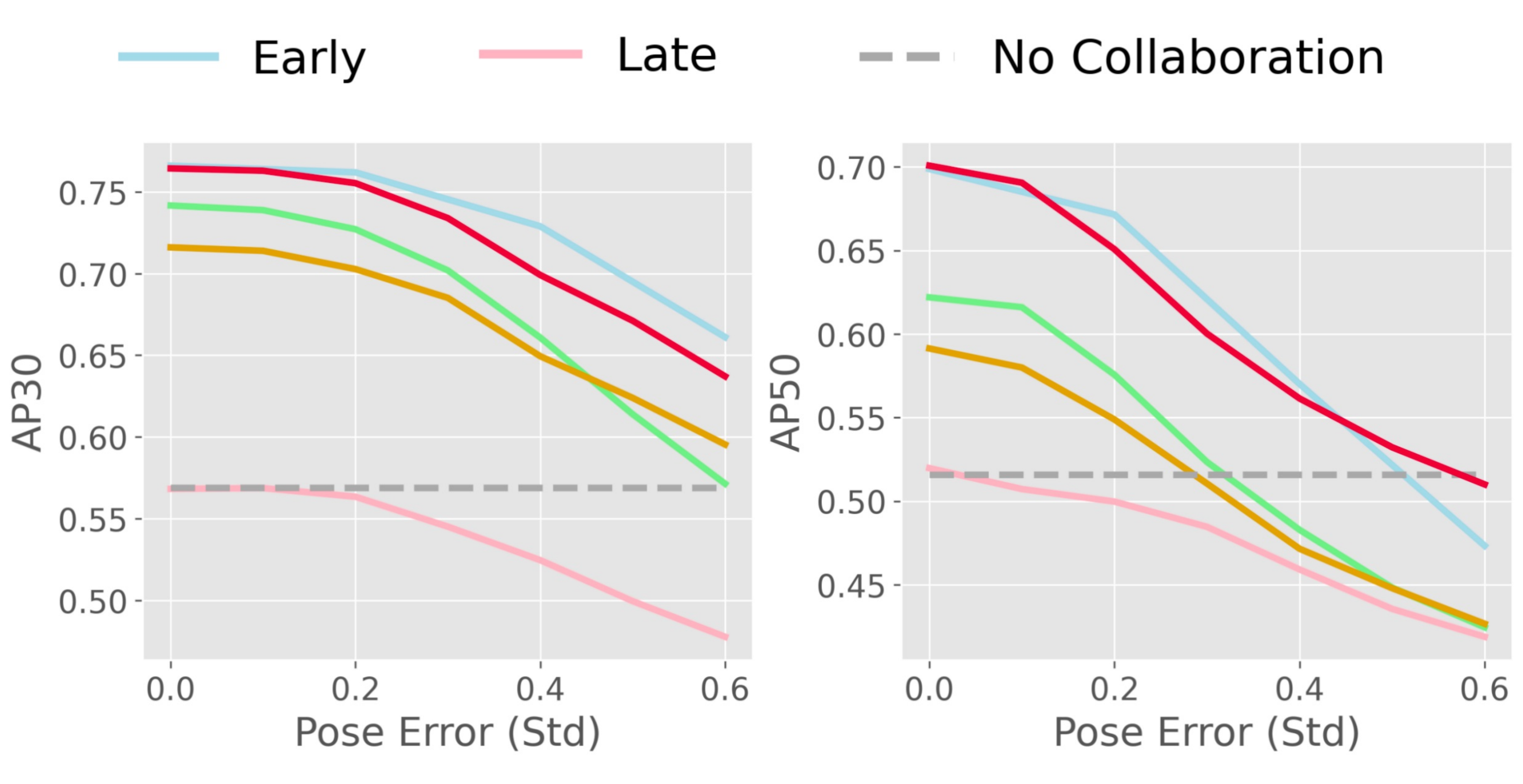}}
    \vspace{-3mm}
  \caption{V2V4Real}
  \vspace{-0mm}
  \label{Fig:V2V4Real_PE}
  \end{subfigure}
\begin{subfigure}{0.48\linewidth}
    \includegraphics[width=0.99\linewidth]{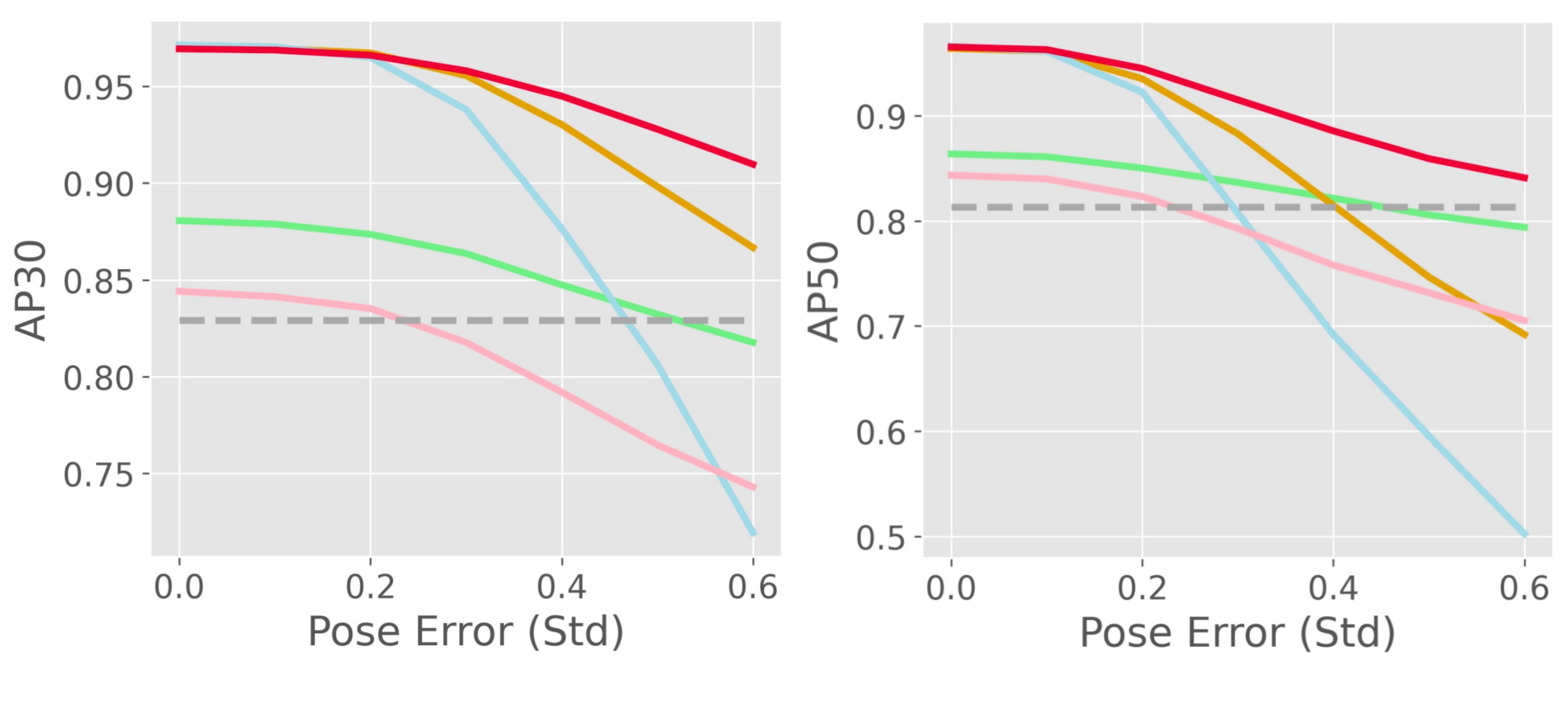}
    \vspace{-3mm}
  \caption{OPV2V}
  \vspace{-1mm}
  \label{Fig:OPV2V_PE}
  \end{subfigure}
\begin{subfigure}{0.48\linewidth}
    \includegraphics[width=0.99\linewidth,height=0.445\linewidth]{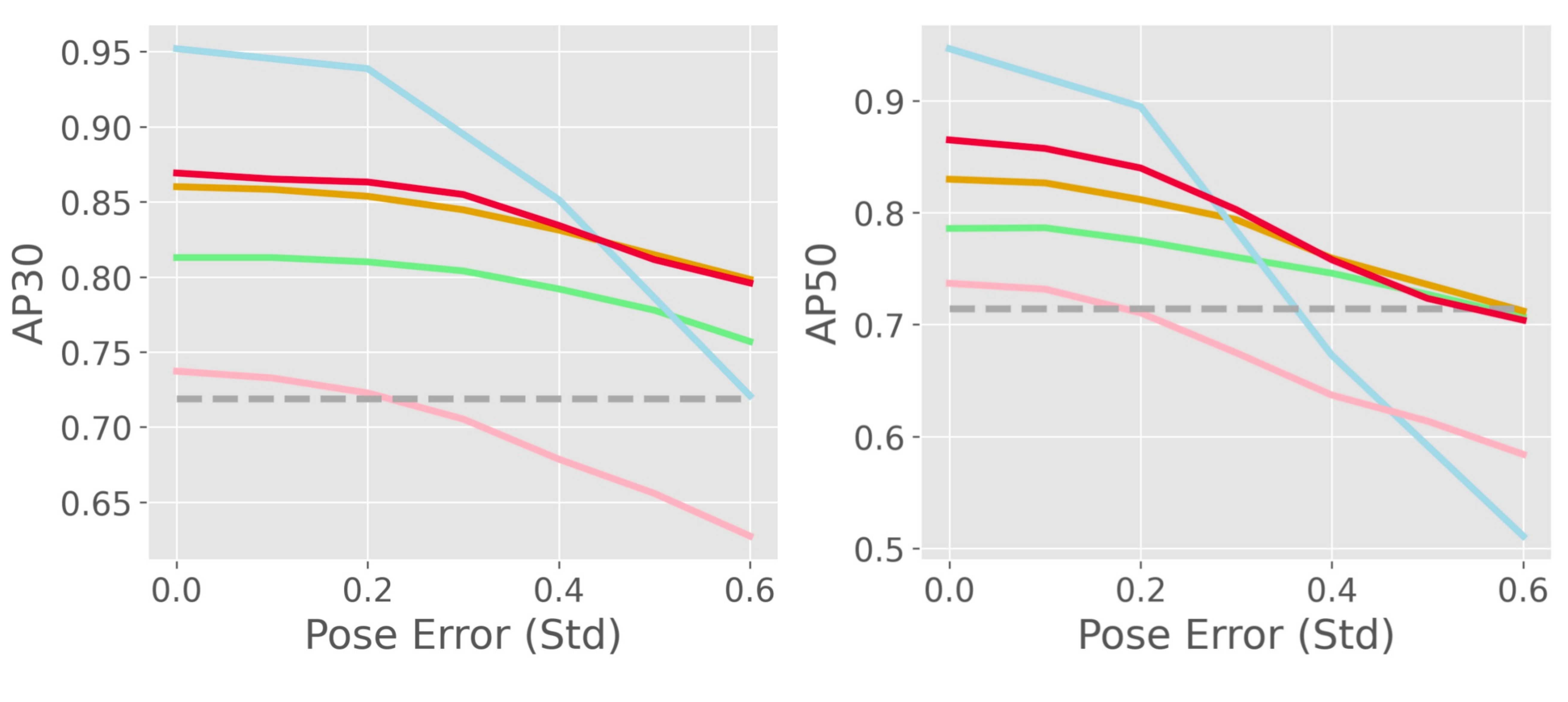}
    \vspace{-3mm}
  \caption{V2XSIM2.0}
  \vspace{-1mm}
  \label{Fig:V2XSIM_PE}
  \end{subfigure}
\vspace{-2mm}
\caption{The robustness to pose error of CoDriving on commonly used collaborative perception real-world and simulation datasets.}
\vspace{-4mm}
\label{Fig:PE}
\end{figure*}

\begin{figure*}[!t]
\centering
\begin{subfigure}{0.48\linewidth}
    \hspace{-2mm}
    \includegraphics[width=0.99\linewidth]{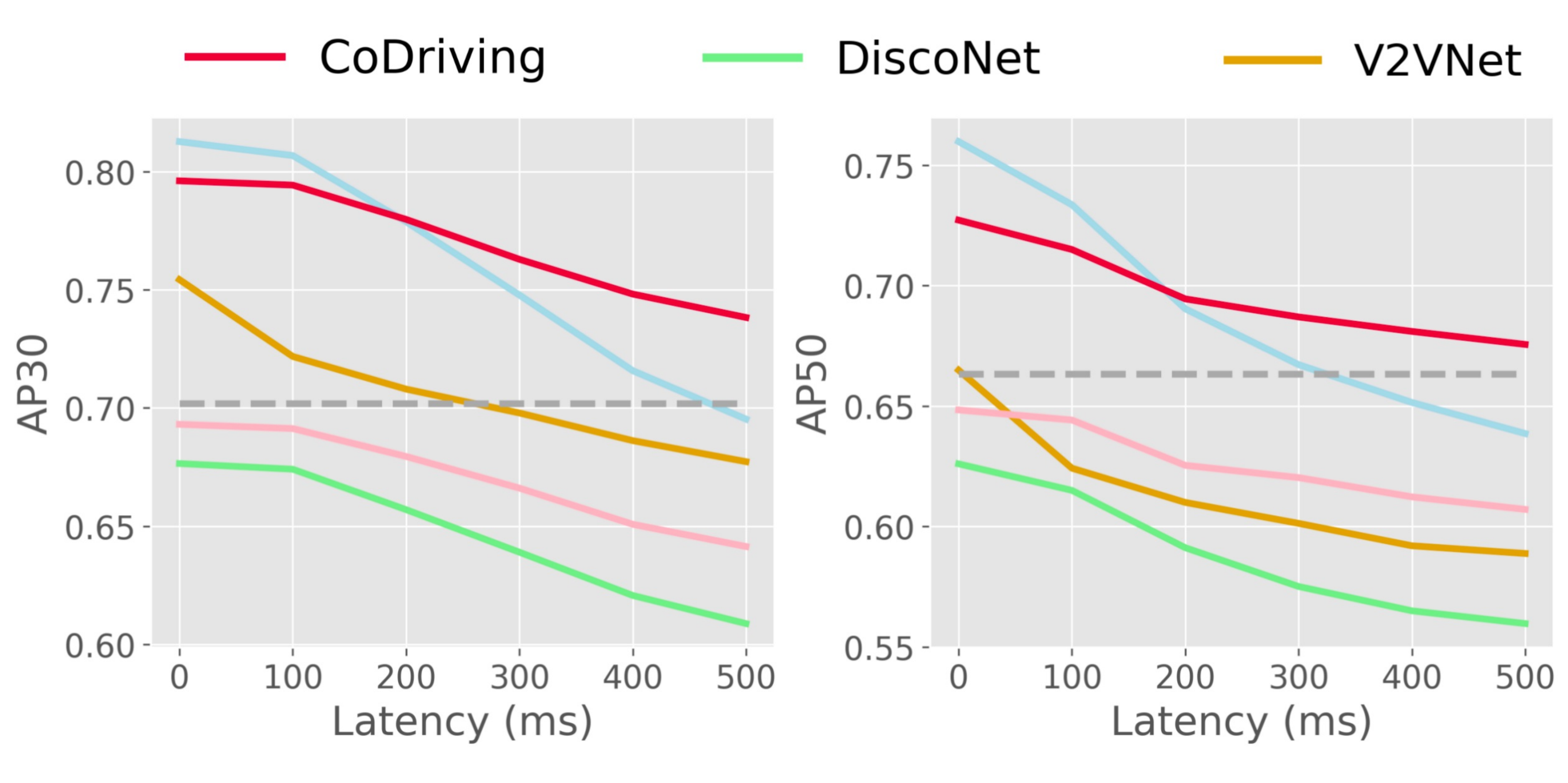}
    \vspace{-3mm}
  \caption{DAIR-V2X}
  \label{Fig:DAIRV2X_latency}
  \end{subfigure}
\begin{subfigure}{0.485\linewidth}
    \hspace{-2mm}
    \raisebox{-0.0mm}{\includegraphics[width=0.99\linewidth, height=0.490\linewidth]{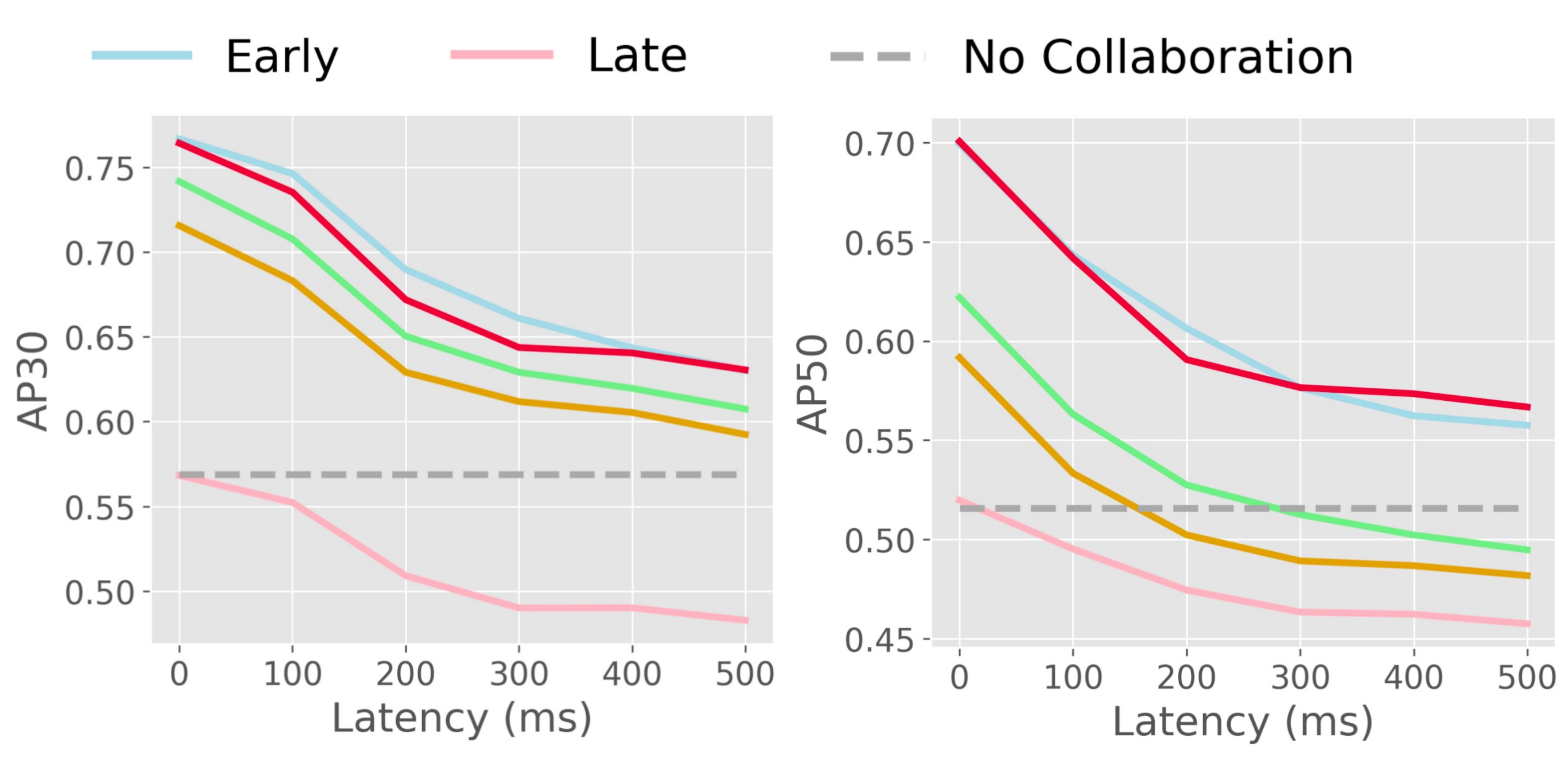}}
    \vspace{-3mm}
  \caption{V2V4Real}
  \label{Fig:V2V4Real_latency}
  \end{subfigure}
\begin{subfigure}{0.48\linewidth}
\hspace{0mm}
    \includegraphics[width=0.97\linewidth]{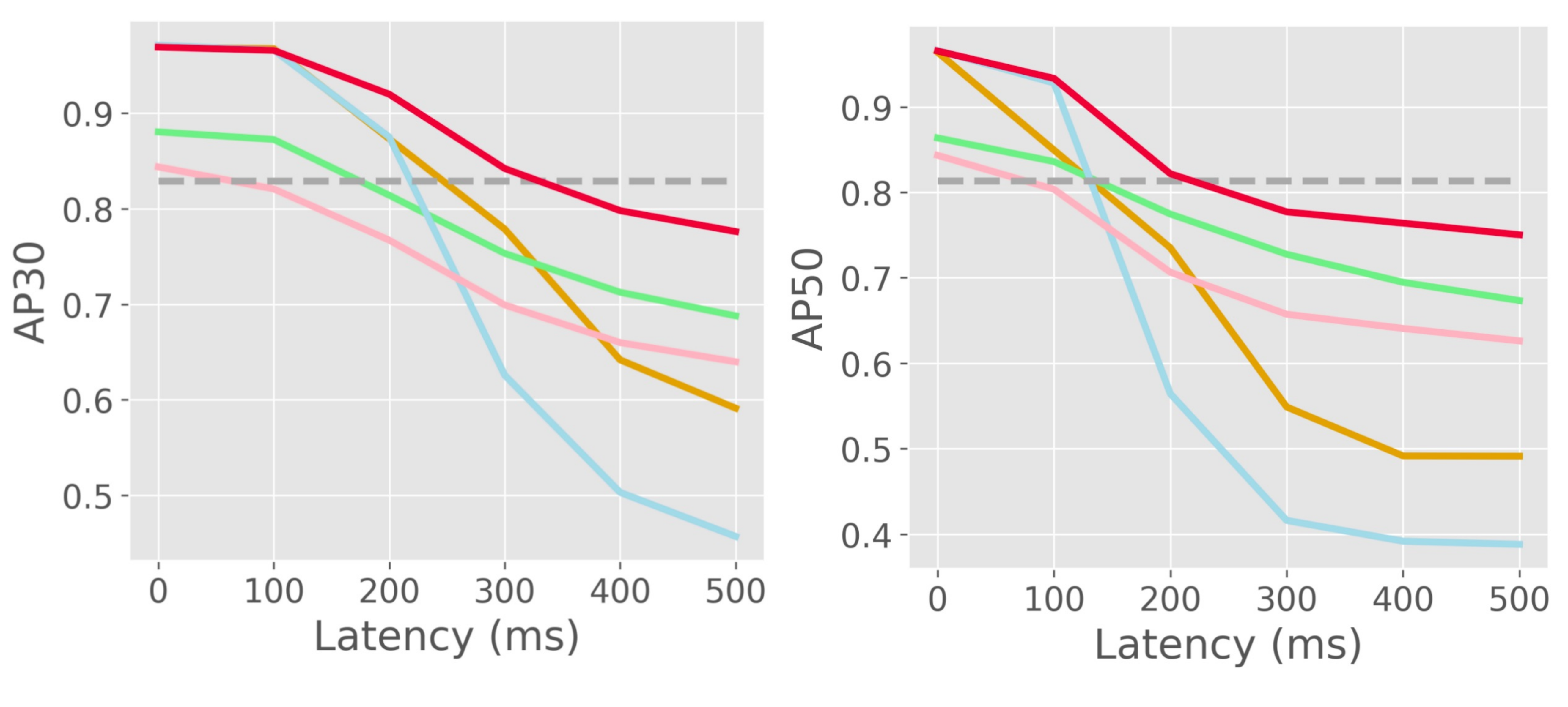}
    \vspace{-3mm}
  \caption{OPV2V}
  \label{Fig:OPV2V_latency}
  \end{subfigure}
\begin{subfigure}{0.487\linewidth}
    \includegraphics[width=0.99\linewidth]{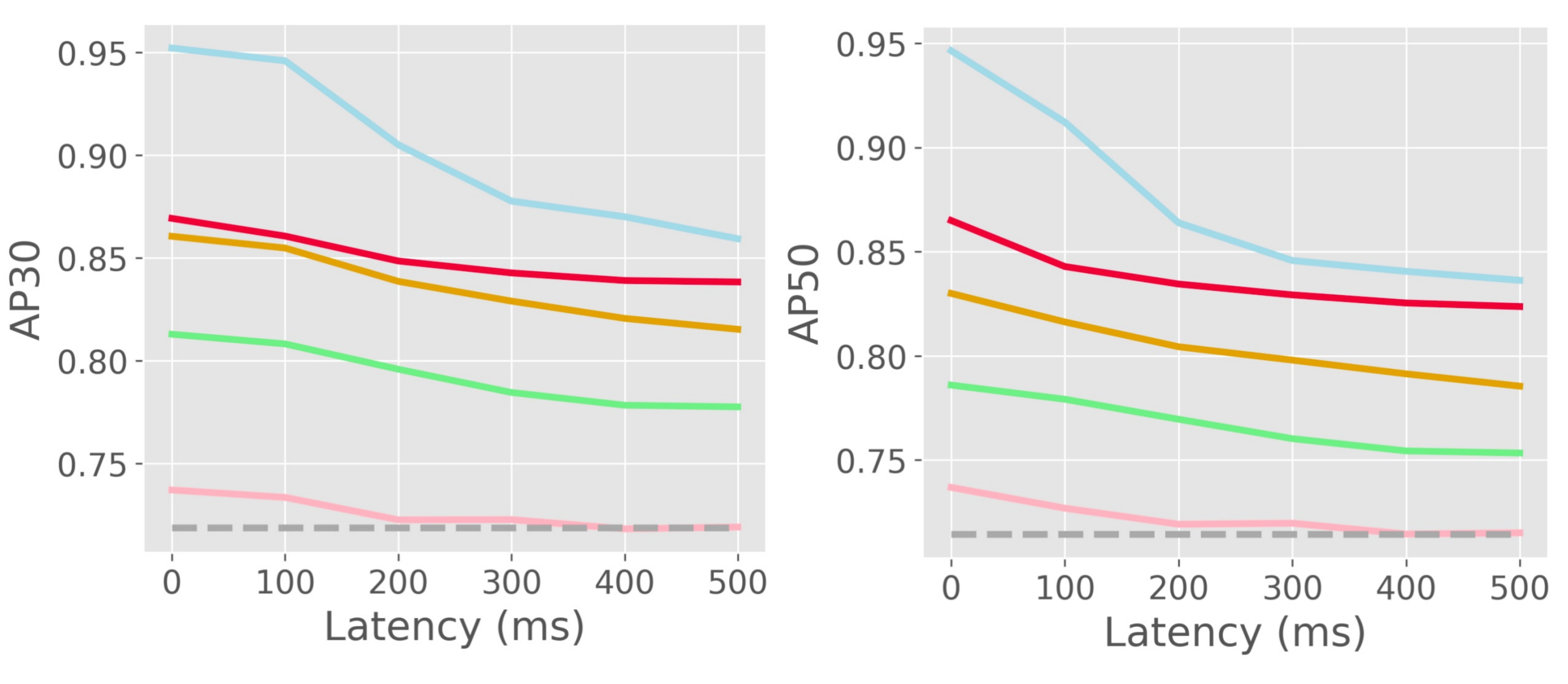}
    \vspace{-3mm}
  \caption{V2XSIM2.0}
  \label{Fig:V2XSIM_latency}
  \end{subfigure}
\vspace{-2mm}
\caption{The robustness to communication latency of CoDriving on commonly used collaborative perception real-world and simulation datasets.}
\vspace{-1mm}
\label{Fig:Latency}
\end{figure*}

\vspace{-3mm}
\subsection{Quantitative results}
\vspace{-1mm}
\label{subsec:perception_results}
\textbf{Perception performance and communication cost trade-off under both homogeneous and heterogeneous scenarios.}
Figure~\ref{Fig:SOTAs} and Figure~\ref{Fig:Hete} compare the proposed CoDriving with previous methods in terms of the trade-off between detection performance and communication bandwidth on real-world datasets DAIR-V2X, V2V4Real, TUMTraf-V2X and simulation datasets OPV2V, V2XSIM2.0 under homogeneous and heterogeneous settings, respectively. \textcolor{blue}{Baselines include no collaboration, 
V2VNet~\cite{WangV2vnet:ECCV20}, DiscoNet~\cite{LiLearning:NeurIPS21}, HMViT~\cite{XiangHMViT:ICCV23}, CoopDet3D~\cite{zimmer2024tumtraf}, early fusion, and late fusion, where agents exchange the detected 3D boxes directly.}
We see that CoDriving: i) achieves a far-more superior perception-communication trade-off across all the communication bandwidth choices on all the collaborative perception datasets under both homogeneous and heterogeneous scenarios; ii) significantly improves the detection performance, especially under extremely limited communication bandwidth, improving the SOTA performance by 47.58/9.66\% on DAIR-V2X and V2XSIM2.0 even when the bandwidth is constrained by a factor of 1K; \textcolor{blue}{ iii) outperforms previous SOTA, DiscoNet, with significantly reduced communication cost: 142/654 times less on V2V4Real and V2XSIM2.0, and achieves competitive performance compared to CoopDet3D on TUMTraf-V2X while reducing communication costs by 8 times; and iv) can adapt to varying communication bandwidth while previous methods are limited to specific communication choices.} The reason is that CoDriving can adjust the information selection under varying bandwidth limits by solving the constrained optimization problem \eqref{sec4:problem_reformulate_2}, while previous methods instinctively transmit the complete feature map.


\noindent
\textbf{Robustness of offline perception performance against pose error and communication latency issues.}
We validate the robustness of CoDriving against pose error and communication latency on V2V4Real, DAIR-V2X, OPV2V, and V2XSIM2.0. 
Figure~\ref{Fig:PE} and~\ref{Fig:Latency} show the detection performances as a function of pose error and latency, respectively. We see: i) while perception performance generally declines with increasing levels of pose error and latency, CoDriving achieves superior performance under all imperfect conditions across all the collaborative perception datasets. Note that early fusion initially outperforms CoDriving under minor pose error perturbations in DARI-V2X and V2XSIM2.0, but its performance declines quickly due to the limited ability to correct communication errors. ii) CoDriving consistently surpasses no collaboration on DAIR-V2X, V2V4Real, V2XSIM2.0 with metric AP30, whereas baselines fail when pose error exceeds 0.4m and latency exceeds 300ms.
\textcolor{blue}{The observed robustness is attributed to: i) the attention fusion helps to filter out the disturbed feature values by assigning lower attention weights; 2) CoDriving fuses features at three resolution level, as described in Section~\ref{subsec:collaboration}, which incorporates low-resolution feature fusion that is less sensitive to positional disturbances.}

\noindent
\textbf{Advantages.} Our advantages are twofold. First, compared to previous collaborative perception methods which only address a subset of evaluation scenarios, we conduct a thorough benchmarking of the existing state-of-the-art (SOTA) approaches under a wide range of evaluation settings, including four commonly used datasets, both homogeneous and heterogeneous sensor configurations, and two practical challenges: pose error and communication latency. Our comprehensive evaluation lowers the barriers to researching the advantages of collaboration facilitated by V2X communication from the perceptual perspective.
Second, this specialized variant of CoDriving consistently outperforms previous SOTAs in collaborative perception across these extensive evaluation scenarios. As a result, CoDriving serves as an effective method within the field of collaborative perception research.

\vspace{-3mm}
\section{\textcolor{blue}{System-level} driving evaluation}
\label{sec:Benchmark_driving}

V2Xverse facilitates the system-level evaluation of collaborative autonomous driving under unified data distribution. \textcolor{blue}{In this section, we evaluate the online closed-loop driving performance of CoDriving within V2Xverse simulation environment, and utilize the offline benchmark of V2Xverse to evaluate the modular performance. }

\begin{table*}
\renewcommand\arraystretch{0.1}
\centering
\scriptsize
\caption{Closed-loop driving performance evaluation. $\uparrow$ means the higher the better. $\downarrow$ means the lower the better. The \textbf{bold} font denotes the best performance. Expert is a powerful handcrafted agent, presented for reference and not included in the comparison. \textbf{Our CoDriving ensures the highest driving score with the second highest speed, and at the same time, achieves low collision rates, providing a reliable and efficient driving solution.}}
\vspace{-2mm}
\label{table:close-loop}
\resizebox{0.9\linewidth}{!}{%
\begin{tblr}{
  rowsep=0.3pt,
  colsep=2.5pt,
  row{1} = {c},
  row{2} = {c},
  row{4} = {c},
  row{5} = {c},
  row{6} = {c},
  row{7} = {c},
  row{9} = {c},
  row{10} = {c},
  row{11} = {c},
  row{12} = {c},
  row{13} = {c},
  row{14} = {c},
  row{15} = {c},
  row{16} = {c},
  cell{1}{1} = {c=9}{},
  cell{2}{1} = {c=2}{},
  cell{3}{1} = {r=5}{},
  cell{3}{2} = {c},
  cell{3}{3} = {c},
  cell{3}{4} = {c},
  cell{3}{5} = {c},
  cell{3}{6} = {c},
  cell{3}{7} = {c},
  cell{3}{8} = {c},
  cell{3}{9} = {c},
  cell{8}{1} = {r=7}{},
  cell{8}{2} = {c},
  cell{8}{3} = {c},
  cell{8}{4} = {c},
  cell{8}{5} = {c},
  cell{8}{6} = {c},
  cell{8}{7} = {c},
  cell{8}{8} = {c},
  cell{8}{9} = {c},
  cell{14}{2} = {c},
  cell{16}{1} = {c=2}{},
  vline{2} = {2,16}{},
  vline{2-3} = {2-16}{},
  vline{3} = {4-7,9-15}{},
  hline{1-3,17} = {-}{},
  hline{8} = {1-9}{},
  hline{16} = {1-9}{},
}
\textbf{Driving Performance}                            &             &                             &                                &                                &                                      &                                   &                                  &                          \\
\SetCell{c=2} Methods                                     &             & {\scriptsize Driving \\Score$\uparrow$} & {\scriptsize Route \\Completion$\uparrow$} & {\scriptsize Infraction \\Score$\uparrow$} & {\scriptsize Pedestrian \\Collision$\downarrow$} & {\scriptsize Vehicle \\Collision$\downarrow$} & {\scriptsize Layout \\Collision$\downarrow$} & {\scriptsize Mean \\Speed$\uparrow$} \\
\begin{sideways}Single\end{sideways}       & LAV \cite{LAV}        & 38.27                       & 55.07                          & 0.64                           & 2.83                                 & 3.97                              & 0.44                             & 2.13                     \\
& WOR  \cite{chen2021learning}       & 12.05                       & 22.27                          & 0.60                           & 3.13                                 & 5.60                              & 7.16                             & 1.30                     \\
& TCP \cite{TCP}         & 47.48                       & 62.14                          & 0.81                           & 1.57                                 & \textbf{0.39}                              & 0.20                              & 2.46                     \\
& TransFuser \cite{TransFuser} & 30.23                        & 73.47                           & 0.47                           & 5.12                                & 1.74                              & 0.26                            & 2.63                     \\
& InterFuser \cite{InterFuser} & 40.31                       & 85.07                          & 0.47                           & 2.63                                 & 2.68                              & 1.02                             & 2.55                     \\
\begin{sideways}Collaborative\end{sideways} & No Fusion & 33.81                       & 89.74                          & 0.37                           & 2.99                                 & 1.71                              & 1.87                             & 2.84                     \\
& Late Fusion      &52.40                        & 90.72                          & 0.57                           &2.07                                  &1.30                               & 1.75                             & 2.89                      \\
& Early Fusion      &59.12                        &90.72                           & 0.63                           &1.41                                  & 2.35                              & 1.76                             & 2.89                     \\
& Fcooper  \cite{ChenFcooper:SEC19}   & 44.00                       & 90.42                          & 0.46                           & 2.59                                 & 1.70                              & 1.94                             & 2.47                     \\
& V2X-ViT  \cite{xu2022v2xvit}    & 39.35                       & 91.98                          & 0.42                           & 2.61                                 & 4.05                              & 0.13                             & \textbf{3.43}                    \\
& \textcolor{blue}{CoopDet3D~\cite{zimmer2024tumtraf}}    & 63.04 & 88.12 & 0.71 & 1.67 & 0.83 & 1.09 & 2.76                    \\
& Coopernaut \cite{CuiCoopernaut:CVPR22} & 8.36                       & 12.38                       & 0.62                & 1.02                            & 3.45                      & 3.20                          & 0.92                 \\
& CoDriving (Ours)   & \textbf{77.15}                       & \textbf{92.34}                          & \textbf{0.82}                           & \textbf{0.73}                                 & 0.47                              & \textbf{0.04}                             & 3.05                     \\
Expert                                      &             & 82.58                       & 92.00                          & 0.89                           & 1.40                                 & 0.16                              & 0.00                          & 2.53                     
\end{tblr}
}
\end{table*}

\begin{table*}[tt]
\centering
\setlength\tabcolsep{4pt}
\renewcommand\arraystretch{1}
\caption{\small Modular tasks performance evaluation on different collaboration strategies \textcolor{blue}{in town5 test set}. On perception/planning, CoDriving outperforms other collaboration strategies.}
\vspace{-3mm}
\resizebox{0.9\linewidth}{!}{
\begin{tabular}{c|cccccccc|cc}
\toprule
\multirow{3}{*}{\begin{tabular}[c]{@{}c@{}}Collaborative \\ methods\end{tabular}} & \multicolumn{8}{c|}{\textbf{Detection}}                                                                                                                                                                                                                                                                    & \multicolumn{2}{c}{\multirow{2}{*}{\begin{tabular}[c]{@{}c@{}}\textbf{Waypoints}\\ \textbf{Planning}\end{tabular}}}  \\
& \multicolumn{3}{c|}{Vehicle}                                                                                  & \multicolumn{2}{c|}{Cyclist}                                             & \multicolumn{2}{c|}{Pedestrian}                                          & \multicolumn{1}{c|}{Mean}            & \multicolumn{2}{c}{}                                                                                                  \\
& \multicolumn{1}{c}{AP30$\uparrow$} & \multicolumn{1}{c}{AP50$\uparrow$} & \multicolumn{1}{c|}{AP70$\uparrow$} & \multicolumn{1}{c}{AP30$\uparrow$} & \multicolumn{1}{c|}{AP50$\uparrow$} & \multicolumn{1}{c}{AP30$\uparrow$} & \multicolumn{1}{c|}{AP50$\uparrow$} & \multicolumn{1}{c|}{mAP30$\uparrow$} & \multicolumn{1}{c}{ADE$\downarrow$} & \multicolumn{1}{c}{FDE$\downarrow$}                                             \\ 
\midrule
No Fusion                                                                         & 0.89                               & 0.84                               & \multicolumn{1}{c|}{0.73}                                & 0.40                               & \multicolumn{1}{c|}{0.30}                                & 0.41                               & \multicolumn{1}{c|}{0.24}                                & 0.57                                 & 0.636                               & 1.460                                                                           \\
Late Fusion                                                                       & 0.88                               & 0.86                               & \multicolumn{1}{c|}{0.81}                                & 0.43                               & \multicolumn{1}{c|}{0.38}                                & 0.45                               & \multicolumn{1}{c|}{0.27}                                & 0.59                                 & 0.631                               & 1.454                                                                           \\
Fcooper \cite{ChenFcooper:SEC19}                                                                          & 0.93                               & 0.82                               & \multicolumn{1}{c|}{0.68}                                & 0.44                               & \multicolumn{1}{c|}{0.29}                                & 0.56                               & \multicolumn{1}{c|}{0.33}                                & 0.64                                 & 0.627                               & 1.446                                                                           \\
V2X-ViT \cite{xu2022v2xvit}                                                                          & 0.93                               & \textbf{0.91}                               & \multicolumn{1}{c|}{\textbf{0.84}}                                & 0.50                               & \multicolumn{1}{c|}{0.36}                                & 0.41                               & \multicolumn{1}{c|}{0.12}                                & 0.61                                 & 0.629                               & 1.447                                                                           \\
\textcolor{blue}{CoopDet3D~\cite{zimmer2024tumtraf}}                                                                      & 0.93                               & 0.90                               & \multicolumn{1}{c|}{0.81}                                & 0.48                               & \multicolumn{1}{c|}{\textbf{0.41}}                                & 0.53                               & \multicolumn{1}{c|}{0.31}                                & 0.65                                 & 0.623                               & 1.439                                                                           \\
CoDriving (Ours)                                                                  & \multicolumn{1}{c}{\textbf{0.94}}           & \textbf{0.91}                               & \multicolumn{1}{c|}{0.83}                                & \textbf{0.52}                               & \multicolumn{1}{c|}{\textbf{0.41}}                                & \textbf{0.58}                               & \multicolumn{1}{c|}{\textbf{0.35}}                                & \textbf{0.68}                                 & \textbf{0.619}                               & \textbf{1.413}                                                                           \\
\bottomrule                       
\end{tabular}
}
\label{table:modular}
\end{table*}

\vspace{-2.5mm}
\subsection{Experimental setting and details}

Our experiments comprehensively evaluate CoDriving, including three tasks: the online driving task, the perception (3D detection) task, and the waypoints planning task. 
Two types of input modality are covered in the evaluation, including 3D point cloud and RGB image. We also evaluate the trade-off between performance and communication band-width, and the robustness against two types of practical issues, including communication latency and pose error.

\noindent
\textbf{Metrics.} We evaluate the closed-loop driving performance with a bunch of driving safety and efficiency metrics introduced in Section~\ref{sec:evaluation}, including driving score (DS), route completion ratio (RC), infraction score (IS), pedestrian/vehicle/layout collision rate, and mean speed; see details in Appendix \ref{sec:appendix data descrip}. The 3D detection task is evaluated using the Average Precision (AP) metric for different categories of objects at various IoU thresholds, along with the mean Average Precision (mAP) computed across all categories. The waypoints planning task is evaluated with average displacement error (ADE) and final displacement error (FDE).

\noindent
\textbf{Implementation details.}
In closed-loop evaluation, the ego vehicle can access and communicate with one RSU simultaneously. In the perception and the planning tasks, the ego vehicle can communicate with at most one RSU and one vehicle. Both vehicles and roadside units (RSU) are equipped with cameras and LiDAR. The image resolution is 600 $\times$ 800 and the LiDAR is 64-channel. The detection range is 48m $\times$ 24m, with 36m/12m/12m/12m at front/rear/left/right.

\noindent
\textbf{Baselines.}
We compare our method with representative collaborative methods and single-agent end-to-end driving methods. First, collaborative baselines are categorized into two types: i) collaborative end-to-end driving method, Coopernaut~\cite{CuiCoopernaut:CVPR22}, and ii) collaborative perception methods, V2X-ViT~\cite{xu2022v2xvit}, Fcooper~\cite{ChenFcooper:SEC19}, \textcolor{blue}{ CoopDet3D~\cite{zimmer2024tumtraf}}, perceptual-output-based late collaboration (Late fusion) and raw-inputs-based early collaboration (Early fusion). Note that, to assess the system-level driving performance of each collaborative perception modular method, we integrate their perception components into our driving system, enabling them with the planner and controller of CoDriving, and train the planning modules accordingly. Specifically, we re-implement these collaborative perception methods with PointPillar BEV encoding~\cite{Lang2018PointPillarsFE} and centerpoint detection loss \cite{zhou2019objects}, as done in CoDriving. Second, for individual-agent end-to-end driving baselines, we identify two groups: 1) state-of-the-art methods, including LAV \cite{LAV}, WOR \cite{chen2021learning}, TCP \cite{TCP}, Transfuser \cite{TransFuser}, InterFuser \cite{InterFuser} and 2) a handcrafted expert autopilot system which has access to the privileged information in the CARLA simulator.
















\begin{table*}[t]
\centering
\small
\renewcommand\arraystretch{1}
\setlength\tabcolsep{3pt}
\caption{ Ablation study for multi-modality (C: camera/L: LiDAR) and collaboration ($\times/\checkmark$) in driving performance. Collaboration improves the driving score by 98\%(LiDAR), 18\%(camera).}
\vspace{-2mm}
\resizebox{0.9\linewidth}{!}{%
\begin{tabular}{c|cc|ccccccccc}
\toprule
\multirow{6}{*}{\rotatebox{90}{\textbf{Driving\quad}}} & \multirow{2}{*}{\makecell{Collabo-\\ration}} & \multirow{2}{*}{Modality} & \multirow{2}{*}{\makecell{Driving \\Score$\uparrow$}} & \multirow{2}{*}{\makecell{Route \\Completion$\uparrow$}} & \multirow{2}{*}{\makecell{Infraction \\Score$\uparrow$}} & \multirow{2}{*}{\makecell{Pedestrian \\Collision$\downarrow$}} & \multirow{2}{*}{\makecell{Vehicle \\Collision$\downarrow$}} & \multirow{2}{*}{\makecell{Layout \\Collision$\downarrow$}}  & \multirow{2}{*}{\makecell{Mean \\ Speed$\uparrow$}} \\
&& \\ \cmidrule{2-10}
& \multirow{2}{*}{$\times$} & C  &  38.88     &  92.13   & 0.40  & 4.21 & 3.51 & 1.23 & \textbf{3.12} \\
&  & L   & 33.81                       & 89.74                          & 0.37                           & 2.99                                 & 1.71                              & 1.87                             & 2.84     \\\cmidrule{2-10}
& \multirow{2}{*}{$\checkmark$} & C  & 45.06  & \textbf{95.28} & 0.46 & 2.29 & 2.31 & 0.23 & 2.83 \\
& & L   &  \textbf{77.15}                       & 92.34                         & \textbf{0.82}                           & \textbf{0.73}                                 & \textbf{0.47}                              & \textbf{0.04}                             & 3.05 \\\bottomrule
\end{tabular}}
\vspace{-3mm}
\label{table:modality1}
\end{table*}

\begin{table*}[t]
\centering
\renewcommand\arraystretch{1}
\setlength\tabcolsep{3pt}
\caption{\small 
Ablation study for multi-modality (C: camera/L: LiDAR) and collaboration ($\times/\checkmark$) on perception and planning. Collaboration improves mAP30 by 19.30\% (LiDAR), 37.78\% (camera) and FDE by 3.22\% (LiDAR), 10.19\% (camera).}
\vspace{-3mm}
\resizebox{0.9\linewidth}{!}{
\begin{tabular}{cccccccccccc}
\toprule
                                            &  \multicolumn{1}{c|}{}      & \multicolumn{8}{c|}{\textbf{Detection}}                                                                                                                                                                                                       & \multicolumn{2}{c}{\multirow{2}{*}{\begin{tabular}[c]{@{}c@{}}\textbf{Waypoints}\\ \textbf{Planning}\end{tabular}}} \\
\multicolumn{1}{c}{\multirow{2}{*}{\makecell{Collabo-\\ration}}} & \multicolumn{1}{c|}{\multirow{2}{*}{Modality}} & \multicolumn{3}{c|}{Vehicle}                                                    & \multicolumn{2}{c|}{Cyclist}                         & \multicolumn{2}{c|}{Pedestrian}                               & \multicolumn{1}{c|}{Mean}   & \multicolumn{2}{c}{}                                                                               \\
\multicolumn{1}{c}{}                        & \multicolumn{1}{c|}{}                          & \multicolumn{1}{c}{AP30$\uparrow$} & \multicolumn{1}{c}{AP50$\uparrow$} & \multicolumn{1}{c|}{AP70$\uparrow$} & \multicolumn{1}{c}{AP30$\uparrow$} & \multicolumn{1}{c|}{AP50$\uparrow$} & \multicolumn{1}{c}{AP30$\uparrow$} &   \multicolumn{1}{c|}{AP50$\uparrow$} & \multicolumn{1}{c|}{mAP30$\uparrow$} & ADE$\downarrow$                                              & FDE$\downarrow$                                             \\ \midrule
\multirow{2}{*}{$\times$}                     & \multicolumn{1}{c|}{C}                         & 0.73  & 0.63 & 0.51    & 0.36       & 0.22    &   0.27     & 0.14   &  0.45    & 0.706 & 1.531 \\
                                            & \multicolumn{1}{c|}{L}                          & 0.89               & 0.84           & 0.73           & 0.40           & 0.30           & 0.41           & 0.24           & 0.57           & 0.636                                     & 1.460 \\ \midrule
\multirow{2}{*}{$\checkmark$}                   & \multicolumn{1}{c|}{C}                   & 0.91 & 0.86 &   0.76  &     \textbf{0.52}          &  0.37   &  0.43  & 0.28   &   0.62   & 0.634  & \textbf{1.375} \\
                                            & \multicolumn{1}{c|}{L}                         & \textbf{0.94}               & \textbf{0.91}           & \textbf{0.83}           & \textbf{0.52}           & \textbf{0.41}           & \textbf{0.58}           & \textbf{0.35}           & \textbf{0.68}            & \textbf{0.619}                                     & 1.413  \\ \bottomrule
\end{tabular}}
\label{table:modality2}
\end{table*}

\begin{figure*}[!ht]
  \vspace{-3mm}
  \centering
  \begin{subfigure}{0.32\linewidth}
    \includegraphics[width=1.0\linewidth,height=1\linewidth]{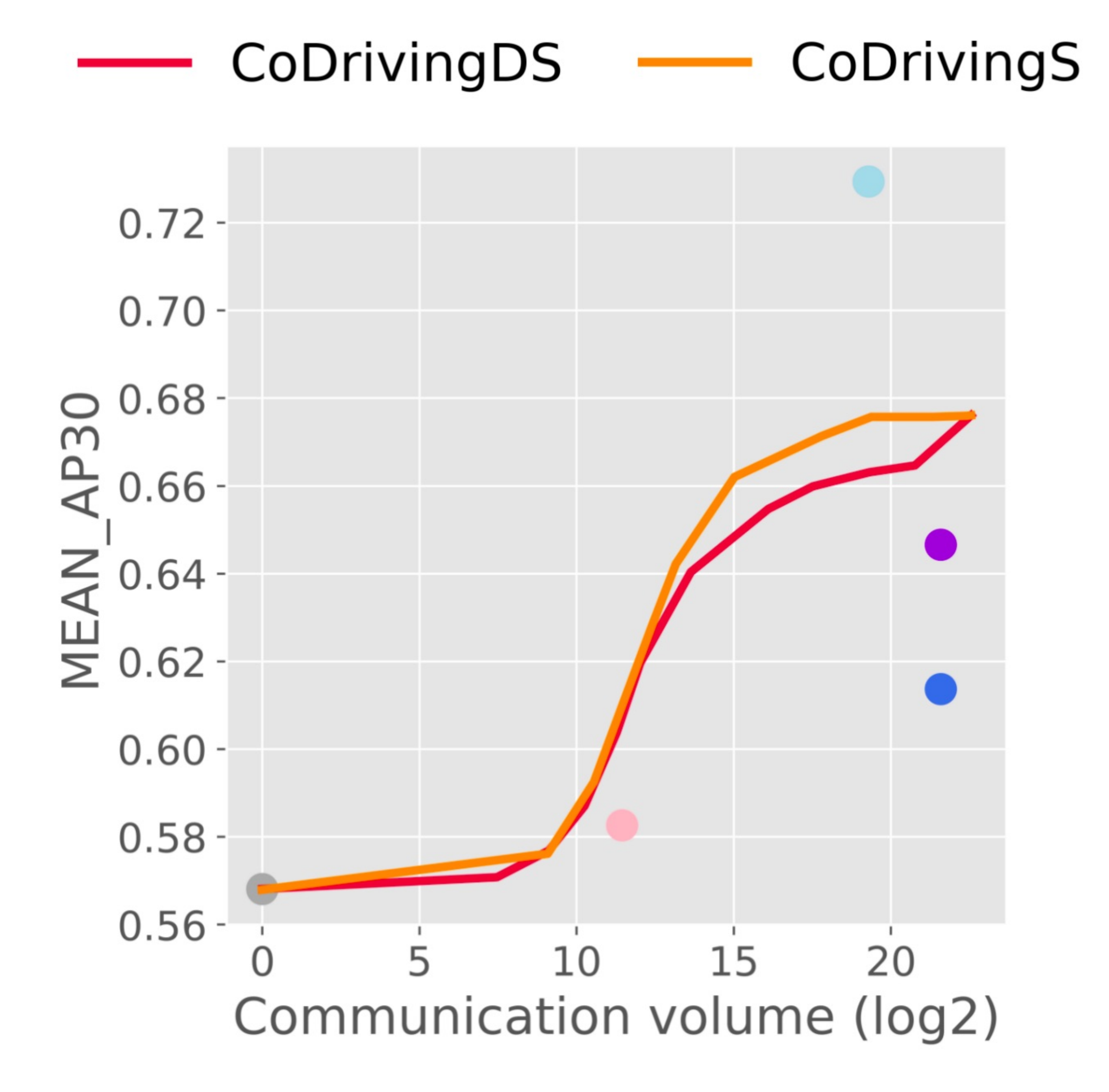}
    \vspace{-6mm}
    \caption{Perception (mAP30$\uparrow$)}
    \label{fig:comm_det}
  \end{subfigure}
  \begin{subfigure}{0.32\linewidth}
    \includegraphics[width=1.0\linewidth,height=1\linewidth]{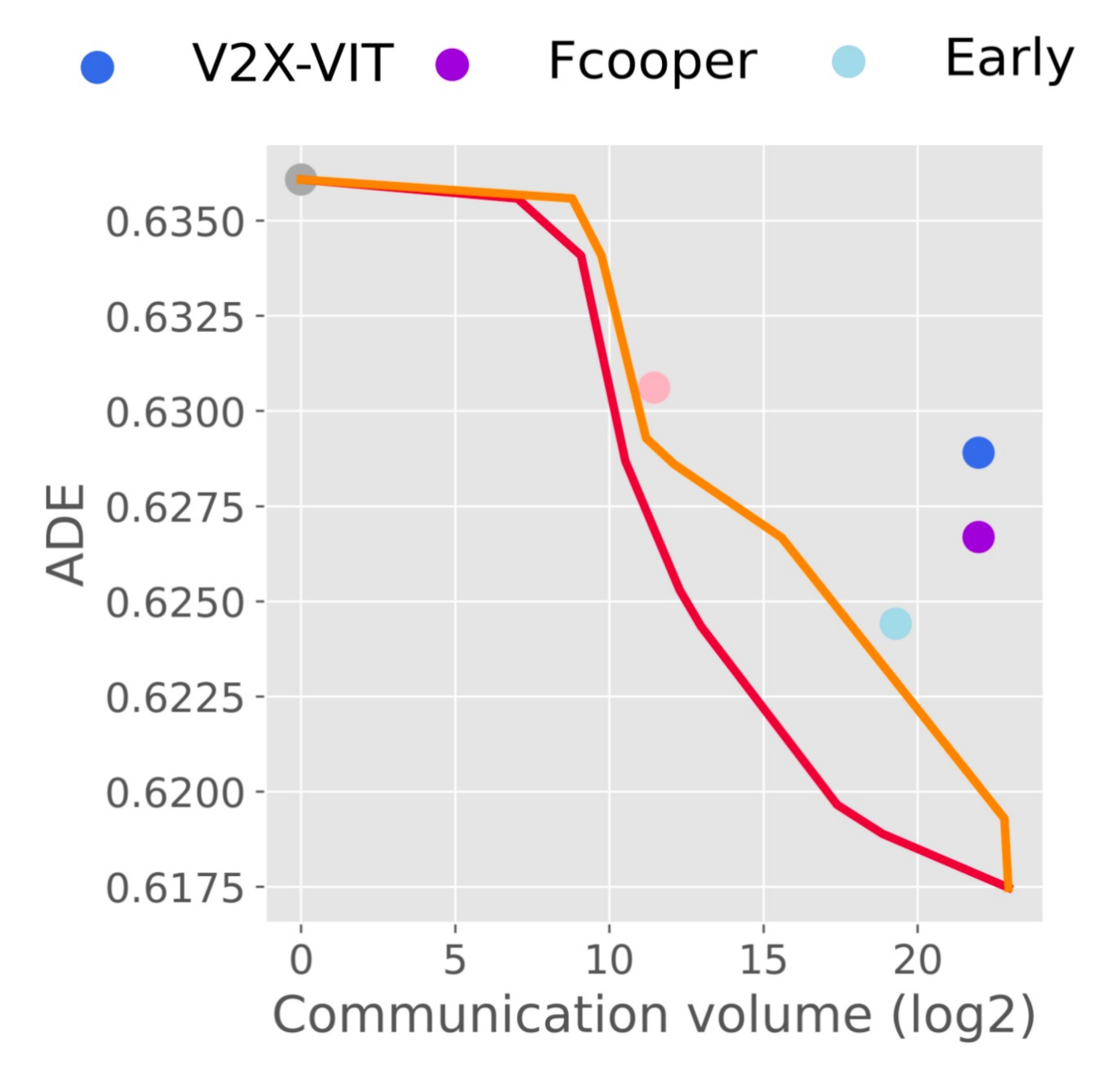}
    \vspace{-6mm}
    \caption{Planning (ADE$\downarrow$)}
  \end{subfigure}
  \begin{subfigure}{0.32\linewidth}
    \raisebox{0.00cm}{\includegraphics[width=1.0\linewidth,height=1\linewidth]{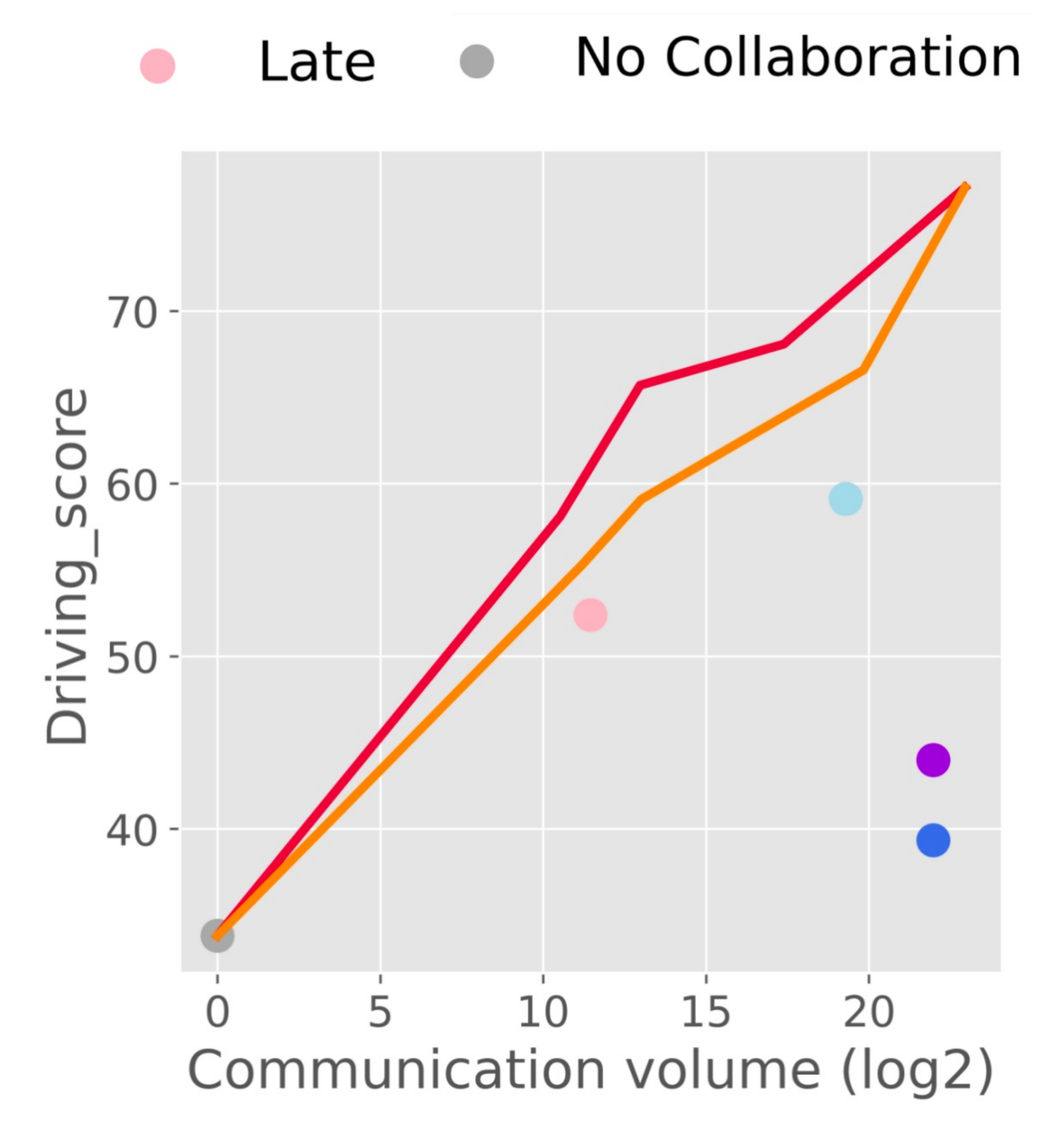}}
    \vspace{-6mm}
    \caption{Driving (driving score$\uparrow$)}
  \end{subfigure}
  \vspace{-2mm}
  \caption{The trade-off between performance and communication bandwidth. CoDrivingDS/S denotes CoDriving with/without driving-request. Compared with previous collaboration strategies, CoDriving with driving-request-aware communication consistently achieves superior performance over varying bandwidths in perception, planning and driving task.}
  \vspace{-3mm}
  \label{Fig:comm-bandwidth}
\end{figure*}

\begin{figure*}[!ht]
  \centering
  \vspace{-1mm}
  \begin{subfigure}{0.32\linewidth}
    \includegraphics[width=0.95\linewidth,height=0.95\linewidth]{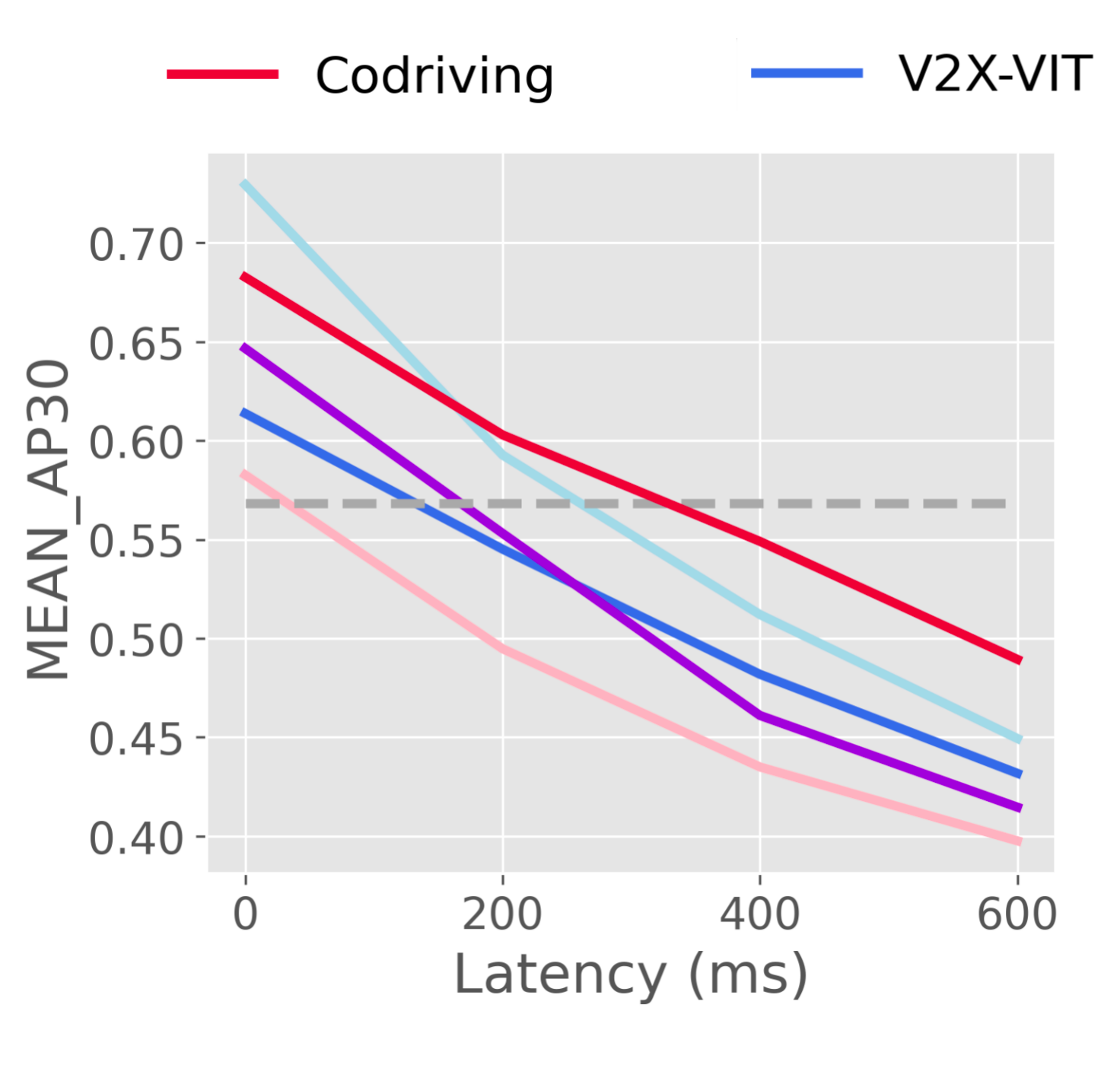}
    \vspace{-3mm}
    \caption{Perception (mAP30$\uparrow$)}
  \end{subfigure}
  \begin{subfigure}{0.32\linewidth}
    \includegraphics[width=0.95\linewidth,height=0.95\linewidth]{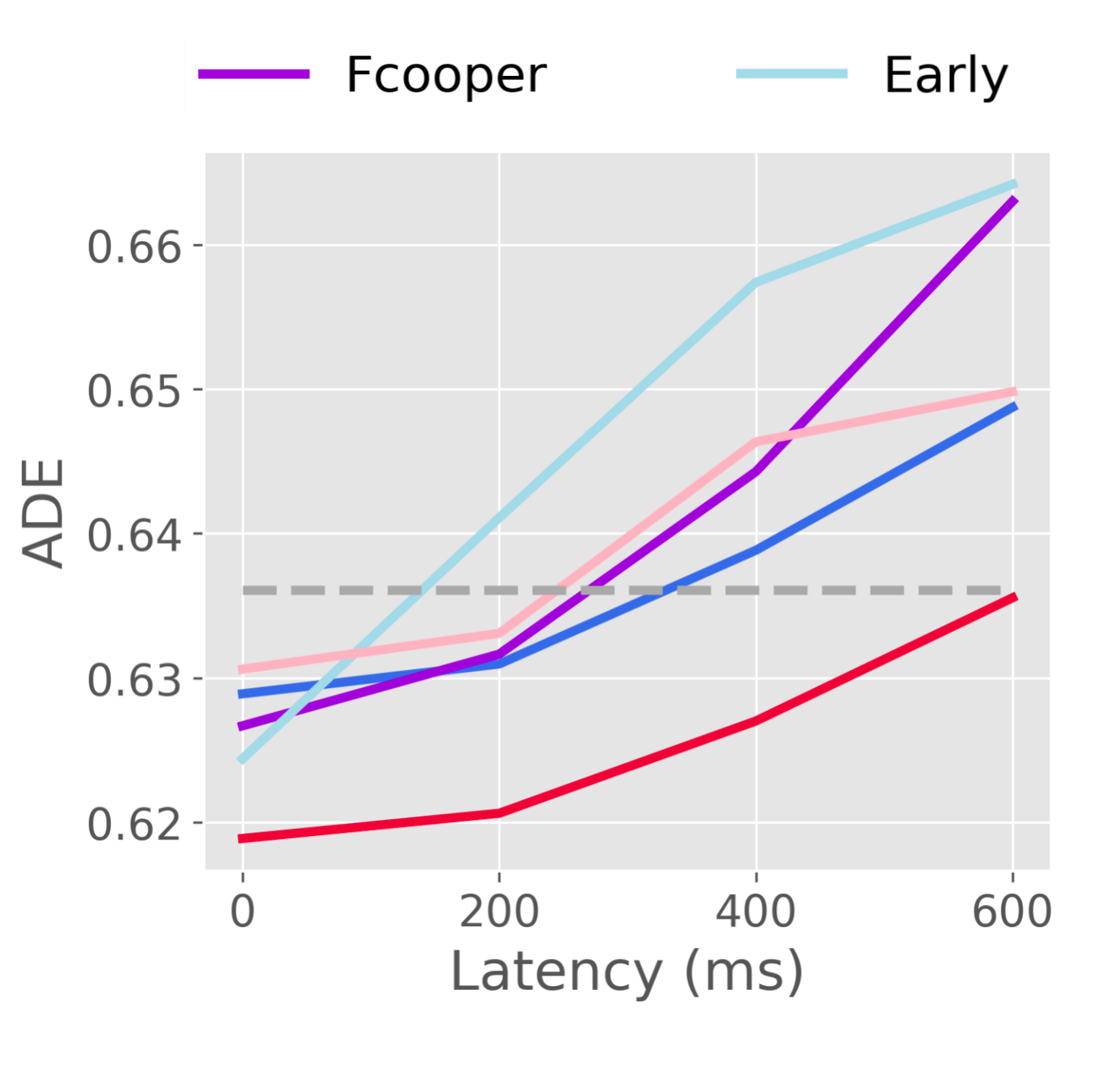}
    \vspace{-3mm}
    \caption{Planning (ADE$\downarrow$)}
  \end{subfigure}
  \begin{subfigure}{0.32\linewidth}
    \includegraphics[width=0.953\linewidth,height=0.953\linewidth]{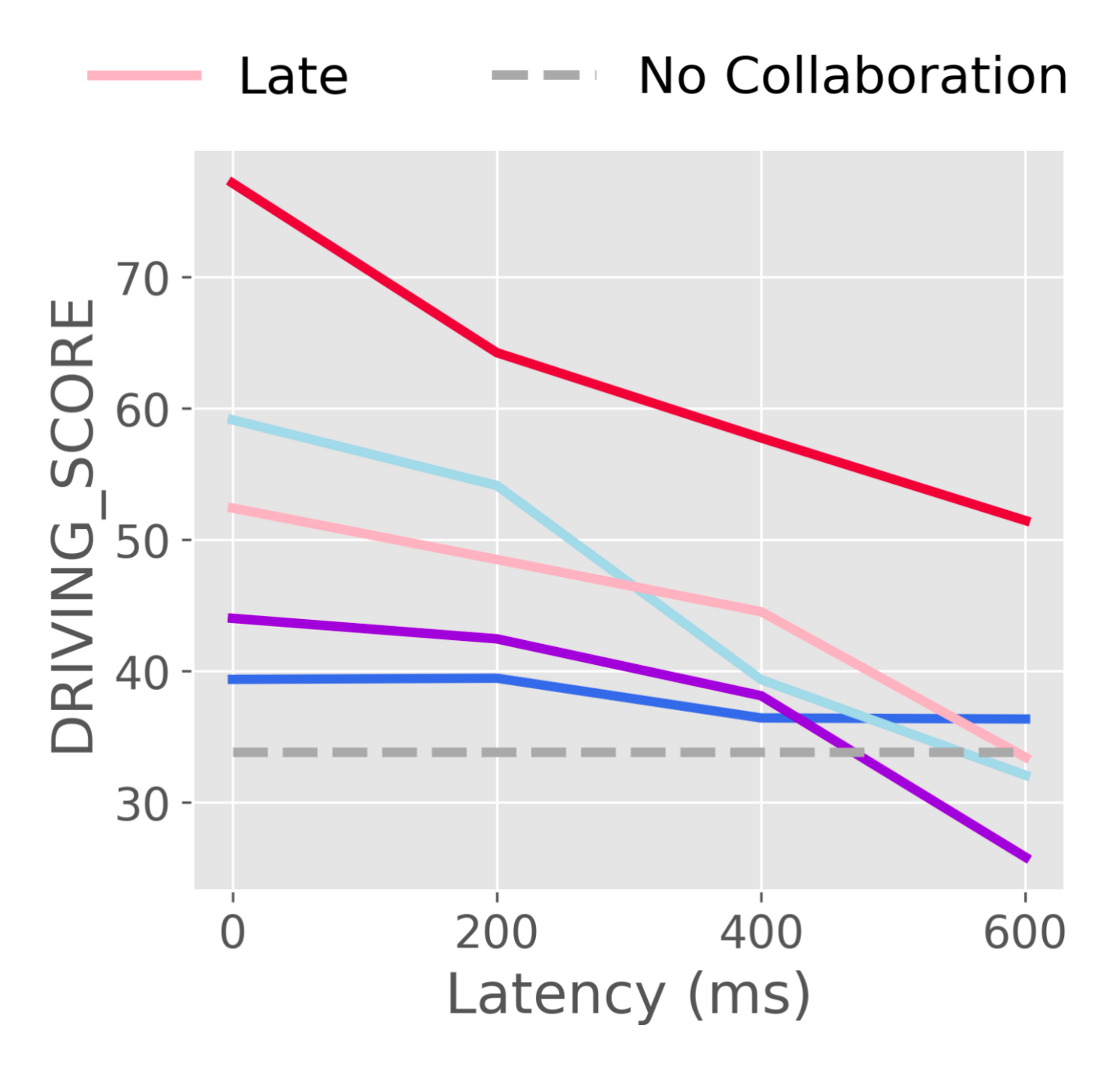}
    \vspace{-3mm}
    \caption{Driving (driving score$\uparrow$)}
  \end{subfigure}
  \vspace{-3mm}
  \caption{Robustness to latency on V2Xverse. CoDriving consistently outperforms previous collaboration strategies.}
  \vspace{-4mm}
  \label{Fig:Latency_performance}
\end{figure*}

\begin{figure*}[!ht]
  \centering
  \vspace{-1mm}
  \begin{subfigure}{0.32\linewidth}
    \includegraphics[width=0.95\linewidth,height=0.95\linewidth]{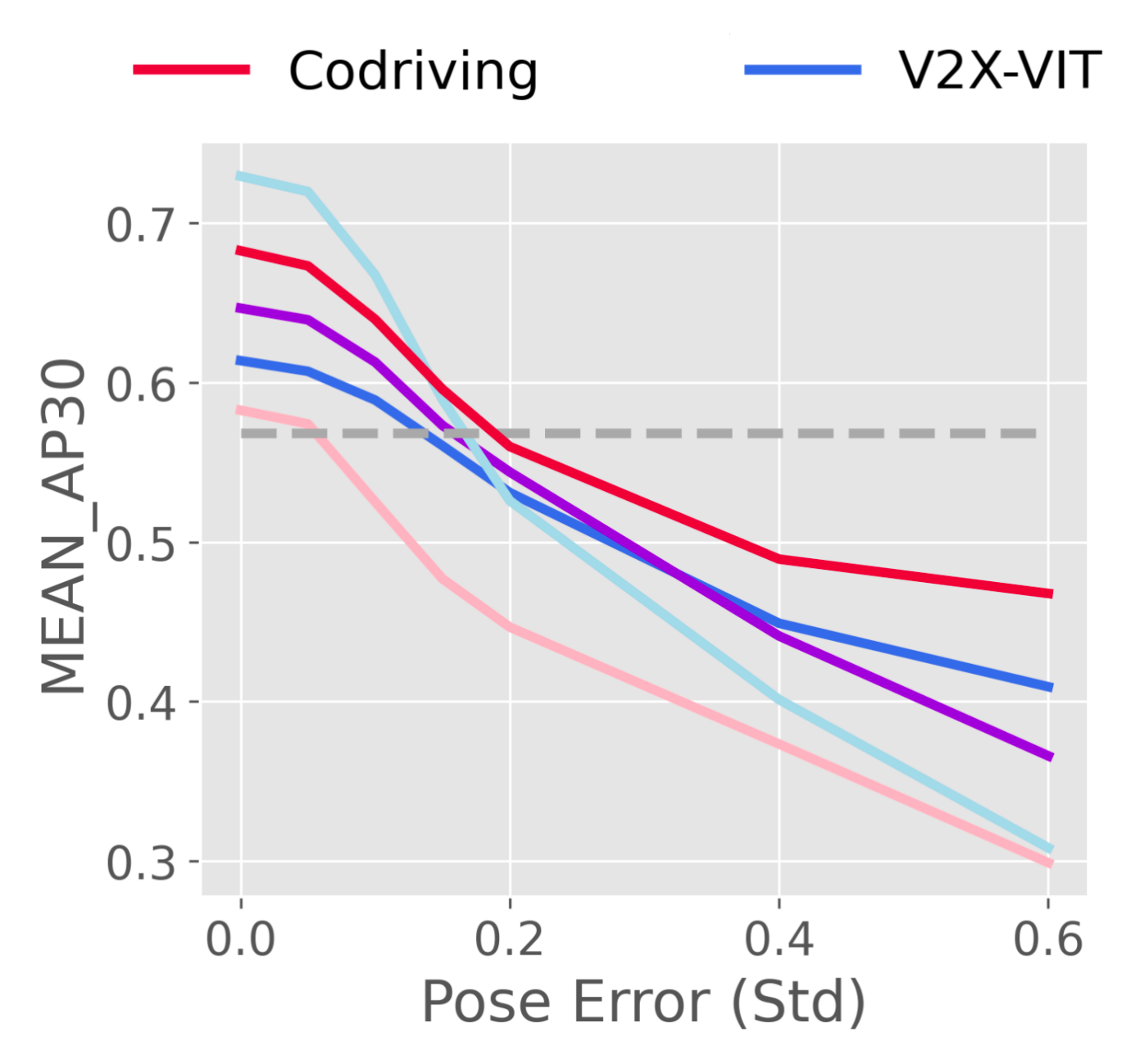}
    \vspace{-3mm}
    \caption{Perception (mAP30$\uparrow$)}
  \end{subfigure}
  \begin{subfigure}{0.32\linewidth}
    \includegraphics[width=0.95\linewidth,height=0.95\linewidth]{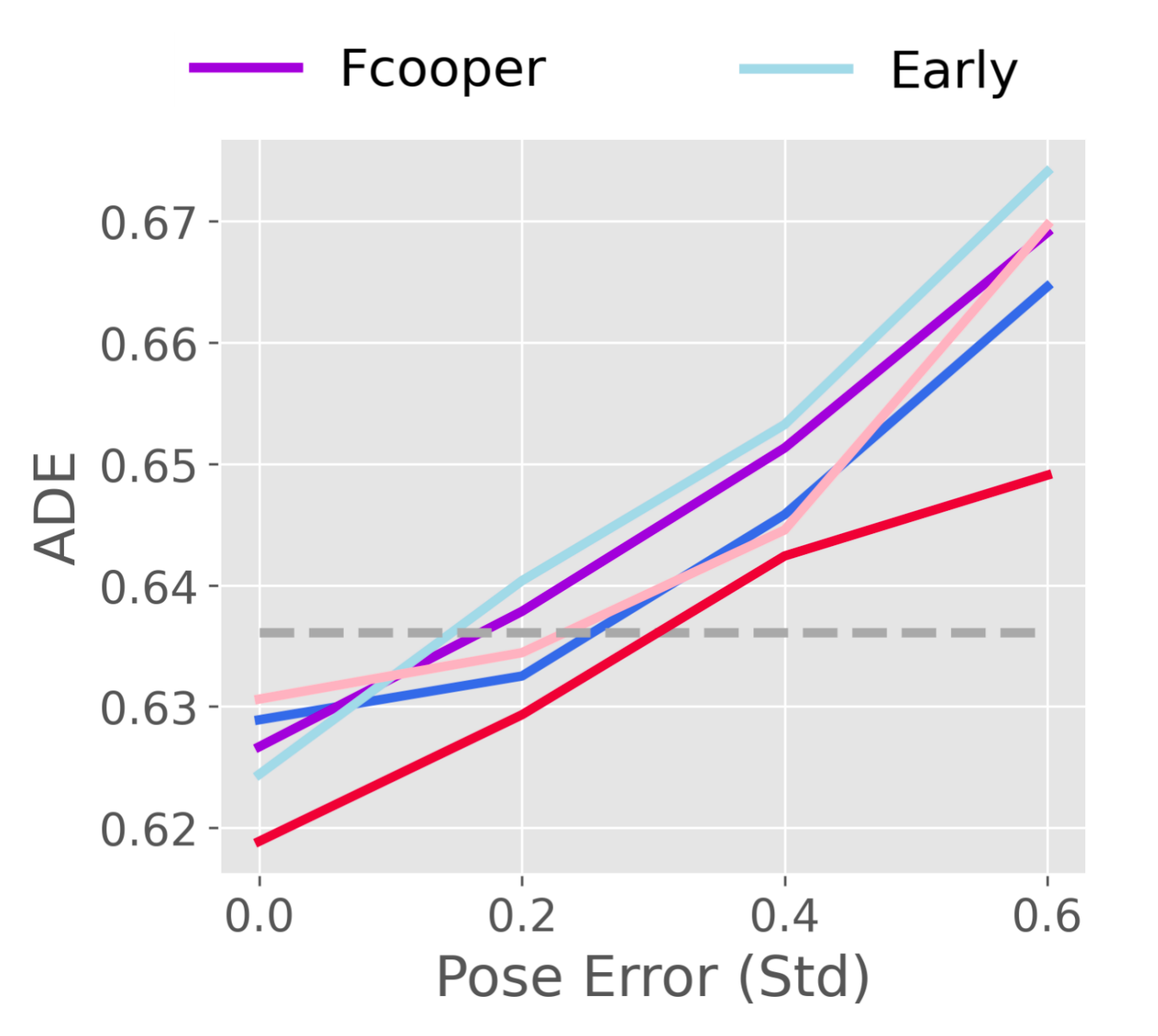}
    \vspace{-3mm}
    \caption{Planning (ADE$\downarrow$)}
  \end{subfigure}
  \begin{subfigure}{0.32\linewidth}
    \includegraphics[width=0.95\linewidth,height=0.95\linewidth]{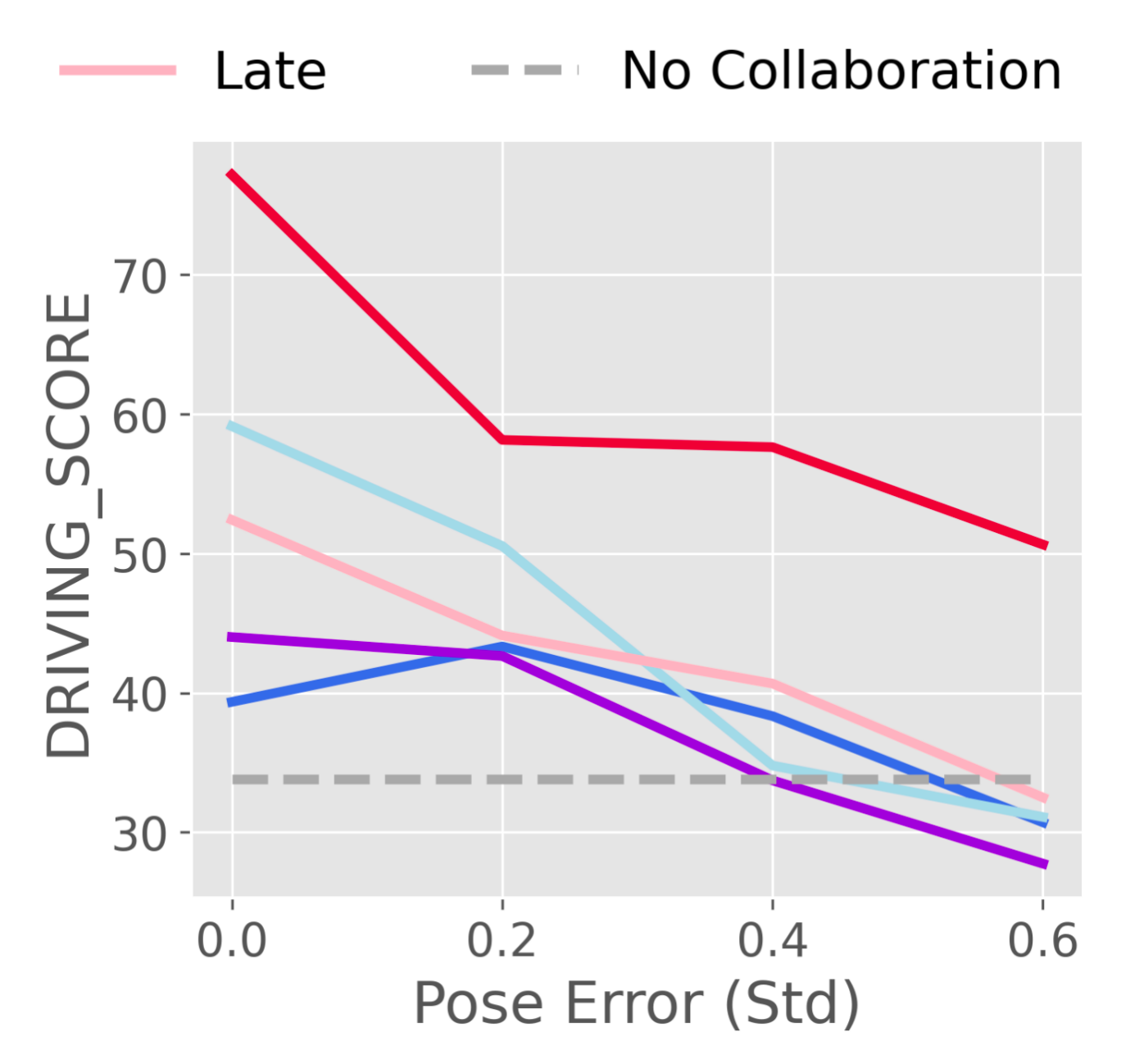}
    \vspace{-3mm}
    \caption{Driving (driving score$\uparrow$)}
  \end{subfigure}
  \vspace{-3mm}
  \caption{Robustness to pose error on V2Xverse. CoDriving consistently outperforms previous collaboration strategies.
    }
  \vspace{-4mm}
  \label{Fig:pose_error_performance}
\end{figure*}

\vspace{-2mm}
\subsection{Quantitative results}
\vspace{-2mm}
\textbf{Closed-loop driving performance evaluation.} 
Table \ref{table:close-loop} presents the comparison of our CoDriving with representative state-of-the-art methods  in terms of closed-loop driving performance. We see that i) our CoDriving achieves the highest driving score
and the highest infraction score while maintaining the second highest speed,
providing a safe and efficient driving solution; ii) compared to Coopernaut, the existing end-to-end collaborative driving method, our CoDriving achieves significantly more effective driving behavior
with a higher mean speed by \textbf{231.52\%} improvement and a higher driving score by \textbf{822.85\%} improvement. In contrast, Coopernaut experiences the lowest route completion due to its overfitted policy; iii) compared with other collaborative perception approaches like Fcooper, V2X-Vit, and CoopDet3D, our method achieves the lowest pedestrian and vehicle collision rate, showing the effectiveness of our collaborative perception method on pedestrian and vehicle detection;
iv) compared with TCP, the state-of-the-art end-to-end individual driving framework, CoDriving achieves a more efficient driving behavior with a higher mean speed by \textbf{23.98\%} improvement and a higher route completion by \textbf{48.60\%} improvement. Meanwhile, CoDriving ensures safety with a significantly lower pedestrian collision rate by \textbf{53.50\%}. TCP takes a relatively conservative driving strategy, resulting in a low route completion rate due to exceeding the time limit. 


\noindent
\textbf{Modular tasks evaluation on different collaboration strategies.}
Table \ref{table:modular} compares CoDriving with other collaboration strategies on modular tasks including 3D object detection and waypoints planning. We see that: i) all the collaboration strategies outperform no collaboration on both perception and planning task; ii) our CoDriving achieves the best detection and planning performance.

\begin{figure*}[!t]
  \centering
  \begin{subfigure}{0.48\linewidth}
    \includegraphics[width=0.98\linewidth]{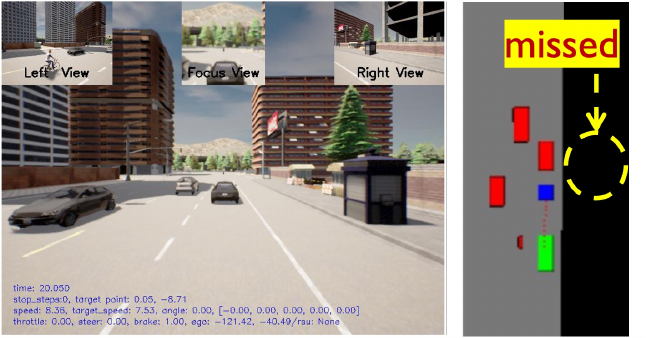}
    \caption{No collaboration at $T_0$}
    \label{fig:V-PnP}
  \end{subfigure}
  \begin{subfigure}{0.48\linewidth}
    \includegraphics[width=0.96\linewidth]{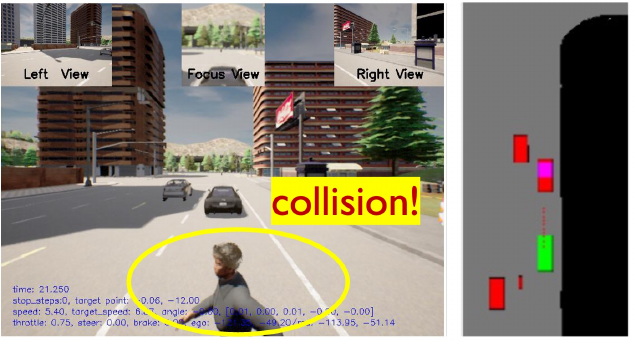}
    \caption{No collaboration at $T_{1}$}
    \label{fig:V-Prediction}
  \end{subfigure} \\
  \begin{subfigure}{0.48\linewidth}
    \includegraphics[width=0.98\linewidth,trim=-1.0mm 0 2mm -2mm,]{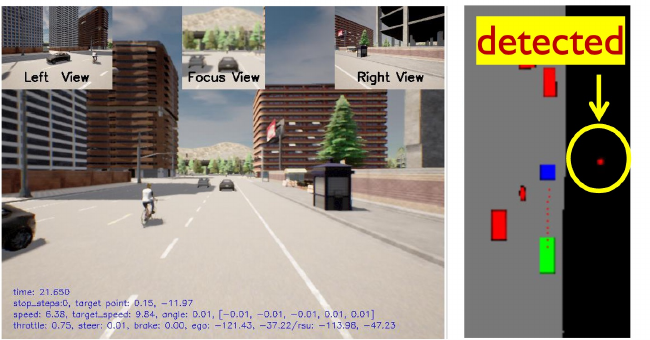}
    \caption{CoDriving at $T_0$}
    \label{fig:V2X-PnP}
  \end{subfigure}
  \begin{subfigure}{0.48\linewidth}
    \includegraphics[width=0.96\linewidth,trim=-1.7mm 5mm 0 -3mm,]{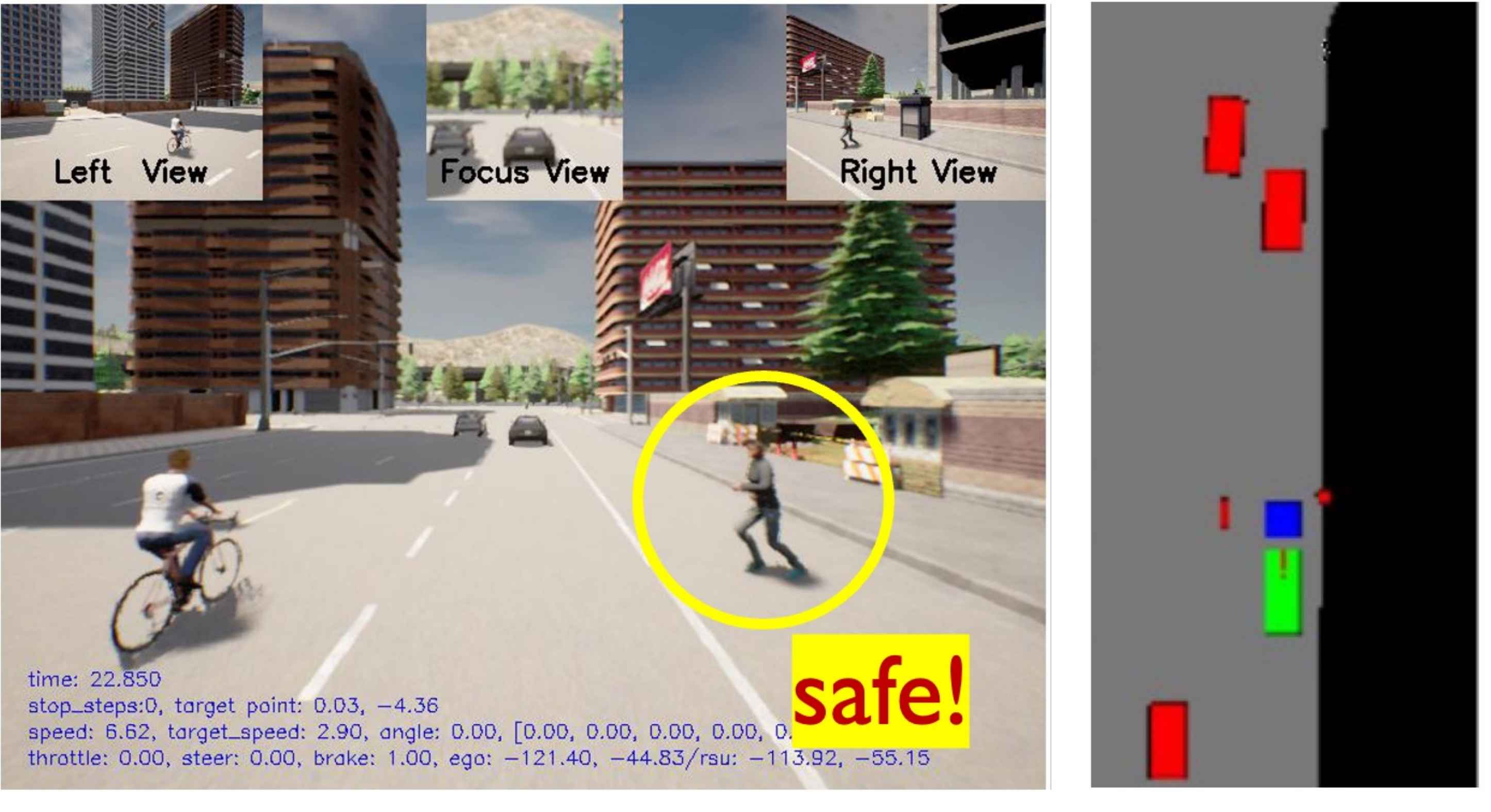}
    \caption{CoDriving at $T_{1}$}
    \label{fig:V2X-Prediction}
  \end{subfigure}
  \vspace{-2mm}
  \caption{\small Visualization of a potentially dangerous scenario and detection results. Compared to no collaboration, our CoDriving perceives the actors better with the complement of RSU information and avoids collision.}
  \label{fig:DrivingSample}
  \vspace{-5mm}
\end{figure*}

\noindent
\textbf{Driving and modular tasks evaluation on different modalities.}
Table \ref{table:modality1} and Table \ref{table:modality2} present
the comparison of no collaboration and Codriving collaboration under different input modalities: Camera-based (C) and LiDAR-based (L). We evaluate the whole system and report the performance of perception, planning and driving. From Table \ref{table:modality1} and Table \ref{table:modality2}, we see that i) for both modalities, the communication-enabled multi-agent collaboration significantly enhances the driving system, validating the generalizability of collaboration; and ii) collaboration benefits the entire driving system, improving the driving score by 98.43\%, the detection mAP30 by 19.30\%, and reducing the planning FDE by 3.22\% based on LiDAR inputs.

\noindent
\textbf{Trade-off between performance and communication bandwidth.}
\textcolor{blue}{To compare communication results straightforwardly and fairly under different communication bandwidths, we do not consider any extra data/feature/model compression. Figure~\ref{Fig:comm-bandwidth} compares CoDriving with/without (red/orange curve) the proposed driving-request-aware communication with other collaboration strategies, illustrating the trade-off between performance and communication bandwidth in the driving task and modular tasks.
The absence of driving request map indicates that the communication is determined solely by the confidence map. We see that:
i) CoDriving outperforms other intermediate collaboration strategies across all communication bandwidth on all three tasks, reducing the communication cost 90/1176 times less on perception/planning than Fcooper and 630/4389 times less than V2X-VIT.
ii) Compared to perceptual-confidence-only communication, the proposed driving-request-aware communication achieves a superior trade-off between performance and communication cost in both planning and driving tasks. 
In the ultimate closed-loop driving task, the driving-request mechanism achieves an average improvement of $9.6\%$ in driving score across communication volume ranging from $2^{11}$ to $2^{20}$.
Note that the driving-request mechanism slightly reduces perception performance, since it prioritizes objects in regions near the predicted waypoints, and the detection precision for objects in other areas, which constitute a substantial portion, is somewhat compromised. Nevertheless, the driving-request mechanism achieves superior performance in driving and planning by allocating precise communication bandwidth to driving-critical regions, which improves communication efficiency.}


\noindent
\textbf{Robustness of driving performance against communication latency and pose errors.} 
\textcolor{blue}{For communication latency, Figure \ref{Fig:Latency_performance} presents the comparison of our CoDriving with other collaboration strategies under various communication latency settings. We see that: i) the system-level driving performance and module-level perception, planning performance degrade as latency increases, while CoDriving consistently achieves superior performance; ii) in planning and driving tasks, CoDriving surpasses no collaboration even under extreme asynchrony, while other collaboration strategies deteriorate to the level of no collaboration.
For pose errors, Figure \ref{Fig:pose_error_performance} compares CoDriving with other collaboration strategies under various pose error levels. The experimental setting follows Section~\ref{subsec:det_metric}. We see that: i) the system and modular performance degrade as pose error increases, while CoDriving achieves superior performance consistently; ii) even in an extreme noise of 0.6 Std, CoDriving surpasses no collaboration by $47.06\%$ in driving score.
The observed robustness is attributed to the multi-scale attention fusion strategy, which helps reduce the sensitivity to disturbances and filter out noise,
as analyzed in Section~\ref{subsec:perception_results}. This advantage extends from perception to the entire driving system, improving both planning and driving performance. See further ablation studies in Appendix~\ref{subsec:robustness_latency}.}

\vspace{-4mm}
\subsection{Qualitative analysis}


\textbf{Does collaboration benefit driving behavior?}
Figure~\ref{fig:DrivingSample} depicts a safety-critical scenario caused by occlusion, a reckless pedestrian abruptly emerges from a telephone booth. From Figure \ref{fig:DrivingSample}(a)(b), we observe that the single-agent driving system would encounter a severe pedestrian collision caused by the miss detection. In contrast, from Figure \ref{fig:DrivingSample}(c)(d), we observe that our CoDriving system
avoids a catastrophic collision by braking in advance as it detects the occluded pedestrian based on the complementary information shared by roadside unit through communication. Compared to single-agent driving, CoDriving provides a safer and more reliable autonomous driving solution.

\vspace{-4mm}
\section{Conclusion and Limitations}\label{sec:conclusion}
This work advances collaborative autonomous driving through both a simulation platform and an end-to-end system. We develop a comprehensive closed-loop V2X fully autonomous driving simulation platform V2Xverse. This platform enables the complete pipeline for developing collaborative autonomous driving systems that target the ultimate driving performance. Meanwhile, V2Xverse maintains adaptability and extensibility to integrate and validate single-functional modules and single-agent driving systems.
Furthermore, we propose a novel end-to-end collaborative autonomous driving system, CoDriving, which enhances driving performance while optimizing communication efficiency by sharing driving-critical perceptual information. Comprehensive evaluations on the entire driving system show that CoDriving significantly outperforms single-agent systems across varying communication bandwidths. 

\textcolor{blue}{This work presents significant potential for future extensions. From a platform perspective, while V2Xverse implements V2X-aided driving scenarios through rule-based simulation, there is room to improve the realism of traffic trajectories and the fidelity of image rendering by incorporating data-driven generative simulation. From a methodological perspective, although our approach leverages low-level planning outputs to enable efficient sharing of perceptual features, an open question remains as to how shared high-level planning intentions could further facilitate planning negotiation. Exploring this direction could maximize the benefits of information exchange, contributing to safer and more efficient autonomous driving.}

\appendices


\vspace{-4mm}
\ifCLASSOPTIONcompsoc
  \section*{Acknowledgments}
\else
  \section*{Acknowledgment}
\fi
\vspace{-1mm}

This research is partially supported by National Natural Science Foundation of China under Grant 62171276, the Science and Technology Commission of Shanghai Municipal under Grant 21511100900.

\ifCLASSOPTIONcaptionsoff
  \newpage
\fi



%


\vspace{-4mm}
\bibliographystyle{IEEEtran}
\bibliography{main}

\vspace{-7mm}
\begin{IEEEbiography}
[{\includegraphics[width=1.0in,height=1.25in, clip,keepaspectratio]{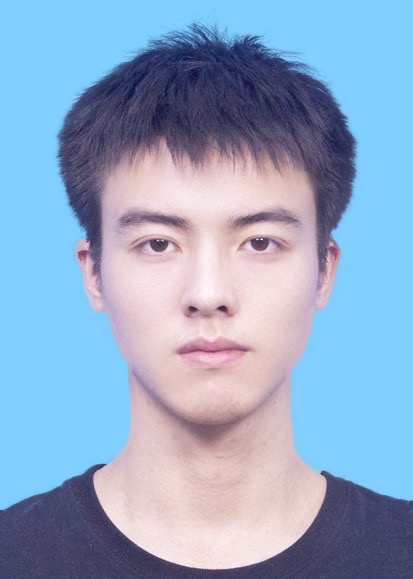}}]{Genjia Liu} is a Ph.D.  student at Cooperative Medianet Innovation Center in Shanghai Jiao Tong University. He received the B.E. degree in information engineering from Shanghai Jiao Tong University in 2021. His research interests include graph signal process and collaborative autonomous driving.
\end{IEEEbiography}
\vspace{-5mm}

\begin{IEEEbiography}[{\includegraphics[width=1.0in,height=1.25in,clip,keepaspectratio]{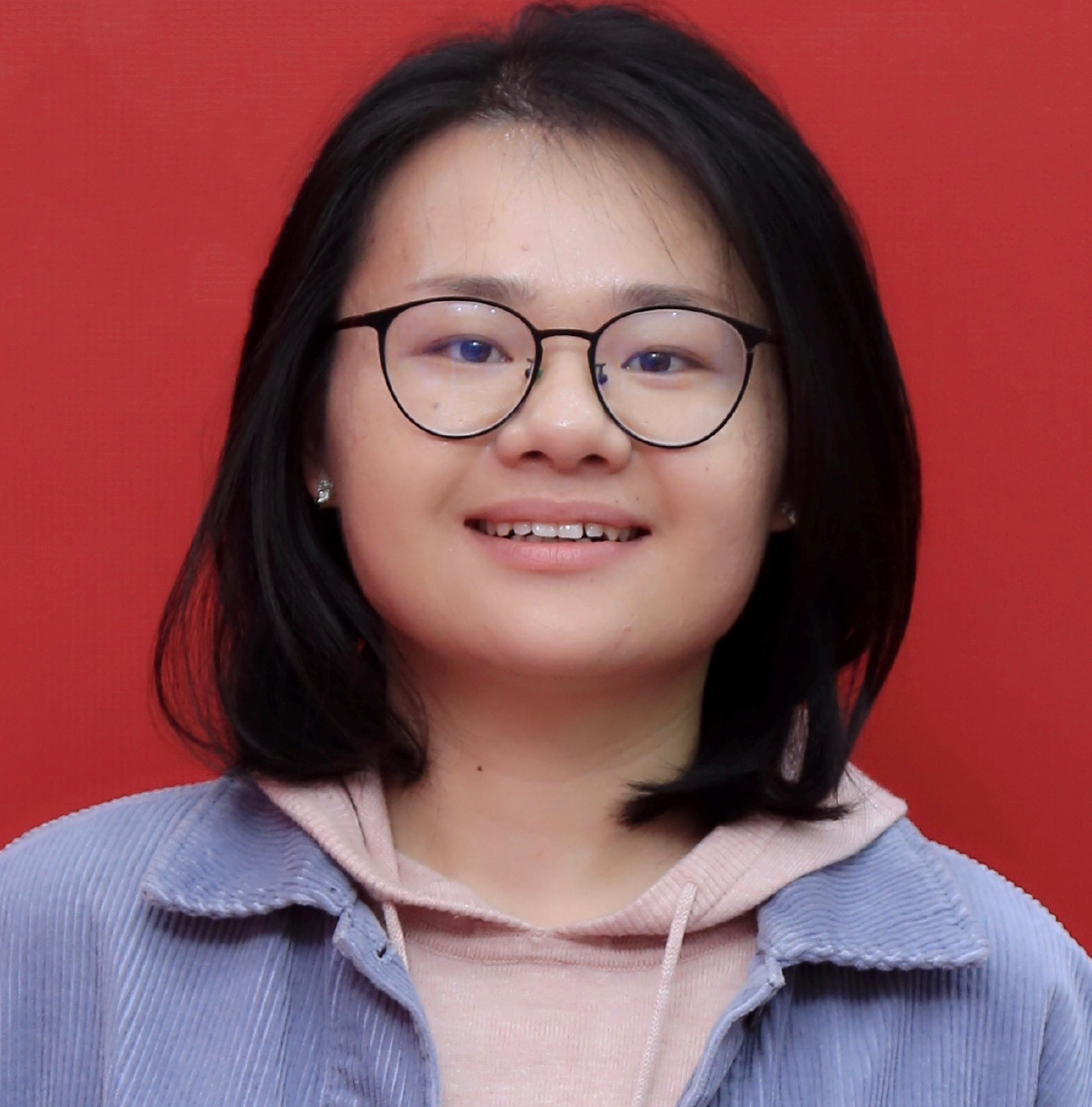}}]{Yue Hu} is working toward the Ph.D. degree at Cooperative Medianet Innovation Center at Shanghai Jiao Tong University since 2021. She received the M.S. degree and B.E. degree in information engineering from Shanghai Jiao Tong University, Shanghai, China, in 2020 and 2017. Her research interests include multi-agent collaboration, communication efficiency, and 3D vision.
\end{IEEEbiography}
\vspace{-3mm}

\vspace{-3mm}
\begin{IEEEbiography}
[{\includegraphics[width=1.0in,height=1.25in, clip,keepaspectratio]{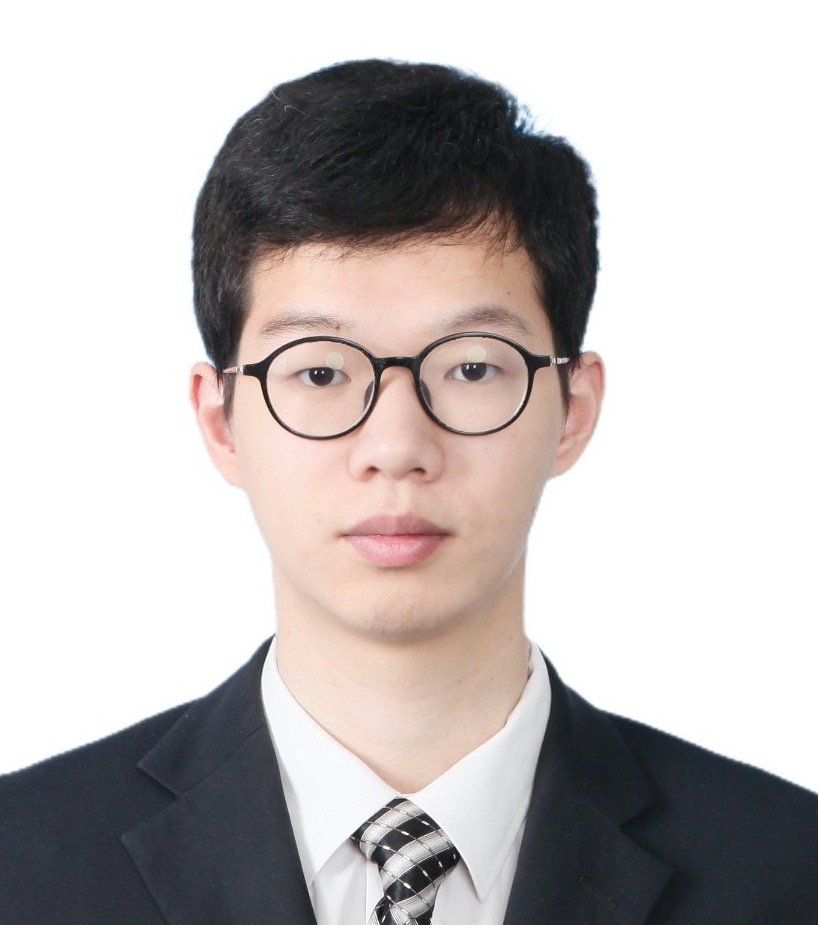}}]{Chenxin Xu} is working toward the joint Ph.D. degree at Cooperative Medianet Innovation Center in Shanghai Jiao Tong University and at Electrical and Computer Engineering in National University of Singapore since 2019. His research interests include trajectory prediction and multi-agent system.  He is the reviewer of some prestigious international journals
and conferences, including IEEE-TPAMI, CVPR, ICCV, ICML and NeurIPS.
\end{IEEEbiography}
\vspace{-3mm}

\vspace{-3mm}
\begin{IEEEbiography}
[{\includegraphics[width=1.0in,height=1.25in, clip,keepaspectratio]{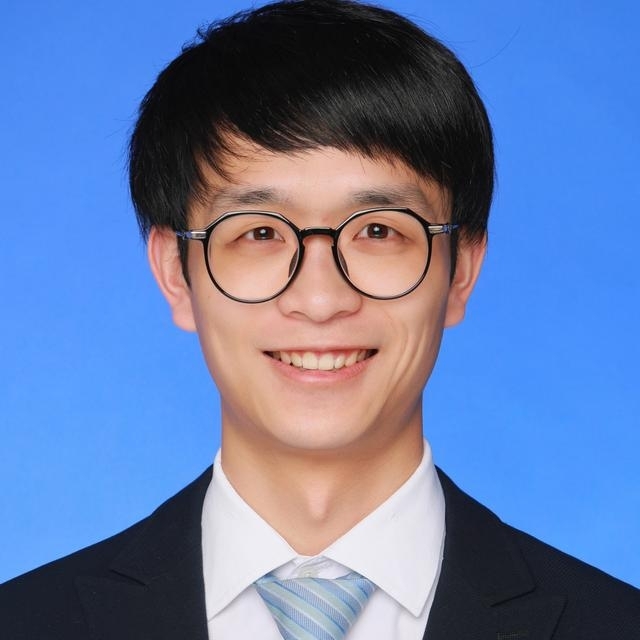}}]{Weibo Mao} was working toward the M.S. degree when the work was done. He received the B.E. degree from Shanghai Jiao Tong University in 2021. His research interests include trajectory prediction and memory intelligence. 
\end{IEEEbiography}
\vspace{-3mm}

\vspace{-3mm}
\begin{IEEEbiography}
[{\includegraphics[width=1.0in,height=1.25in, clip,keepaspectratio]{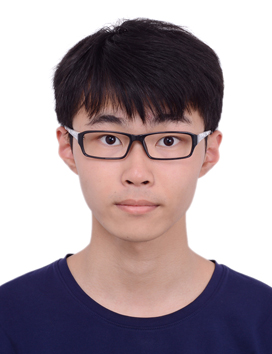}}]{Junhao Ge} is working toward the B.E. degree at Cooperative Medianet Innovation Center in Shanghai Jiao Tong University. His current research field includes traffic simulation and collaborative data generation.
\end{IEEEbiography}
\vspace{-3mm}

\vspace{-3mm}
\begin{IEEEbiography}
[{\includegraphics[width=1.0in,height=1.25in, clip,keepaspectratio]{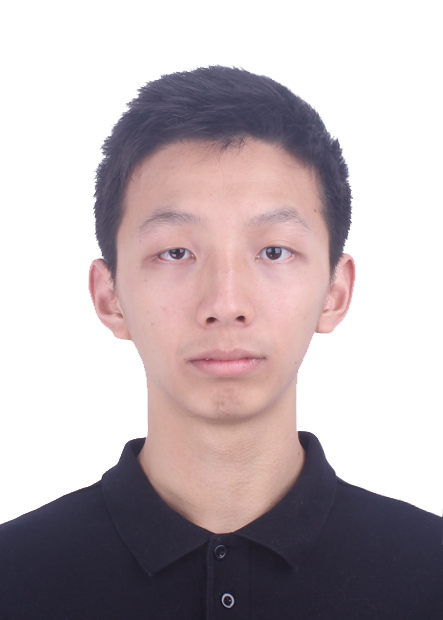}}]{Zhengxiang Huang} is working toward the B.E. degree at Computer Science Department in Shanghai Jiao Tong University. His current research field includes  collaborative data generation and machine learning system design.
\end{IEEEbiography}
\vspace{-3mm}

\vspace{-3mm}
\begin{IEEEbiography}
[{\includegraphics[width=1.0in,height=1.25in, clip,keepaspectratio]{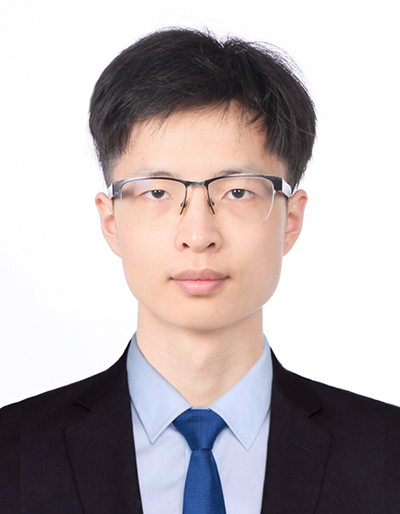}}]{Yifan Lu} is working toward the M.S. degree at Cooperative Medianet Innovation Center in Shanghai Jiao Tong University. He received the B.E. degree in computer science from Shanghai Jiao Tong University in 2022. His current research field includes 3D scene data simulation and generation.
\end{IEEEbiography}
\vspace{-3mm}

\vspace{-3mm}
\begin{IEEEbiography}
[{\includegraphics[width=1.0in,height=1.25in, clip,keepaspectratio]{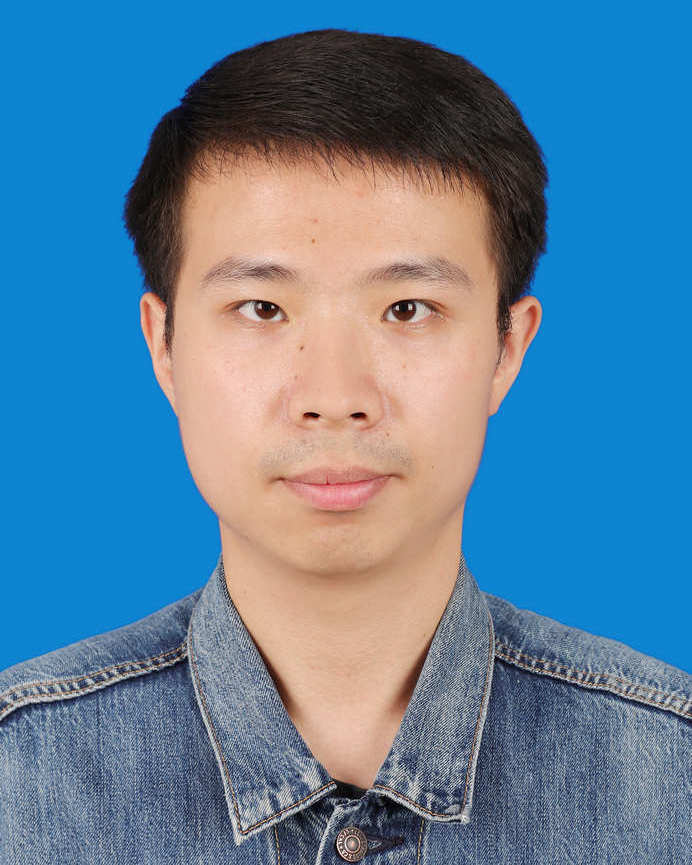}}]{Yinda Xu} is working toward a Ph.D. degree in Computer Science at the Cooperative Medianet Innovation Center in Shanghai Jiao Tong University since 2023. He received both the B.S. degree and M.S. degree in Electrical Engineering from Zhejiang University in 2017 and 2021, respectively. He also served as an Algorithm Engineer in the industry of Autonomous Driving (Momenta.ai and Deeproute.ai, respectively) between 2021 and 2023. His research interests include Robot Learning and Embodied AI.
\end{IEEEbiography}
\vspace{-3mm}

\vspace{-3mm}
\begin{IEEEbiography}
[{\includegraphics[width=1.0in,height=1.25in, clip,keepaspectratio]{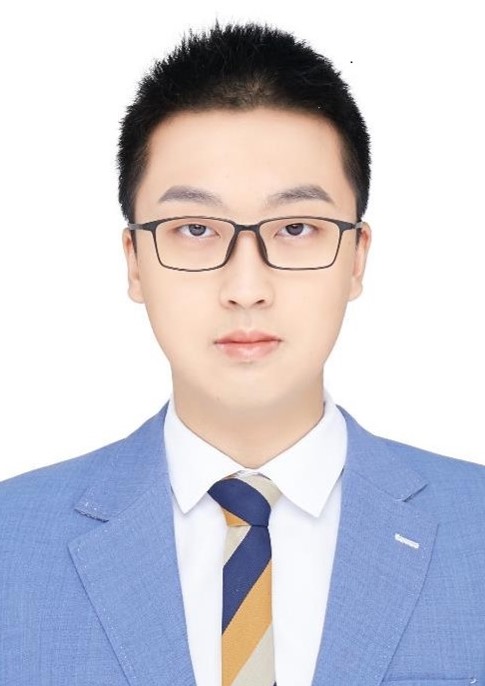}}]{Junkai Xia} is working toward the M.S. degree at SJTU Paris Elite Institute of Technology at Shanghai Jiao Tong University and the Ingénieur Polytechnicien Program at École Polytechnique. He received the B.E.
degree in information engineering from Shanghai
Jiao Tong University in 2022. His research interests include traffic scenarios generation and collaborative autonomous driving.
\end{IEEEbiography}
\vspace{-3mm}

\vspace{-3mm}
\begin{IEEEbiography}[{\includegraphics[width=1.0in,height=1.25in, clip,keepaspectratio]{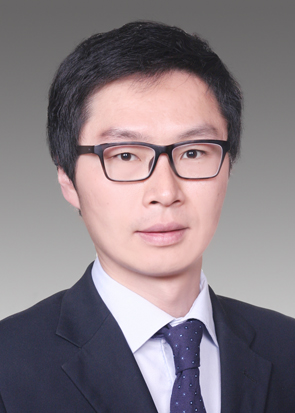}}]{Yafei Wang} (Member, IEEE) received the B.S. degree in internal combustion engine from Jilin University, Changchun, China, in 2005, the M.S. degree in vehicle engineering from Shanghai Jiao Tong University, Shanghai, China, in 2008, and the Ph.D. degree in electrical engineering from The University of Tokyo, Tokyo, Japan, in 2013.
From 2008 to 2010, he was with automotive industry for nearly two years. From 2013 to 2016, he was a Postdoctoral Researcher with The University of Tokyo. He is currently a Associate Professor with the School of Mechanical Engineering, Shanghai Jiao Tong University. His research interests include state estimation and control for connected and automated
vehicles.
\end{IEEEbiography}
\vspace{-3mm}

\vspace{-3mm}
\begin{IEEEbiography}[{\includegraphics[width=1.0in,height=1.25in, clip,keepaspectratio]{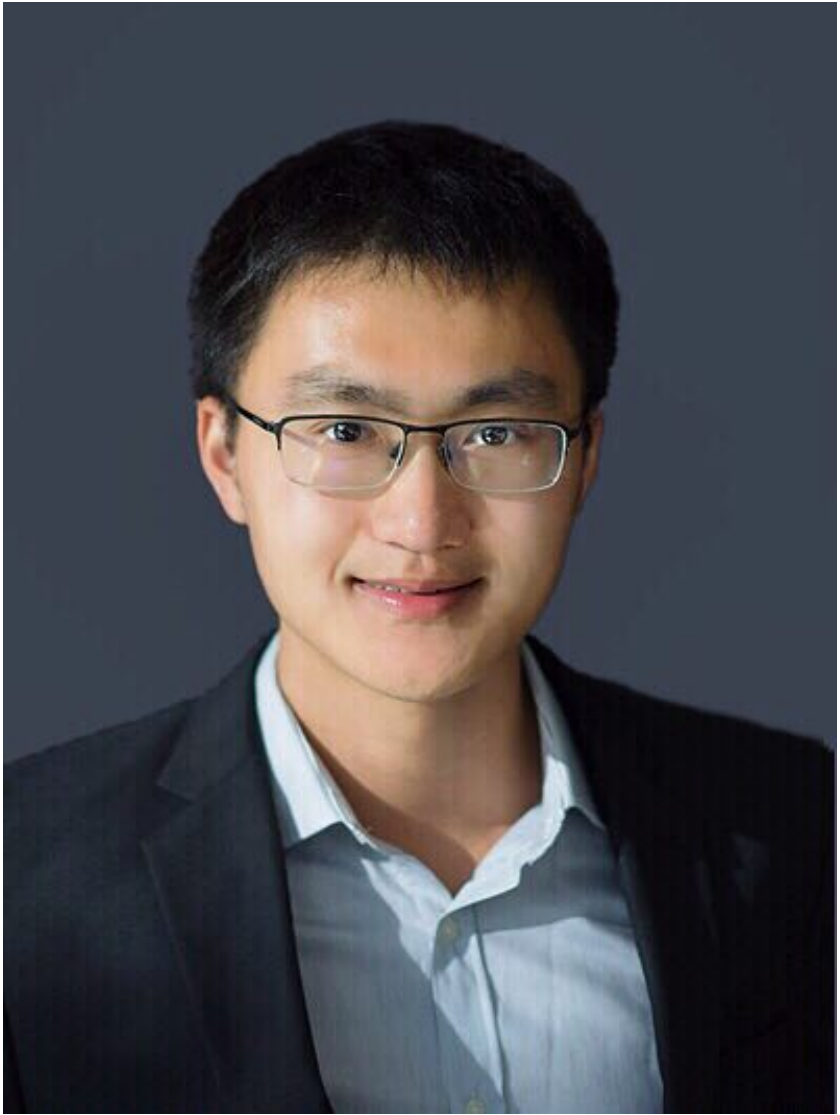}}]{Siheng Chen} is a tenure-track associate professor of Shanghai Jiao Tong University. He was a research scientist at Mitsubishi Electric Research Laboratories (MERL), and an autonomy engineer at Uber Advanced Technologies Group (ATG), working on self-driving cars. Dr. Chen received his doctorate from Carnegie Mellon University in 2016. Dr. Chen's work on sampling theory of graph data received the 2018 IEEE Signal Processing Society Young Author Best Paper Award. He contributed to the project of scene-aware interaction, winning MERL President's Award. His research interests include graph machine learning and collective intelligence.
\end{IEEEbiography}
\vspace{-3mm}

\clearpage
\appendices

\begin{figure}[h]
\centering
\includegraphics[width=0.95\linewidth]{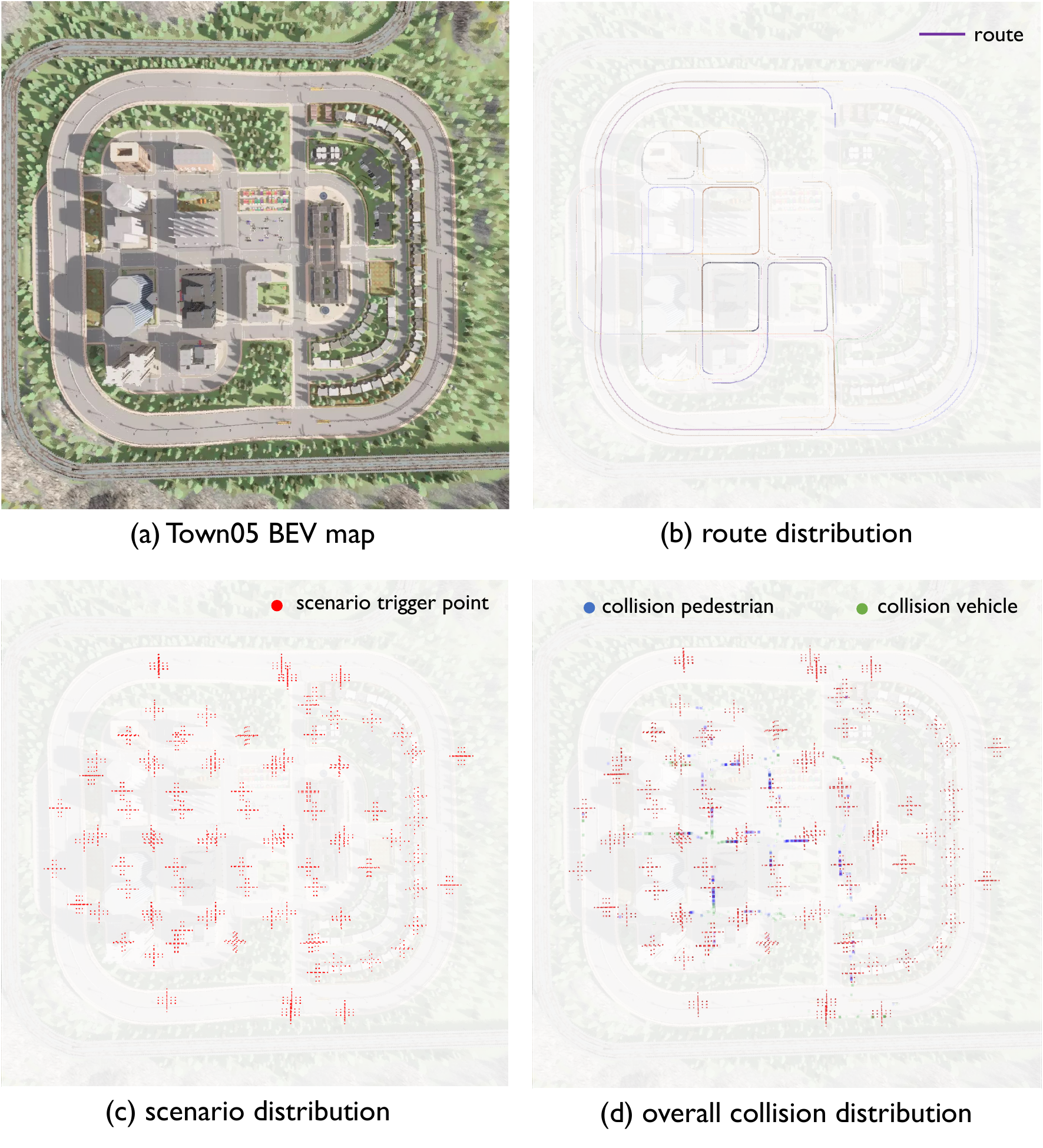}
\vspace{-2mm}
\caption{\small Town05's map, routes, scenario densities, and collision distribution of CoDriving for reference. A darker color represents a larger collision rate.}
\label{fig:map_exmaple}
\vspace{-5mm}
\end{figure}

\vspace{-3mm}
\section{V2Xverse}

\subsection{Platform construction}
\label{sec_A1}
\textbf{Map creation.}
Similar to the setup in~\cite{InterFuser}, we split 5 open-source towns (town1-town4, town6) for offline training, 2 towns (town7, town10) for offline validating, and town5 for testing.
We evaluate driving safety across 67 test routes, covering five scenario configurations. Each configuration defines the map according to four main attributes: vehicle amount, pedestrian amount, probability of pedestrian to violate traffic rules, and the trigger distance of the trigger-based scenarios.
Figure~\ref{fig:map_exmaple} presents the map, routes, density of trigger-based scenarios, and collision rates in Town5.


\noindent
\textbf{Deployment of entities.}
The non-communicating vehicles and pedestrians are initialized with random starting positions and a destination positions, and they will be controlled by a rule-based policy to reach their destination during data collection.
For each route, the communicating vehicles are initially positioned according to the following guidelines: i) Communicating vehicles are placed close to each other; ii) the first communicating vehicle will be positioned at the start point of route. The second and third ones will be positioned 12 meters ahead and behind the first vehicle, respectively. The fourth and fifth communicating vehicles will be placed on the adjacent right and left lanes near the first vehicle; iii) To ensure appropriate positioning, we check their initial states and adjust placement by 1-meter increments if any overlap with other vehicles or boundaries occurs.
RSUs are strategically positioned on the road side alongside the ego car to ensure a comprehensive view of the traffic conditions.
Specifically, to ensure that ego vehicles have at least one roadside unit for valid collaboration at each moment, we adopt a strategy to adjust the position of the roadside unit constantly. A roadside unit will be placed on the right side of the road 12 meters in front of the ego vehicle and the position is updated every five seconds as the ego vehicle moves.

\noindent
\textcolor{blue}{
\textbf{Scenario generation.}
We design challenging scenarios to test the robustness of the autonomous driving system to handle highly dynamic surrounding objects in driving situations with blind spots. 
To increase the dynamics of surrounding objects, we make the following designs: 1) We increase the number of pedestrians and cyclists in the virtual towns, as they are difficult to detect and have flexible trajectories. 2) Pedestrians and cyclists are programmed to ignore traffic rules and common sense. For example, they are allowed to move in vehicle lanes, and they are programmed with a certain probability to not avoid oncoming vehicles. This setup ensures that the driving task remains challenging, as it would lose its difficulty if all other traffic participants efficiently avoided the ego vehicle. 3) Traffic lights were turned off to increase traffic disorder. 
To create blind spots, we make the following designs: 1) We increase the number of large vehicles in the scenes by selecting several types of larger cars from CARLA and increasing their proportion. 2) We create specific trigger-based scenarios, such as pedestrians and cyclists suddenly emerging from behind obstacles, and unprotected left turns at intersections with blocked views. These special scenarios are spread throughout various locations in virtual towns.}

\noindent
\textbf{Sensor setup.}
The collaborative vehicles are equipped with cameras facing 4 directions, including front, left/right angled at ±60°, and rear. Each camera has a field of view (FoV) of 100° and the resolution is 800×600, depth information and semantic segmentation are provided as well. The intrinsic and extrinsic of each camera are recorded. Besides, vehicles are equipped with lidar with 64 channels at a height of 1.9m, the upper FOV is 10° and the lower FOV is -30°. RSU follows a similar sensor configuration and is set up to a height of 7.5m, the side cameras are angled at ±90°, and the lower FOV of lidar is -60°. The position and rotation of each lidar sensor are recorded.

\noindent
\textbf{Data collection.}
During data collection, we deploy 2 expert autonomous vehicles to drive concurrently, spanning 108 routes and 8 towns. Each route is defined by a route file with key points, which are then planned as fine-grained route points by CARLA's built-in navigation module. The two expert vehicles start from neighboring points and follow the same route, ensuring effective collaboration distance. Meanwhile, RSUs are constantly deployed as described in Section~\ref{sec:Platform construction}. The data are collected along the way using a range of heterogeneous sensors from both vehicles and RSUs, with a sampling frame rate of 5 FPS.

\noindent
\textbf{Simulation efficiency.}
We assess the efficiency of V2Xverse by evaluating the time cost during the data collection phase using expert agents. Table~\ref{table:platform_efficiency} shows a detailed breakdown of simulation times per route (650frames) based on varying numbers of agents/sensors. We focus on two major components of time cost: the time to initially load the scenario and the game duration time, where game duration mainly depends on the time to extract and save sensor data. We see that: i) the overall simulation time grows linearly with number of agents; and ii) the average time taken for one simulation step remains under 1.67 seconds, demonstrating the efficiency of V2Xverse.
Therefore, it is fast enough to practically simulate large scale dataset using V2Xverse. To generate a dataset with its volume be comparable to previous autonomous driving datasets, e.g. KITTI(15k)~\cite{Geiger2012kitti}, NuScenes(40K)~\cite{Caesar2020nuScenesAM}, only 15 hours are needed using 2 GeForce RTX 3090 to capture 70K frames.

\begin{table} [!t]
\begin{center}
\small
\resizebox{0.9\linewidth}{!}{
\begin{tabular}{c c c c}
\toprule
\makecell{Sensor-equipped \\ agents} &	\makecell{Load \\ scenario(s)} &	\makecell{Game \\ duration(s)}	& Overall(s)   \\
\midrule
1 vehicle &	26.86	& 175.50&	202.36\\
\midrule
\makecell{1 vehicle \\ and 1 RSU}	& 27.04	& 533.00 & 	560.04 \\
\midrule
2 vehicles &	46.01&	526.50&	572.51 \\
\midrule
\makecell{2 vehicles \\ and 2 RSUs}	& 52.64	&1085.50	&1138.14 \\
\bottomrule
\end{tabular}
}
\end{center}
\vspace{-3mm}
\caption{\small Detailed breakdown of simulation times per route (650 frames) based on varying numbers of agents/sensors.} 
\label{table:platform_efficiency}
\end{table}

\begin{figure*}[!t]
  \centering
  \begin{subfigure}{0.36\linewidth}
    \includegraphics[width=1.0\linewidth]{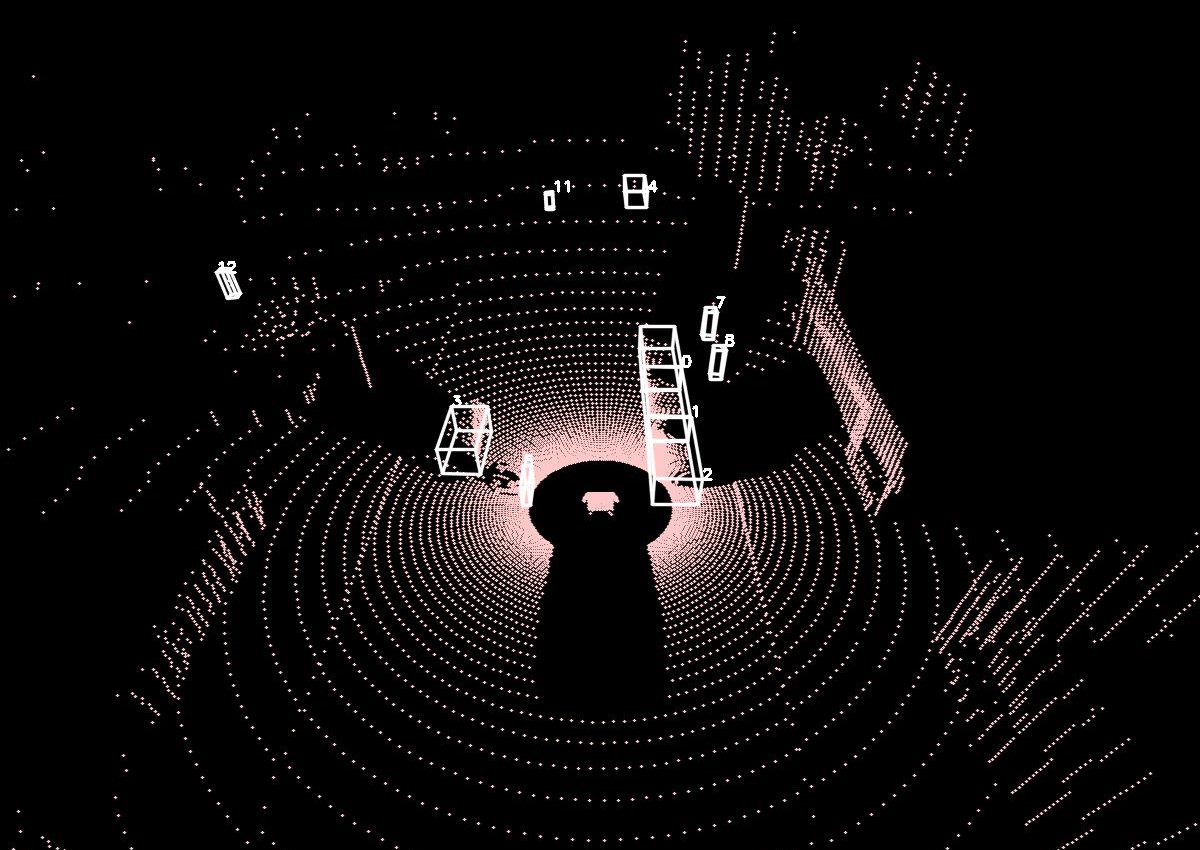}
    \caption{Lidar data of ego vehicle}
    \label{fig:data_demo_1}
  \end{subfigure}
  \begin{subfigure}{0.36\linewidth}
    \includegraphics[width=1.0\linewidth]{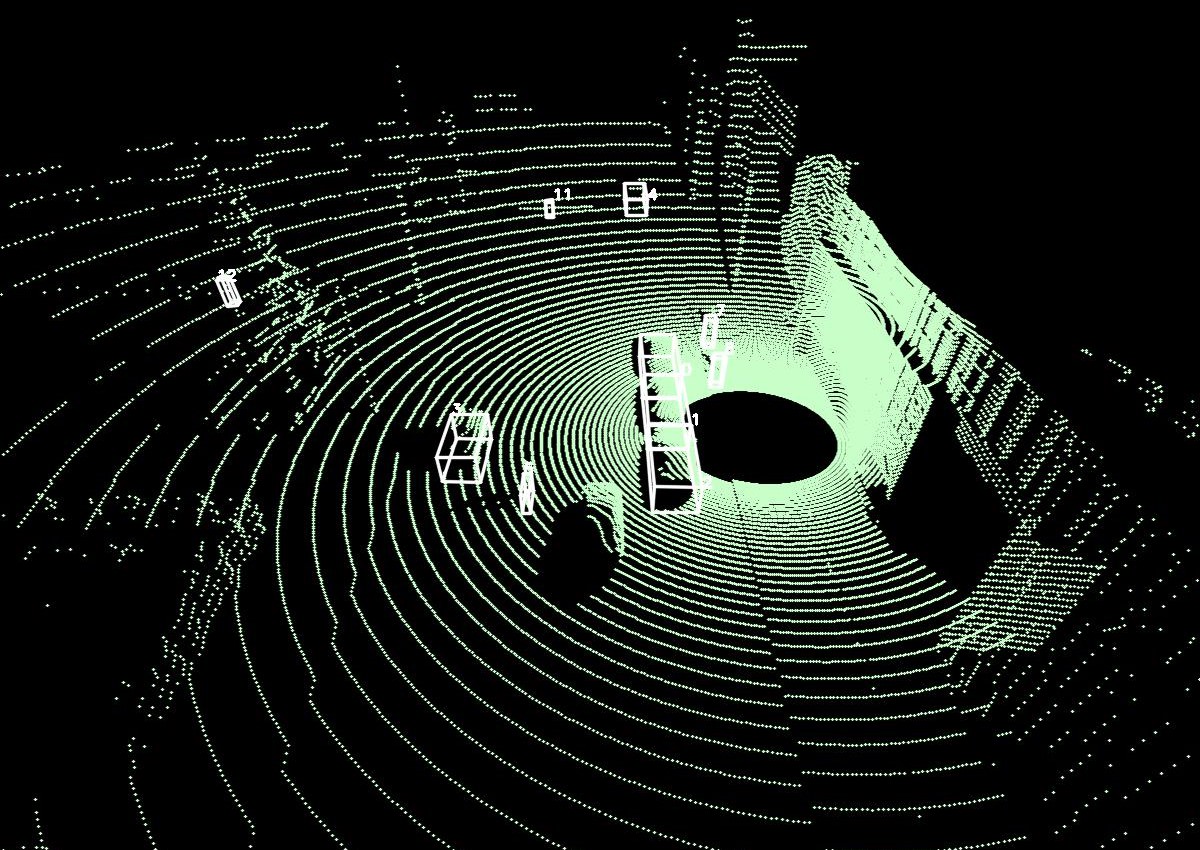}
    \caption{Lidar data of RSU}
    \label{fig:data_demo_2}
  \end{subfigure}
  \begin{subfigure}{0.36\linewidth}
    \includegraphics[width=1.0\linewidth]{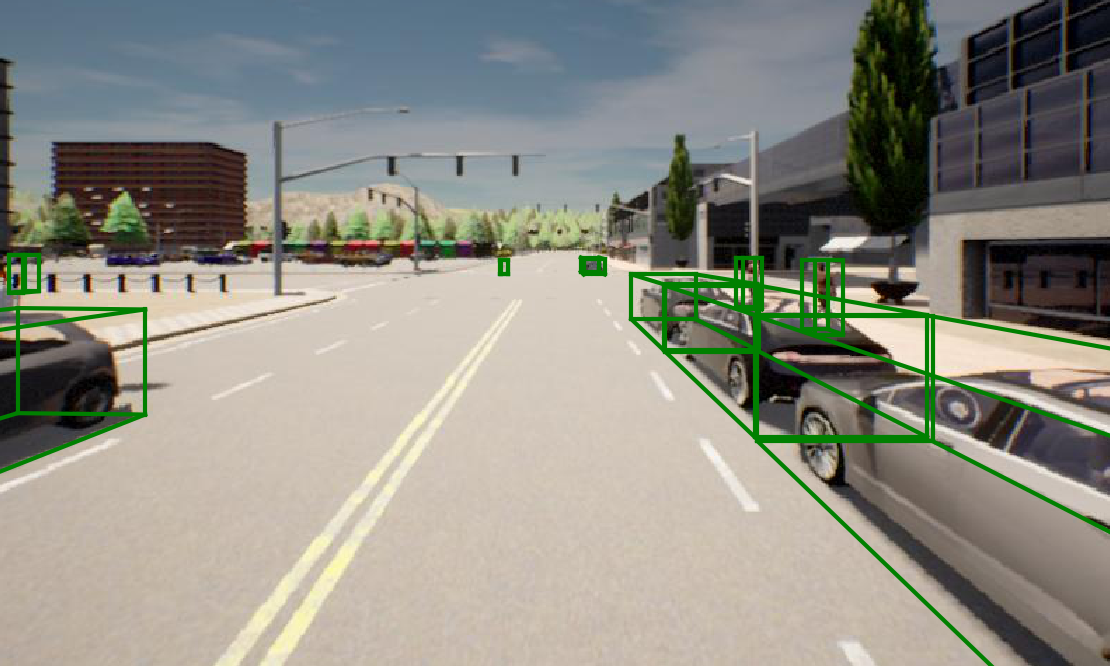}
    \caption{Front image of ego vehicle}
    \label{fig:data_demo_3}
  \end{subfigure}
  \begin{subfigure}{0.36\linewidth}
    \includegraphics[width=1.0\linewidth]{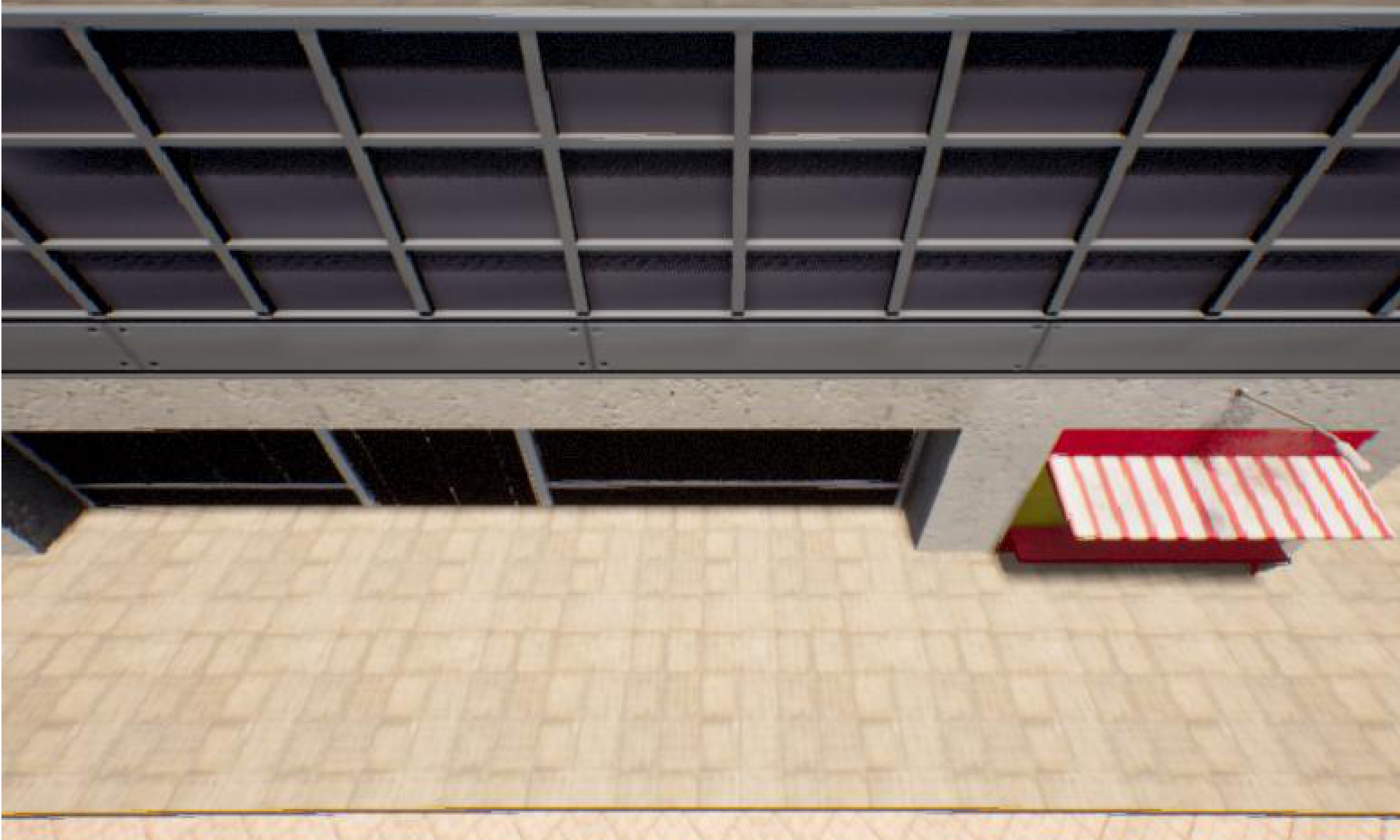}
    \caption{Front image of RSU}
    \label{fig:data_demo_4}
  \end{subfigure}
    \begin{subfigure}{0.36\linewidth}
    \includegraphics[width=1.0\linewidth]{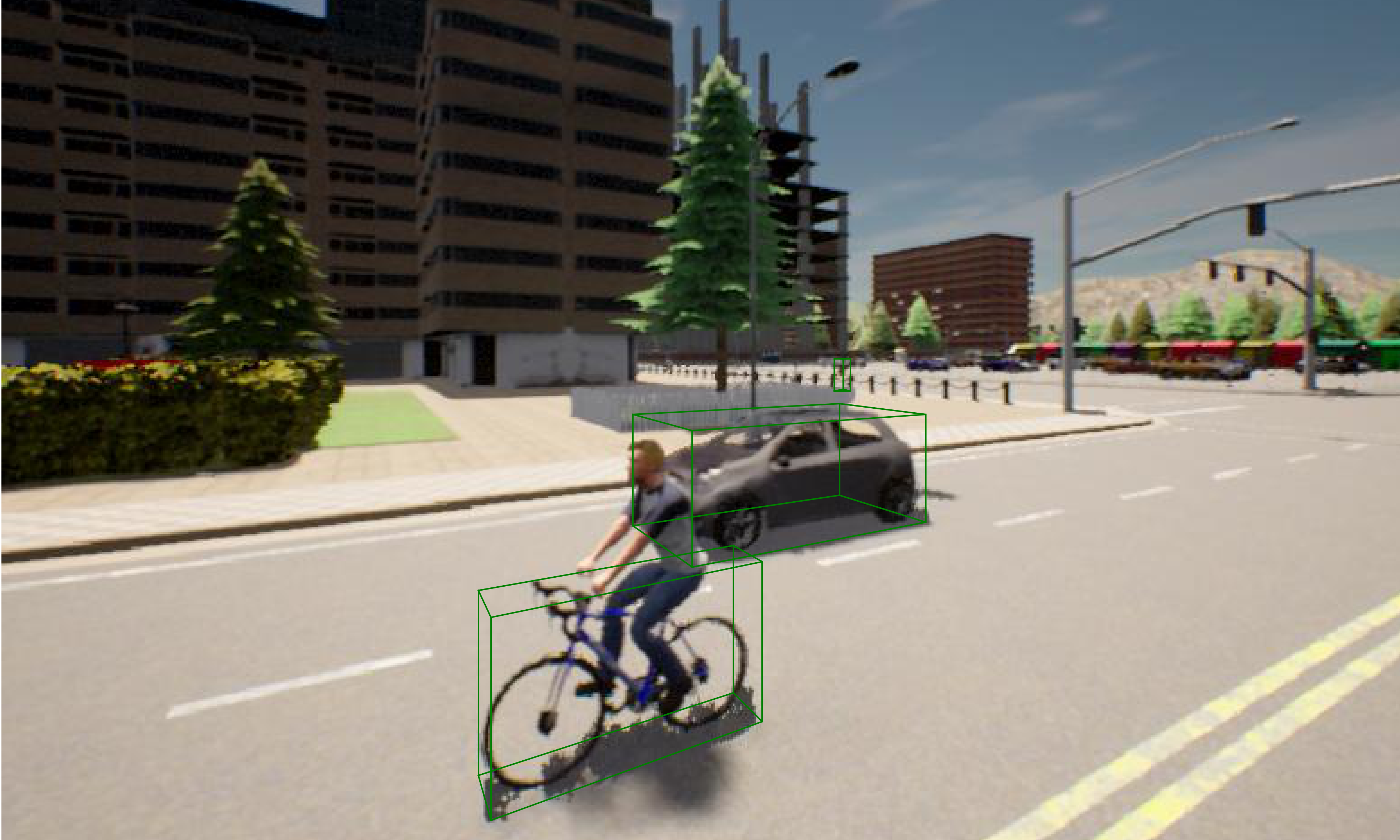}
    \caption{Left side image of ego vehicle}
    \label{fig:data_demo_5}
  \end{subfigure}
  \begin{subfigure}{0.36\linewidth}
    \includegraphics[width=1.0\linewidth]{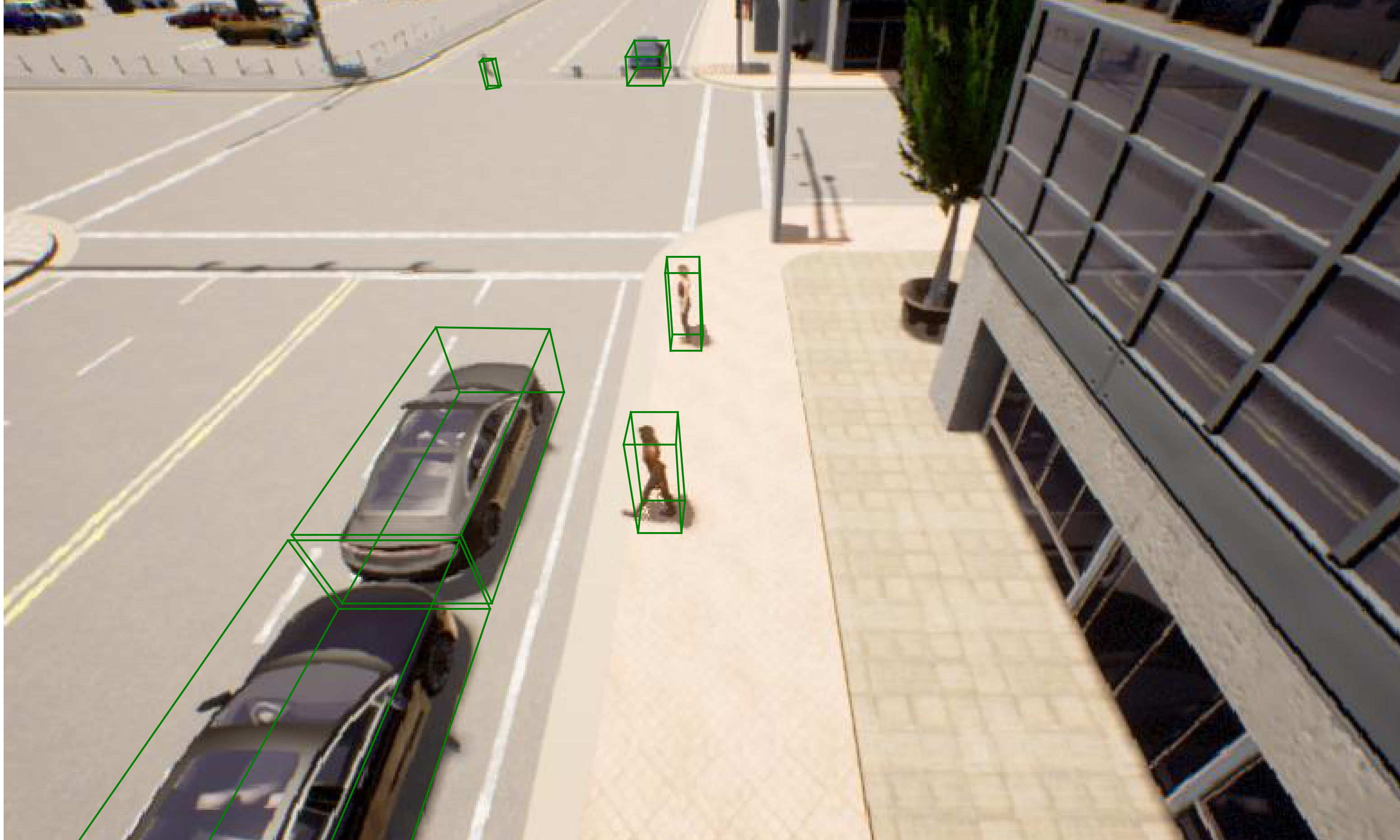}
    \caption{Left side image of RSU}
    \label{fig:data_demo_6}
  \end{subfigure}
    \begin{subfigure}{0.36\linewidth}
    \includegraphics[width=1.0\linewidth]{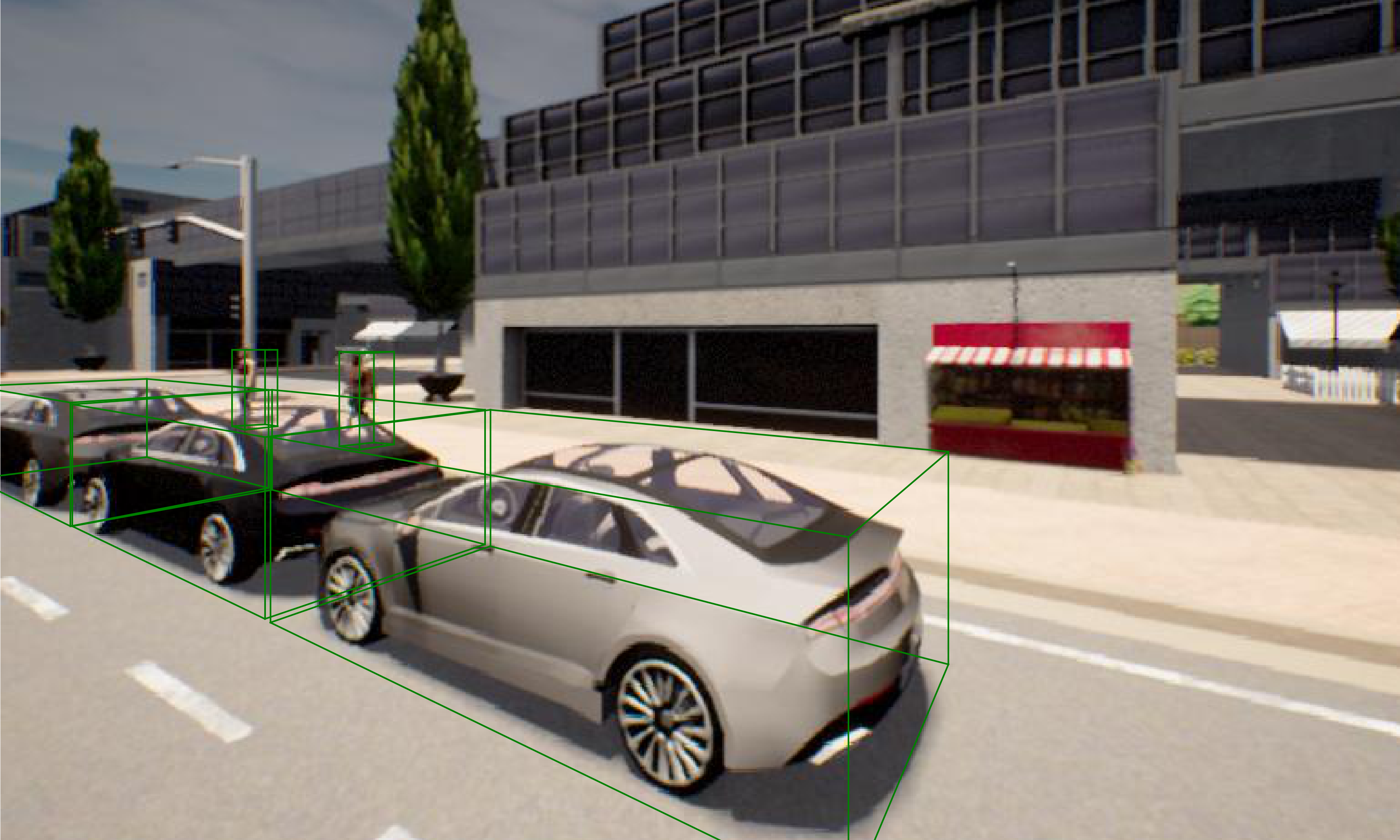}
    \caption{Right side image of ego vehicle}
    \label{fig:data_demo_7}
  \end{subfigure}
  \begin{subfigure}{0.36\linewidth}
    \includegraphics[width=1.0\linewidth]{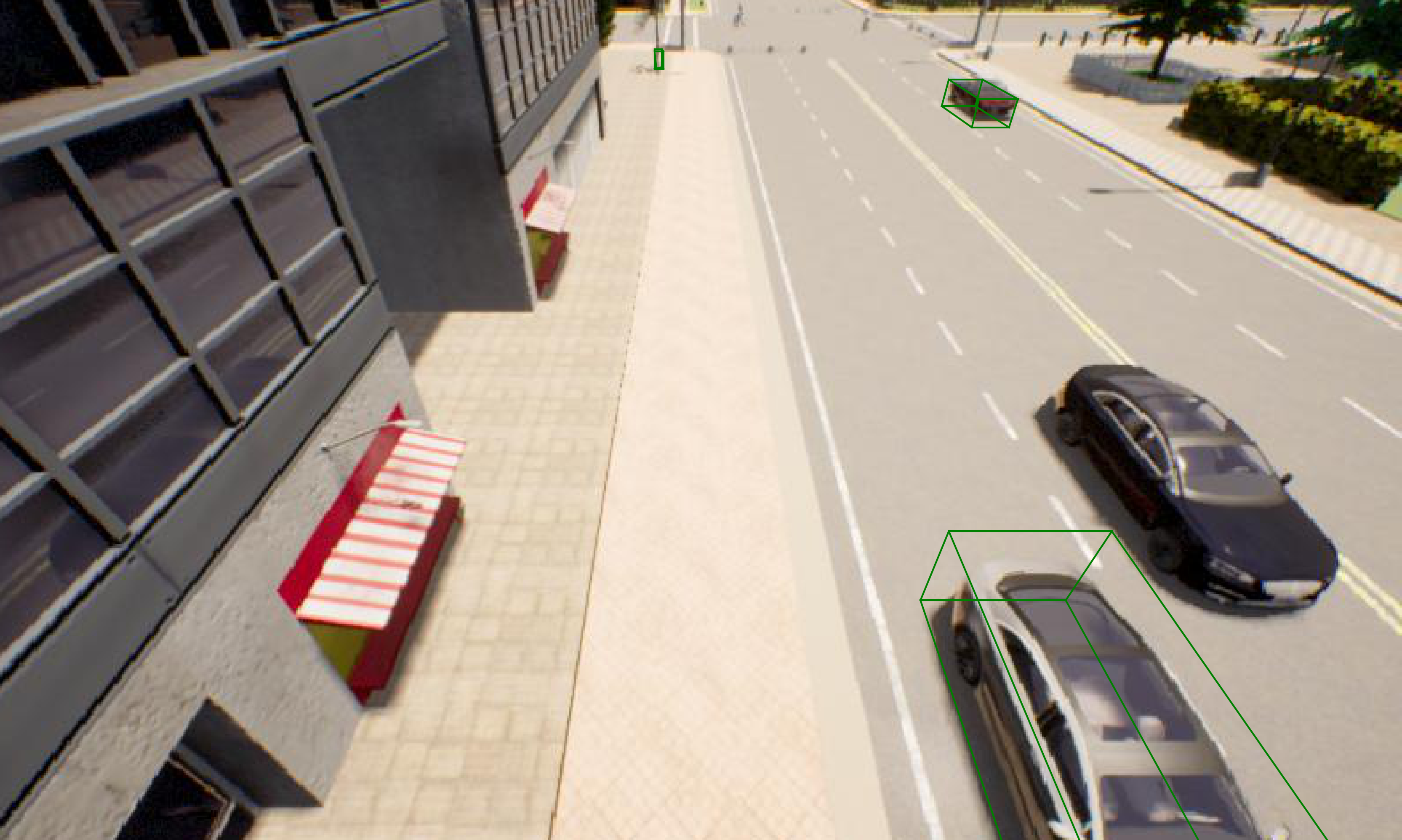}
    \caption{Right side image of RSU}
    \label{fig:data_demo_8}
  \end{subfigure}
    \begin{subfigure}{0.36\linewidth}
    \includegraphics[width=1.0\linewidth]{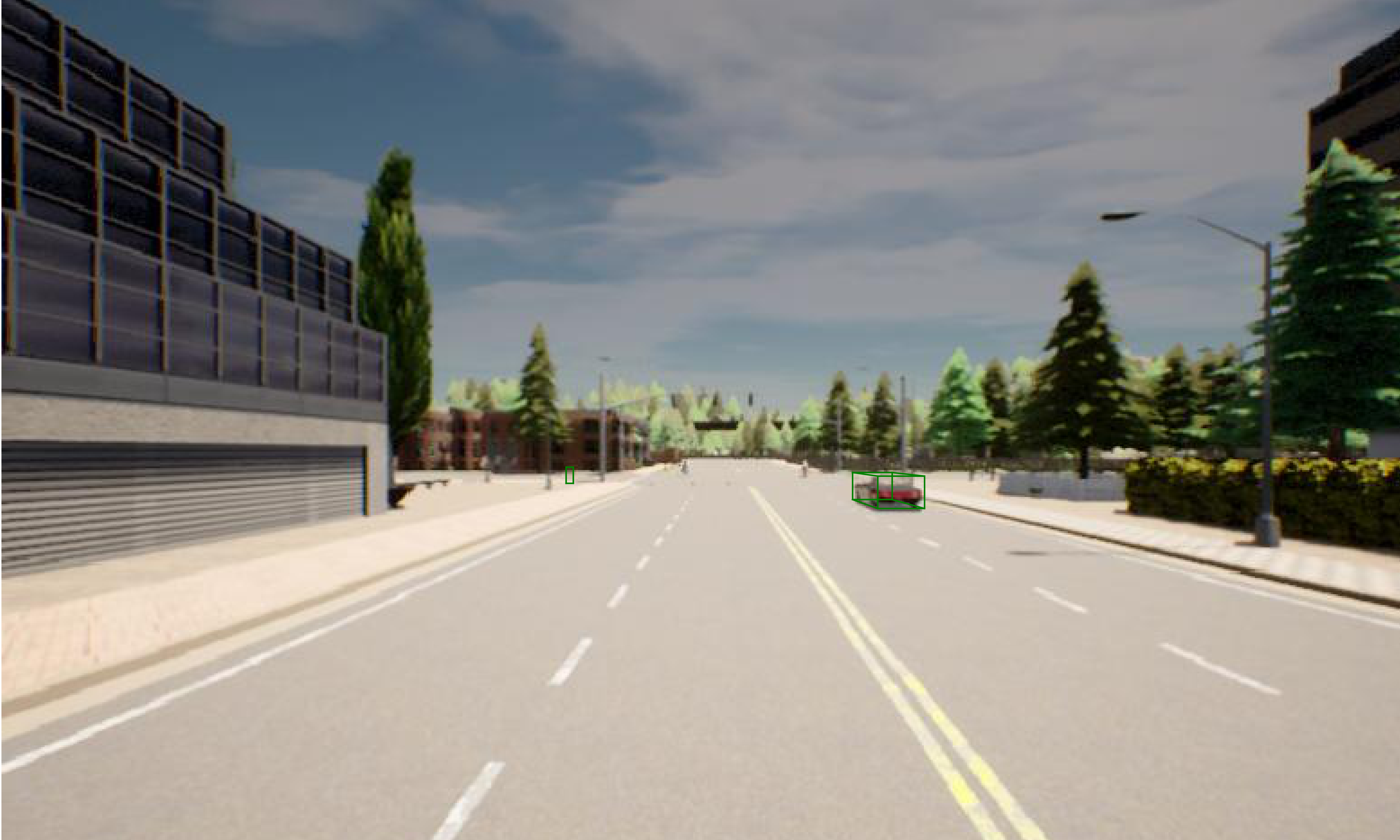}
    \caption{Rear image of ego vehicle}
    \label{fig:data_demo_9}
  \end{subfigure}
  \begin{subfigure}{0.36\linewidth}
    \includegraphics[width=1.0\linewidth]{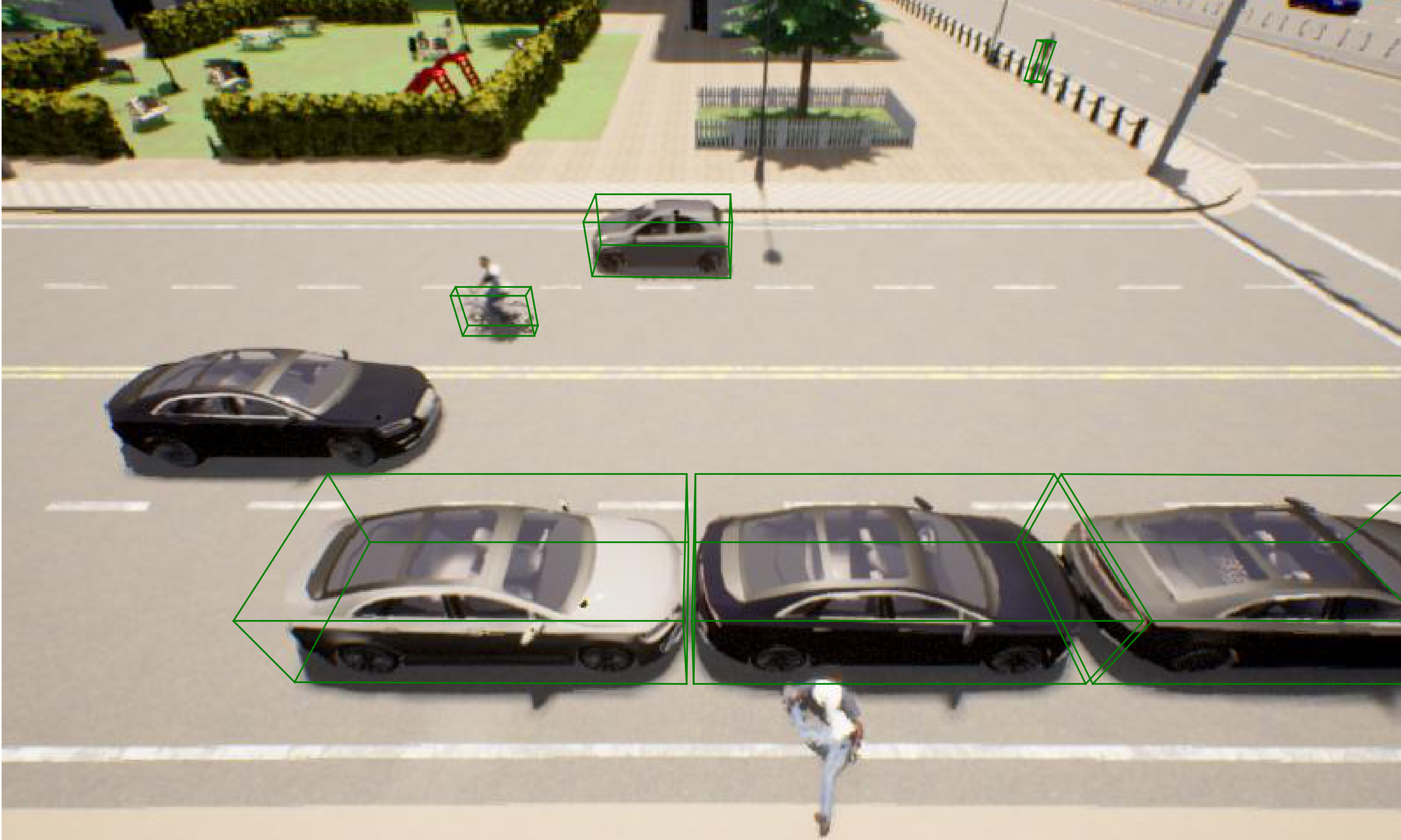}
    \caption{Rear image of RSU}
    \label{fig:data_demo_10}
  \end{subfigure}
  \vspace{-1mm}
  \caption{\small Data and annotations in a single frame. The left column and right column present the vehicle view and RSU view respectively. It includes LiDAR data and RGB images from different views, along with 3D bounding boxes. The bounding box of ego vehicle is removed here.}
  \label{fig:sample_example}
  \vspace{-3mm}
\end{figure*}

\begin{table*}[t]
\centering
\caption{\small \textcolor{blue}{Comparison of 3D cooperative V2X perception benchmarks with our proposed V2Xverse. V2Xverse surpasses previous benchmarks in the volume of sensor data and detection annotations.}}
\resizebox{0.97\linewidth}{!}{
\begin{tabular}{l|cccccccc}
\toprule
\multicolumn{1}{l|}{{  Benchmarks}} & {  OPV2V~\cite{XuOPV2V:ICRA22}} & {  V2XSet~\cite{xu2022v2xvit}} & {  V2X-Sim~\cite{LiV2XSim:RAL22}} & {  V2V4Real~\cite{XuV2V4Real:CVPR23}} & {  DAIR-V2X~\cite{YuDAIRV2X:CVPR22}} & {  V2X-Seq~\cite{YuV2XSeq:CVPR23}} & {  TUMTraf~\cite{zimmer2024tumtraf}} & V2Xverse (Ours) \\ \midrule
{  Year} & {  2022} & {  2022} & {  2022} & {  2022} & {  2022} & {  2023} & {  2024} & \multicolumn{1}{c}{2024} \\
\multicolumn{1}{l|}{{  Point Clouds}} & {  11k} & {  11k} & {  10k} & {  20k} & {  39k} & {  15k} & {  2.0k} & 72k \\
{  Images} & {  44k} & {  44k} & {  60k} & {  40k+} & {  39k} & {  15k} & {  5.0k} & 288k \\
{  3D Boxes} & {  233k} & {  233k} & {  26k} & {  240k} & {  464k} & {  10.45k} & {  29.38k} & 482k \\
{  Location} & {  CARLA} & {  CARLA} & {  CARLA} & {  USA} & {  China} & {  China} & {  Germany} & CARLA \\ \bottomrule
\end{tabular}}
\label{compare_v2xverse_Dataset}
\end{table*}

\vspace{-3mm}
\subsection{Data description}
\label{sec:appendix data descrip}
Fully-annotated data are provided in our proposed dataset, including synchronous images with pixel-wise semantic labels, 3D point clouds from LiDAR, 2D \& 3D bounding boxes of vehicles and pedestrians, and binary BEV drivable area map. Figure~\ref{fig:sample_example} shows an example of the existing data with annotations, where the ten images are captured by cameras/LiDARs on vehicles and RSU simultaneously.

\noindent
\textbf{LiDAR data.}
We collect synchronous 3D point clouds data from LiDAR on both ego vehicles and RSUs. The information of the LIDAR measurement is encoded as 4D points. Being the first three, the space points in xyz coordinates and the last one intensity loss during the travel. 

\noindent
\textbf{Camera data.}
We collect synchronous images from all cameras on each agent, which are 8 images in a sample with one ego vehicle and one RSU. Camera intrinsics and extrinsic in global coordinates are provided to support coordinate transformation across different views.

\noindent
\textbf{Depth data.}
We collect depth maps that potentially contain useful spatial information. We use the depth map to determine whether an object is occluded.

\noindent
\textbf{Waypoints.}
The world coordinates $(x,y,z)$ of 50 nearest target waypoints that the ego vehicle will drive to are recorded. These points are planned and interpolated by the navigation module of CARLA and the distance between two adjacent waypoints is approximately 1m.

\noindent
\textbf{Control.}
Command and output behavior of the expert agent are recorded at each frame, including the nearest waypoint to be reached, macro driving command (void, turn left, turn right, go straight, lane follow, lane change left, lane change right). Output behavior includes values of steer, throttle, brake. The id of those hazard actors that caused the expert model to behave brake are recorded at each frame.

\noindent
\textbf{Bounding boxes.}
During data collection, 3D bounding boxes of objects are recorded synchronously, including location $(x,y,z)$ in world coordinate, rotation cosine of roll, yaw, pitch, and half value of length, width and height. The location $(x,y,z)$ is the bottom of the bounding box for vehicles and the center of the bounding box for pedestrians. Three types of dynamic objects are recorded, including pedestrians, bicycles, and motor vehicles.

\begin{wrapfigure}{R}{0.45\linewidth}
    \centering
    \includegraphics[width=\linewidth]{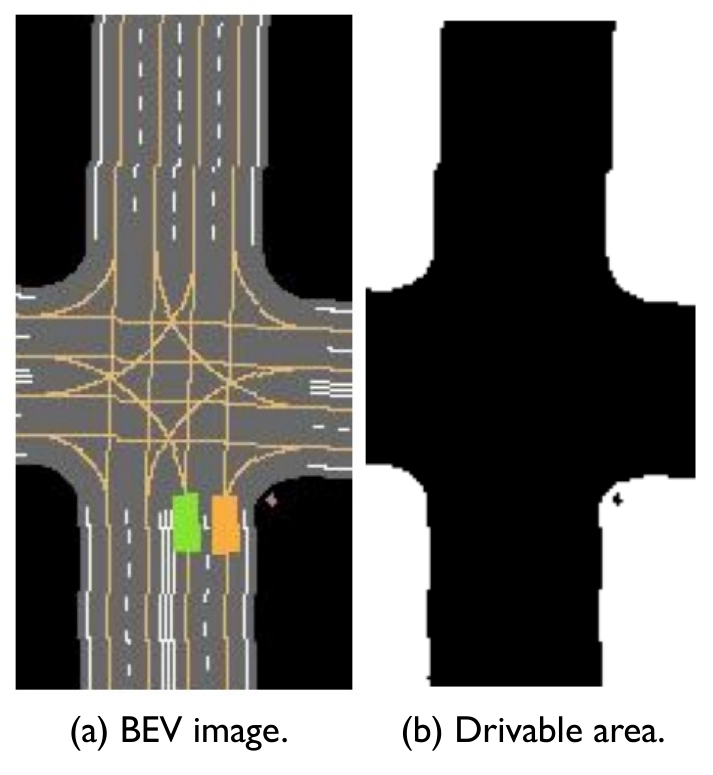}
    \caption{\small Example of the drivable area. (a) shows the original BEV image. (b) shows the processed drivable area image (black for drivable).}
    \label{fig:drivable_area}
\end{wrapfigure}

\noindent
\textbf{Drivable area.} To prevent the waypoint planner from unrealistic predictions (e.g., driving on the sidewalk), we produce the coarse-grained drivable area in extra from the BEV image to indicate whether there is a road or not; see Figure~\ref{fig:drivable_area}. The original BEV image has the range of 96m $\times$ 96m with a resolution of 0.16m/pixel where the car is located as the center. We first crop the image to 24m $\times$ 48m, matching the detection range. We then resize the image from 120$\times$240 to 96$\times$192 and generate the final drivable area using the threshold segmentation with the pixel threshold of 200. Although the original BEV image contains more information including lane segmentation, vehicle position, and road direction, this information is usually unreachable during the closed-loop evaluation. Instead, we use threshold segmentation to generate the coarse-grained drivable area to provide a sense for the planning module.

\noindent
\textbf{Evaluation metrics.} For both safe and efficient driving, we use several metrics to comprehensively evaluate driving performance.
These metrics include the driving score (DS), route completion ratio (RC), infraction score (IS), pedestrian/vehicle/layout collision rate, and mean speed.
We also employ several widely-used metrics like average precision (AP) for the perception task and average displacement error (ADE) and final displacement error (FDE) for the waypoints planning task. 
Here we explain the details of these metrics.
Specifically, the driving score is derived from the product of route completion and infraction score. Route completion indicates the percentage of route distance completed by an agent.
The infraction score tracks several types of infractions triggered by an agent, aggregating them as a geometric series. Each agent starts with an ideal base infraction score of 1.0, which is reduced by a specific ratio each time an infraction is committed. The reduction factors are: collisions with pedestrians/other vehicles/static elements — 0.50/0.60/0.65; scenario timeout — 0.7; failure to maintain minimum speed — 0.7; failure to yield to emergency vehicle — 0.7. Once all routes are completed, a global DS/RC/IS is calculated as the average of individual route scores. Besides, we record collision rates for pedestrians, vehicles, and layout, measured as occurrences per kilometer. 

\noindent
\textbf{Comparisons with other datasets.}
\textcolor{blue}{
Table~\ref{compare_v2xverse_Dataset} presents comparisons of 3D cooperative V2X perception datasets with the dataset generated by V2Xverse, showing that our dataset surpasses previous ones in size. To ensure diversity, we collected data across 108 routes in 8 virtual towns. Specifically, our dataset includes driving data from 36,130 frames, where 2 vehicles and 2 roadside units (RSUs) simultaneously collect sensor data. For 3D boxes, we capture objects within a 50-meter radius of the vehicles and RSUs, and provide visibility attributes for boxes in the view of each sensor. Compared to previous benchmarks~\cite{XuOPV2V:ICRA22,xu2022v2xvit,LiV2XSim:RAL22,XuV2V4Real:CVPR23,YuDAIRV2X:CVPR22,YuV2XSeq:CVPR23,zimmer2024tumtraf}, V2Xverse provides the largest-scale image and point cloud data, along with the most extensive 3D detection box annotations, establishing a robust benchmark for evaluating perception functionality in V2X-aided autonomous driving.}

\begin{figure*}[!t]
\centering
\includegraphics[width=1.0\linewidth]{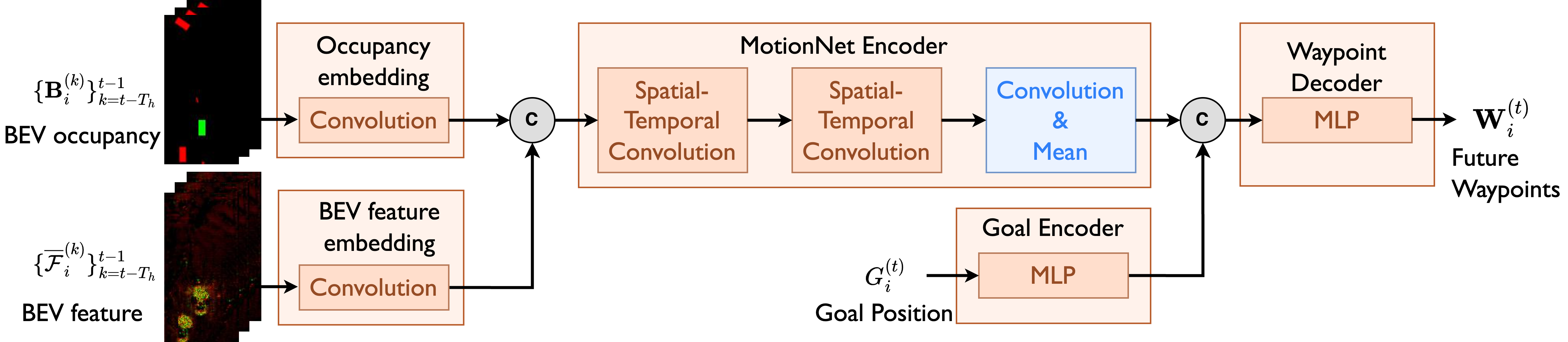}
\caption{\small Architecture of the planning module, which plans the future waypoints according to the historical observed environment and goal position.}
\label{fig:Planner}
\vspace{-2mm}
\end{figure*}

\vspace{-3mm}
\section{Configurations of CoDriving }
This section presents supplementary introduction of the modules in our CoDriving system, including object detector, waypoint predictor, and controller, from network architecture to training strategy aspects.

\noindent
\textbf{3D object detector.}
Our CoDriving system supports input modality of either RGB image or 3D point cloud. For camera-based 3D object detection, we implement the detector following CADDN~\cite{ReadingCategorical:CVPR2021}. 
For LiDAR-based 3D object detection, our detector follows PointPillar~\cite{Lang2018PointPillarsFE}.
\textcolor{blue}{Given a point cloud $pc_i^{(t)}\in\mathbb{R}^{N\times4}$ observed by the $i$-th agent at timestamp $t$, where each point is assigned with a 3D coordinate and reflectance, we follow PointPillars~\cite{Lang2018PointPillarsFE} to extract BEV features. The detection range is set to $x$ (vertical) $\in$ [-12m, 36m], $y$ $\in$ [-12m, 12m], with a voxel grid size of [0.125m, 0.125m]. The point cloud is first converted into a set of pillars $\mathcal{P}_i^{(t)}\in \mathbb{R}^{P_i^{(t)}\times n\times D}$ by re-sampling and grouping, where $P_i^{(t)}$ denotes the number of non-empty pillars, $n$ denotes the number of points per pillar, and $D=10$ is the feature dimension, including reflectance (1), original 3D coordinates (3), and coordinates relative to the center of pillar (3), coordinates relative to the arithmetic mean of all points in the pillar (3). 
For all points, a linear layer $(10\rightarrow64)$ is applied, followed by Batch-Normalization, ReLU, and a pillar-wise max operation to generate an output feature of size $(P_i^{(t)},64)$. This flattened feature is then scattered back to the original pillar locations to create a BEV feature of size (64, 192, 384). A BEV encoder $\Phi_{\rm enc}(\cdot)$ is then applied to further encode the BEV feature. The encoder first reduces the spatial resolution through three convolution blocks. Each block has $L$ 2D convolution layers (K,S,F) with kernel size K, stride S, and F output channels. Each layer is followed by Batch-Normalization and ReLU. The output features from the three convolution blocks are upsampled by three transposed convolution layers and concatenated to obtain the final output of size (384, 96, 192). Our BEV encoder follows the backbone in PointPillars, and the detailed configuration is presented in Table~\ref{BEV_encoder}.
}

\vspace{-2mm}
\begin{table}[H]
\centering
\caption{\small \textcolor{blue}{Configuration of BEV encoder $\Phi_{\rm enc}(\cdot)$ in CoDriving}}
\vspace{-2mm}
\resizebox{0.95\linewidth}{!}{
\begin{tabular}{l|c|c|c}
\toprule
Modules & \begin{tabular}[c]{@{}c@{}}Convolution layers \\ (kernel size, stride, channel)\end{tabular} & \begin{tabular}[c]{@{}c@{}}Input size\\ (C,H,W)\end{tabular} & \begin{tabular}[c]{@{}c@{}}Output size\\ (C,H,W)\end{tabular} \\ \midrule
Conv block 1 & (3,2,64)*1, (3,1,64)*5 & (64, 192, 384) & (64, 96, 192) \\
Conv block 2 & (3,2,128)*1, (3,1,128)*7 & (64, 96, 192) & (128, 48, 96) \\
Conv block 3 & (3,2,256)*1, (3,1,256)*9 & (128, 48, 96) & (256, 24, 48) \\
Transconv layer 1 & (1,1,128) & (64, 96, 192) & (128, 96, 192) \\
Transconv layer 2 & (2,2,128) & (128, 48, 96) & (128, 96, 192) \\
Transconv layer 3 & (4,4,128) & (256, 24, 48) & (128, 96, 192) \\ \bottomrule
\end{tabular}}
\label{BEV_encoder}
\end{table}
\vspace{-2mm}

We conduct training with an initial learning rate of 1e-3 and a decay of 0.1 every 8 epochs. 
It takes approximately 15 hours to converge with 2 NVIDIA-A100 GPUs for LiDAR-based 3D object detection, and about 30 hours to converge for camera-based 3D object detection.

\noindent
\textbf{Waypoints planning.} The waypoint planning module serves as a planner, which takes a sequence of historical perceptual information and the future goal position as inputs to generate trajectories towards the goal; see Figure~\ref{fig:Planner}. The input perceptual information includes BEV feature maps $\{\overline{\mathcal{F}}_i^{(k)}\}_{k=t-T_{h}}^{t-1}$ and structural BEV occupancy images $\{\mathbf{B}_{i}^{(k)}\}_{k=t-T_{h}}^{t-1}$. $\overline{\mathcal{F}}_i^{(k)}$ is the enhanced BEV feature with 128 channels, providing an informative representation of BEV observation. 
Meanwhile, we incorporate occupancy map with auxiliary spatial information, resulting in a 6-channel BEV occupancy image $\mathbf{B}_{i}^{(k)}$: the rasterized occupancy map (1), goal position (1), ego position (1), X-Y coordinate (2), and the drivable area (1). Here we render the ego-vehicle position, goal position, and the $(x,y)$ coordinate in an image of the same size as the BEV occupancy map (96 $\times$ 192, 0.25m/pixel).
We concatenate the last 1-second sequence (5 frames) of both BEV feature and BEV occupancy image to retain historical information. Then, we pre-embed the BEV features and BEV occupancy images using convolution layers and concatenate them as a complete BEV representation.
We use MotionNet~\cite{Wu2020MotionNetJP} to extract the spatial-temporal features from the BEV representation and apply an average-pooling operator to reduce the spatial dimension, resulting in a feature vector.
Furthermore, we use an MLP-based goal encoder to explicitly encode the goal position information into a feature vector.
The two encoded feature vectors are then concatenated and sent to an MLP-based waypoint decoder to generate the final future waypoints. The MLP in the goal encoder has the structure of $2\rightarrow 16 \rightarrow 64 \rightarrow 128$ while the MLP in the waypoint decoder has the structure of $256$ (image feature)+$128$ (waypoint feature) $\rightarrow 1024 \rightarrow 512 \rightarrow 20 \ (10 \times 2)$.
We train the whole network for 50 epochs with an initial learning rate of 5e-4, a weight decay of 0.005, and a batch size of 128, where we introduce random feature selection rates in the last 20 epoches. It takes approximately 30 hours to converge with 6 NVIDIA-3090 GPUs.

\noindent
\textbf{Controller.} The controller obtains executable driving actions from the predicted waypoints, including steer, throttle and brake. We employ two PID controllers, a lateral controller, and a longitudinal controller, to generate the control signals. The lateral signal (turn signal) is calculated using the angle between the last two predicted waypoints, while the longitudinal signal (speed signal) is calculated using the average displacement in the predicted waypoints. Subsequently, we use the PID controller to generate a relatively smooth output. In mathematical terms, let $E \in \mathbb{R}^{N}$ be the historical signal with a time length of $N$, each PID controller takes the current signal $x$ as input and outputs 
$
x' = K_P * x + K_I * \mathrm{MEAN} (E) + K_D * (E[-1] - E[-2])
$
, where [$K_P$, $K_I$, $K_D$, N] forms a set of hyperparameters for a PID controller. Specifically, we set [1, 0.2, 0.1, 5] for the lateral controller and [5, 1, 0.1, 20] for the longitudinal controller. Note that there is no trainable parameter in the controller.


\begin{table*}[t]
\caption{\small Ablation study of CoDriving components on V2Xverse, evaluating robustness to pose errors ($\sigma_t/\sigma_r$) across driving, planning, and perception.}
\vspace{-2mm}
\resizebox{1.0\linewidth}{!}{
\begin{tabular}{ccc|cccc|cccc|cccc}
\toprule
\multirow{2}{*}{\textbf{\begin{tabular}[c]{@{}c@{}}Feature\\ aggregation\end{tabular}}} & \multirow{2}{*}{\textbf{\begin{tabular}[c]{@{}c@{}}Multi-scale\\ fusion\end{tabular}}} & \multirow{2}{*}{\textbf{\begin{tabular}[c]{@{}c@{}}Planning w\\ BEV feature\end{tabular}}} & \multicolumn{4}{c|}{Driving (driving score$\uparrow$)} & \multicolumn{4}{c|}{Waypoints planning (ADE$\downarrow$)} & \multicolumn{4}{c}{ 3D detection (mAP30$\uparrow$)} \\ \cline{4-15} 
 &  &  & 0.0/0.0 & 0.2/0.2 & 0.4/0.4 & 0.6/0.6 & 0.0/0.0 & 0.2/0.2 & 0.4/0.4 & 0.6/0.6 & 0.0/0.0 & 0.2/0.2 & 0.4/0.4 & 0.6/0.6 \\ \midrule
Max & $\checkmark$ & $\checkmark$ & 60.61 & 49.33 & 44.39 & 41.29 & 0.629 & 0.638 & 0.658 & 0.705 & 0.65 & 0.53 & 0.40 & 0.35 \\
Average & $\checkmark$ & $\checkmark$ & 40.53 & 38.25 & 32.35 & 30.12 & 0.634 & 0.685 & 0.783 & 0.855 & 0.61 & 0.46 & 0.31 & 0.23 \\
Attention &  & $\checkmark$ & 63.76 & 54.28 & 52.06 & 46.83 & 0.627 & 0.635 & 0.670 & 0.690 & 0.66 & 0.55 & 0.46 & 0.44 \\
Attention & $\checkmark$ &  & 61.98 & 55.02 & 50.56 & 46.01 & 0.652 & 0.661 & 0.679 & 0.685 & \textbf{0.68} & \textbf{0.57} & \textbf{0.49} & \textbf{0.47} \\
Attention & $\checkmark$ & $\checkmark$ & \textbf{77.15} & \textbf{58.13} & \textbf{57.61} & \textbf{50.63} & \textbf{0.619} & \textbf{0.629} & \textbf{0.642} & \textbf{0.649} & \textbf{0.68} & \textbf{0.57} & \textbf{0.49} & \textbf{0.47} \\ \midrule
\multicolumn{3}{c|}{No collaboration} & \multicolumn{4}{c|}{33.81} & \multicolumn{4}{c|}{0.636} & \multicolumn{4}{c}{0.57} \\ \bottomrule
\end{tabular}}
\label{table:PE_v2xverse}
\end{table*}

\begin{table*}[t]
\caption{ \small
Ablation study of CoDriving components on V2Xverse, evaluating robustness to communication latency (ms) across driving, planning, and perception.
}
\centering
\vspace{-2mm}
\resizebox{1.0\linewidth}{!}{
\begin{tabular}{ccc|cccc|cccc|cccc}
\toprule
\multirow{2}{*}{\textbf{\begin{tabular}[c]{@{}c@{}}Feature\\ aggregation\end{tabular}}} & \multirow{2}{*}{\textbf{\begin{tabular}[c]{@{}c@{}}Multi-scale\\ fusion\end{tabular}}} & \multirow{2}{*}{\textbf{\begin{tabular}[c]{@{}c@{}}Planning w\\ BEV feature\end{tabular}}} & \multicolumn{4}{c|}{Driving (driving score$\uparrow$)} & \multicolumn{4}{c|}{Waypoints planning (ADE$\downarrow$)} & \multicolumn{4}{c}{3D detection (mAP30$\uparrow$)} \\ \cline{4-15} 
 &  &  &  0ms  &  200ms  &  400ms  &  600ms  &   0ms   &  200ms  &  400ms  &  600ms  &   0ms   &  200ms  &  400ms  &  600ms  \\ \midrule
Max & $\checkmark$ & $\checkmark$ & 60.61 & 53.96 & 48.23 & 38.45 & 0.629 & 0.638 & 0.653 & 0.659 & 0.65 & 0.55 & 0.48 & 0.43 \\
Average & $\checkmark$ & $\checkmark$ & 40.53 & 39.06 & 40.08 & 33.67 & 0.634 & 0.674 & 0.735 & 0.763 & 0.61 & 0.46 & 0.36 & 0.32 \\
Attention &  & $\checkmark$ & 63.76 & 56.58 & 50.73 & 45.74 & 0.627 & 0.638 & 0.654 & 0.677 & 0.66 & 0.59 & 0.52 & 0.45 \\
Attention & $\checkmark$ &  & 61.98 & 55.48 & 48.09 & 44.79 & 0.652 & 0.657 & 0.660 & 0.675 & \textbf{0.68} & \textbf{0.60} & \textbf{0.55} & \textbf{0.49} \\
Attention & $\checkmark$ & $\checkmark$ & \textbf{77.15} & \textbf{64.23} & \textbf{57.73} & \textbf{51.44} & \textbf{0.619} & \textbf{0.621} & \textbf{0.627} & \textbf{0.636} & \textbf{0.68} & \textbf{0.60} & \textbf{0.55} & \textbf{0.49} \\ \midrule
\multicolumn{3}{c|}{No collaboration} & \multicolumn{4}{c|}{33.81} & \multicolumn{4}{c|}{0.636} & \multicolumn{4}{c}{0.57} \\ \bottomrule
\end{tabular}}
\label{table:latency_v2xverse}
\end{table*}

\begin{table*}[t]
\caption{ \small
Robustness to pose errors and communication latency on V2Xverse. We compare CoDriving to other collaboration methods.
}
\vspace{-2mm}
\centering
\resizebox{1.0\linewidth}{!}{
\begin{tabular}{c|ccc|ccc|ccc}
\toprule
\multirow{2}{*}{\textbf{Method}} & \multicolumn{3}{c|}{Driving (driving score$\uparrow$)} & \multicolumn{3}{c|}{Waypoints planning (ADE$\downarrow$)} & \multicolumn{3}{c}{3D detection (mAP30$\uparrow$)} \\ \cline{2-10} 
 & \begin{tabular}[c]{@{}c@{}}Ideal\\ situation\end{tabular} & \begin{tabular}[c]{@{}c@{}}Pose noise\\ 0.6/0.6($\sigma_t/\sigma_r$)\end{tabular} & \begin{tabular}[c]{@{}c@{}}Latency\\ 600ms\end{tabular} & \begin{tabular}[c]{@{}c@{}}Ideal\\ situation\end{tabular} & \begin{tabular}[c]{@{}c@{}}Pose noise\\ 0.6/0.6($\sigma_t/\sigma_r$)\end{tabular} & \begin{tabular}[c]{@{}c@{}}Latency\\ 600ms\end{tabular} & \begin{tabular}[c]{@{}c@{}}Ideal\\ situation\end{tabular} & \begin{tabular}[c]{@{}c@{}}Pose noise\\ 0.6/0.6($\sigma_t/\sigma_r$)\end{tabular} & \begin{tabular}[c]{@{}c@{}}Latency\\ 600ms\end{tabular} \\ \midrule
Late Fusion & 52.40\textcolor{gray}{($\downarrow0\%$)} & 32.49\textcolor{gray}{($\downarrow38\%$)} & 33.41\textcolor{gray}{($\downarrow36\%$)} & 0.631\textcolor{gray}{($\uparrow0\%$)} & 0.669\textcolor{gray}{($\uparrow6\%$)} & 0.650\textcolor{gray}{($\uparrow3\%$)} & 0.59\textcolor{gray}{($\downarrow0\%$)} & 0.30\textcolor{gray}{($\downarrow48\%$)} & 0.40\textcolor{gray}{($\downarrow31\%$)} \\
Fcooper~\cite{ChenFcooper:SEC19} & 44.00\textcolor{gray}{($\downarrow0\%$)} & 27.74\textcolor{gray}{($\downarrow37\%$)} & 25.80\textcolor{gray}{($\downarrow40\%$)} & 0.627\textcolor{gray}{($\uparrow0\%$)} & 0.671\textcolor{gray}{($\uparrow7\%$)} & 0.663\textcolor{gray}{($\uparrow6\%$)} & 0.64\textcolor{gray}{($\downarrow0\%$)} & 0.36\textcolor{gray}{($\downarrow44\%$)} & 0.41\textcolor{gray}{($\downarrow36\%$)} \\
V2X-ViT~\cite{xu2022v2xvit} & 39.35\textcolor{gray}{($\downarrow0\%$)} & 30.71\textcolor{gray}{($\downarrow22\%$)} & 36.33\textcolor{gray}{($\downarrow8\%$)} & 0.629\textcolor{gray}{($\uparrow0\%$)} & 0.664\textcolor{gray}{($\uparrow6\%$)} & 0.649\textcolor{gray}{($\uparrow3\%$)} & 0.61\textcolor{gray}{($\downarrow0\%$)} & 0.41\textcolor{gray}{($\downarrow33\%$)} & 0.43\textcolor{gray}{($\downarrow30\%$)} \\
CoDriving & \textbf{77.15}\textcolor{gray}{($\downarrow0\%$)} & \textbf{50.63}\textcolor{gray}{($\downarrow35\%$)} & \textbf{51.44}\textcolor{gray}{($\downarrow33\%$)} & \textbf{0.619}\textcolor{gray}{($\uparrow0\%$)} & \textbf{0.649}\textcolor{gray}{($\uparrow5\%$)} & \textbf{0.636}\textcolor{gray}{($\uparrow3\%$)} & \textbf{0.68}\textcolor{gray}{($\downarrow0\%$)} & \textbf{0.47}\textcolor{gray}{($\downarrow31\%$)} & \textbf{0.49}\textcolor{gray}{($\downarrow28\%$)} \\ \bottomrule
\end{tabular}}
\label{table:robust_v2xverse}
\end{table*}

\vspace{-3mm}
\begin{table}[t]
\caption{ \small
Ablation study of CoDriving components on OPV2V and DAIR-V2X, evaluating robustness to pose errors($\sigma_t/\sigma_r$) in perception task.
}
\centering
\resizebox{0.95\linewidth}{!}{
\begin{tabular}{cc|cccc|cccc}
\toprule
\multirow{2}{*}{\textbf{\begin{tabular}[c]{@{}c@{}}Feature\\ aggregation\end{tabular}}} & \multirow{2}{*}{\textbf{\begin{tabular}[c]{@{}c@{}}Multi-scale\\ fusion\end{tabular}}} & \multicolumn{4}{c|}{OPV2V (AP30$\uparrow$)} & \multicolumn{4}{c}{DAIR-V2X (AP50$\uparrow$)} \\ \cline{3-10} 
 &  & 0.0/0.0 & 0.2/0.2 & 0.4/0.4 & 0.6/0.6 & 0.0/0.0 & 0.2/0.2 & 0.4/0.4 & 0.6/0.6 \\ \midrule
Max & $\checkmark$ & 0.94 & 0.92 & 0.84 & 0.74 & 0.71 & 0.70 & 0.67 & 0.65 \\
Average & $\checkmark$ & 0.89 & 0.87 & 0.80 & 0.69 & 0.70 & 0.69 & 0.67 & \textbf{0.66} \\
Attention &  & 0.96 & 0.95 & 0.92 & 0.89 & 0.71 & 0.69 & 0.67 & 0.65 \\
Attention & $\checkmark$ & \textbf{0.97} & \textbf{0.97} & \textbf{0.94} & \textbf{0.91} & \textbf{0.73} & \textbf{0.71} & \textbf{0.68} & \textbf{0.66} \\ \midrule
\multicolumn{2}{c|}{No collaboration} & \multicolumn{4}{c|}{0.83} & \multicolumn{4}{c}{0.66} \\ \bottomrule
\end{tabular}}
\label{table:PE_opv2v}
\end{table}

\begin{table}[t]
\caption{ \small
Ablation study of CoDriving components on OPV2V and DAIR-V2X, evaluating robustness to communication latency (ms) in perception task.
}
\centering
\resizebox{0.95\linewidth}{!}{
\begin{tabular}{cc|cccc|cccc}
\toprule
\multirow{2}{*}{\textbf{\begin{tabular}[c]{@{}c@{}}Feature\\ aggregation\end{tabular}}} & \multirow{2}{*}{\textbf{\begin{tabular}[c]{@{}c@{}}Multi-scale\\ fusion\end{tabular}}} & \multicolumn{4}{c|}{OPV2V (AP30$\uparrow$)} & \multicolumn{4}{c}{DAIR-V2X (AP50$\uparrow$)} \\ \cline{3-10} 
 &  & 0ms & 200ms & 400ms & 600ms & 0ms & 200ms & 400ms & 600ms \\ \midrule
Max & $\checkmark$ & 0.94 & 0.74 & 0.46 & 0.41 & 0.71 & 0.68 & 0.67 & 0.66 \\
Average & $\checkmark$ & 0.89 & 0.61 & 0.24 & 0.18 & 0.70 & 0.68 & \textbf{0.68} & \textbf{0.67} \\
Attention &  & 0.96 & 0.90 & \textbf{0.80} & 0.76 & 0.71 & 0.66 & 0.66 & 0.65 \\
Attention & $\checkmark$ & \textbf{0.97} & \textbf{0.92} & \textbf{0.80} & \textbf{0.78} & \textbf{0.73} & \textbf{0.69} & \textbf{0.68} & \textbf{0.67} \\ \midrule
\multicolumn{2}{c|}{No collaboration} & \multicolumn{4}{c|}{0.83} & \multicolumn{4}{c}{0.66} \\ \bottomrule
\end{tabular}}
\label{table:latency_opv2v}
\end{table}

\section{\textcolor{blue}{Discussion of methodology designs}}
\subsection{\textcolor{blue}{Robustness to latency and pose errors}}
\label{subsec:robustness_latency}
In this section, we present further ablation studies to investigate the impact of CoDriving components on robustness against pose errors and communication latency.
We focus on the effects of three modules in CoDriving: 1) The feature aggregation function, which transfers the stack of features from neighboring agents $\left[\mathcal{F}_{(j,l)}^{(t)} \right]_{j\in\mathcal{N}_i}$ into the fused feature $\mathcal{F}_{\mathcal{N}_i}$, as described in Section~\ref{subsec:collaboration} of the manuscript. We compare this functionality with two common operations in V2X BEV feature fusion. The first is max fusion~\cite{ChenFcooper:SEC19,zimmer2024tumtraf}, which combines neighboring features by retaining the maximum element from different agents at each pixel. The second is average fusion, which merges features by directly averaging values from all agents, serving as a naive version of attention fusion by simplifying the addition weights to $\frac{1}{\mathcal{N}_i}$. 2) The multi-scale feature fusion strategy, which enhances features across three resolution levels, $\{ \mathcal{F}_{(i,l)}^{(t)} \}_{l=0}^2$, as outlined in Section~\ref{subsec:end2end},~\ref{subsec:collaboration}. We compare this approach to single-scale enhancement, which only enhances feature $\mathcal{F}_{i}^{(t)}$ at one resolution level, as used in previous works~\cite{ChenFcooper:SEC19,WangV2vnet:ECCV20}. 3) The planner with BEV feature input. In CoDriving, structured BEV occupancy maps are used together with detailed BEV features $\{\mathbf{B}_i^{(k)},\mathcal{F}_i^{(t)}\}_{k=t-T_{h}}^{t}$ as inputs to the planning module. We compare this with a paradigm that uses only occupancy maps as planning input, which works with late fusion strategy and non-end-to-end driving pipelines.

\noindent
\textbf{Ablation studies on CoDriving components.}
We evaluate the impact of different module designs on system robustness across varying levels of pose errors and communication latency. This evaluation is performed on our V2Xverse, OPV2V~\cite{XuOPV2V:ICRA22}, and DAIR-V2X~\cite{YuDAIRV2X:CVPR22}, following the experimental settings described in Sections~\ref{sec:Benchmark_perception},~\ref{sec:Benchmark_driving} to incorporate pose errors and latency. Table~\ref{table:PE_v2xverse},~\ref{table:latency_v2xverse} present the driving, waypoint planning, and 3D detection performance on V2Xverse, while Table~\ref{table:PE_opv2v},~\ref{table:latency_opv2v} display the 3D detection performance on OPV2V and DAIR-V2X. Here, we discuss the effects of latency and pose errors together, as both act similarly as disturbances to object positions. The results show that:
1) Compared to max fusion and average fusion (Line 1, Line 2 in Table~\ref{table:PE_v2xverse},~\ref{table:latency_v2xverse},~\ref{table:PE_opv2v},~\ref{table:latency_opv2v}), attention fusion (Line 5 in Table~\ref{table:PE_v2xverse},~\ref{table:latency_v2xverse}, Line 4 in~\ref{table:PE_opv2v},~\ref{table:latency_opv2v}) demonstrates superior robustness across all pose error and latency disturbance conditions. The reason is that attention fusion leverages the ego vehicle’s feature map as reference, weighting feature values from other agents based on vector similarity before aggregation, which helps to filter out noisy feature values.
2) Compared to single-scale feature fusion (Line 3 in Table~\ref{table:PE_v2xverse},~\ref{table:latency_v2xverse},~\ref{table:PE_opv2v},~\ref{table:latency_opv2v}), multi-scale fusion (Line 5 in Table~\ref{table:PE_v2xverse},~\ref{table:latency_v2xverse}, Line 4 in~\ref{table:PE_opv2v},~\ref{table:latency_opv2v}) delivers superior performance by incorporating lower-resolution features, which provide more global scene information and are less sensitive to positional disturbances.
3) Compared to planning with occupancy input alone (Line 4 in Table~\ref{table:PE_v2xverse},~\ref{table:latency_v2xverse}), planning with the additional input of BEV features (Line 5 in Table~\ref{table:PE_v2xverse},~\ref{table:latency_v2xverse}) achieves superior performance in driving and planning tasks across varying levels of disturbance. The reason is that, the additional bev feature provide a more comprehensive understanding of the scenario, benefiting the overall planning capability of the driving system. We see that this hybrid planning input outperforms occupancy-only input, achieving a $24\%$ improvement in driving score under ideal conditions without disturbances. Although both manners are affected by disturbances, the system with hybrid planning input maintains relatively higher performance.
Note that the perception module is frozen during the training of the planning module; therefore, the detection performance remains consistent despite variations in the planning input.

\noindent
\textbf{Comparison with other collaboration methods.}
Table~\ref{table:robust_v2xverse} compares CoDriving with the three collaboration methods discussed in Section~\ref{sec:Benchmark_driving} of the manuscript, showing performance under ideal conditions (no noise or latency), with pose noise of 0.6/0.6 ($\sigma_t/\sigma_r$), and with a latency of 600ms. We see that: 1) Compared to late fusion, which directly gathers the detection results using a non-max suppression (NMS) algorithm and fully reflects the impact of disturbances on detection/planning/driving performance, CoDriving demonstrates stronger robustness. 
The reason is that, CoDriving employs an intermediate fusion strategy, allowing misplaced feature values to be further perceived by feature regions at the correct positions through the following convolutional layers.
2) Compared with Fcooper~\cite{ChenFcooper:SEC19}, which employs max fusion during the feature fusion stage, CoDriving achieves greater robustness by leveraging attention fusion and multi-scale fusion strategies. For further details, see the ablation study and discussion in Table~\ref{table:PE_v2xverse},~\ref{table:latency_v2xverse},~\ref{table:PE_opv2v},~\ref{table:latency_opv2v}.
3) Compared to V2X-ViT~\cite{xu2022v2xvit}, CoDriving demonstrates superior performance in both ideal and disturbed conditions. Notably, V2X-ViT achieves a lower driving score (39.35) than CoDriving (77.15) in ideal situation. One contributing factor is the challenge of detecting pedestrians in V2Xverse scenarios. V2X-ViT struggles with pedestrian detection, achieving only 0.41 AP30, compared to CoDriving’s 0.58 AP30, which is reflected in the pedestrian collision rates—2.61 for V2X-ViT versus 0.73 for CoDriving. This discrepancy ultimately impacts the overall driving score. Under disturbance conditions, the driving/planning/perception performance of V2X-ViT further declines to relatively low levels, highlighting CoDriving’s robustness and superiority.

\begin{table*}[t]
\caption{\small
Performance at various communication Rates. The proposed driving-request mechanism significantly enhances driving and planning performance.
}
\centering
\resizebox{1.0\linewidth}{!}{
\begin{tabular}{cc|ccccc|ccccc|ccccc}
\toprule
\multirow{2}{*}{\begin{tabular}[c]{@{}c@{}}Confidence\\ map\end{tabular}} & \multirow{2}{*}{\begin{tabular}[c]{@{}c@{}}Driving\\ request map\end{tabular}} & \multicolumn{5}{c|}{\begin{tabular}[c]{@{}c@{}}Driving (driving score$\uparrow$)\end{tabular}} & \multicolumn{5}{c|}{\begin{tabular}[c]{@{}c@{}}Waypoints Planning (ADE$\downarrow$)\end{tabular}} & \multicolumn{5}{c}{\begin{tabular}[c]{@{}c@{}}3D Detection (mAP30$\uparrow$)\end{tabular}} \\ 
 &  & $2^0$ & $2^3$ & $2^6$ & $2^9$ & $2^{12}$ & $2^0$ & $2^3$ & $2^6$ & $2^9$ & $2^{12}$ & $2^0$ & $2^3$ & $2^6$ & $2^9$ & $2^{12}$ \\ \midrule[0.5pt]
$\checkmark$ &  & \textbf{77.15} & 66.47 & 61.57 & 59.08 & 53.29 & \textbf{0.617} & 0.623 & 0.627 & 0.628 & 0.630 & \textbf{0.68} & \textbf{0.68} & \textbf{0.66} & \textbf{0.64} & \textbf{0.60} \\
 & $\checkmark$ & \textbf{77.15} & 64.23 & 55.31 & 41.90 & 35.86 & \textbf{0.617} & 0.623 & 0.630 & 0.632 & 0.636 & \textbf{0.68} & 0.61 & 0.59 & 0.57 & 0.57 \\
$\checkmark$ & $\checkmark$ & \textbf{77.15} & \textbf{73.31} & \textbf{67.08} & \textbf{66.19} & \textbf{58.45} & \textbf{0.617} & \textbf{0.618} & \textbf{0.622} & \textbf{0.624} & \textbf{0.627} & \textbf{0.68} & 0.66 & 0.64 & \textbf{0.64} & \textbf{0.60} \\ \bottomrule
\end{tabular}
}

\label{table:ablation_cmap_dmap}
\end{table*}

\begin{table*}[t]
\renewcommand{\arraystretch}{1.05}
\centering
\caption{\small Comparison on different compression strategies in driving, planning and perception. The compression rate ranges from $2^0$ (original feature map) to $2^{12}$.}
\resizebox{1.0\linewidth}{!}{
\begin{tabular}{lr|ccccc|ccccc|ccccc}
\toprule
\multicolumn{2}{c|}{\multirow{2}{*}{Compression Strategy}} & \multicolumn{5}{c|}{\begin{tabular}[c]{@{}c@{}}Driving (driving score$\uparrow$)\end{tabular}} & \multicolumn{5}{c|}{\begin{tabular}[c]{@{}c@{}}Waypoints Planning (ADE$\downarrow$)\end{tabular}} & \multicolumn{5}{c}{\begin{tabular}[c]{@{}c@{}}3D Detection (mAP30$\uparrow$)\end{tabular}} \\ 
\multicolumn{2}{c|}{} & $2^0$ & $2^3$ & $2^6$ & $2^9$ & $2^{12}$ & $2^0$ & $2^3$ & $2^6$ & $2^9$ & $2^{12}$ & $2^0$ & $2^3$ & $2^6$ & $2^9$ & $2^{12}$ \\ \midrule
\multicolumn{1}{l|}{\multirow{6}{*}{\begin{tabular}[c]{@{}l@{}}Spatial Sampling \\ with Confidence Map\end{tabular}}} & random-request & 73.73 & 63.98 & 45.21 & 40.74 & 35.59 & 0.623 & 0.626 & 0.630 & 0.631 & 0.635 & \textbf{0.68} & 0.65 & 0.64 & 0.62 & 0.58 \\
\multicolumn{1}{l|}{} & uncertainty-request & 72.36 & 64.22 & 52.43 & 47.62 & 40.24 & 0.624 & 0.624 & 0.626 & 0.628 & 0.632 & \textbf{0.68} & 0.66 & 0.65 & 0.63 & \textbf{0.60} \\
\multicolumn{1}{l|}{} & nearego-request & 74.17 & 69.83 & 56.52 & 50.88 & 42.37 & 0.620 & 0.622 & 0.623 & 0.627 & 0.630 & \textbf{0.68} & 0.62 & 0.61 & 0.60 & 0.58 \\
\multicolumn{1}{l|}{} & low-confidence-request & 75.43 & 66.32 & 60.34 & 59.36 & 51.54 & 0.620 & 0.622 & 0.627 & 0.628 & 0.631 & \textbf{0.68} & \textbf{0.67} & \textbf{0.66} & \textbf{0.64} & \textbf{0.60} \\
\multicolumn{1}{l|}{} & driving-request (ours) & \textbf{77.15} & \textbf{73.31} & \textbf{67.08} & \textbf{66.19} & \textbf{58.45} & \textbf{0.617} & \textbf{0.618} & \textbf{0.622} & \textbf{0.624} & \textbf{0.627} & \textbf{0.68} & 0.66 & 0.65 & \textbf{0.64} & \textbf{0.60} \\ \cline{1-2}
\multicolumn{1}{l|}{\multirow{2}{*}{\begin{tabular}[c]{@{}l@{}}VAE-based \\ Entropy Coding\end{tabular}}} & \multicolumn{1}{r|}{FactorizedHyperprior~\cite{balle2018variational}} & - & - & - & - & 34.05 & - & - & - & - & 0.635 & - & - & - & - & 0.57 \\
\multicolumn{1}{l|}{} & \multicolumn{1}{r|}{ScaledHyperprior~\cite{balle2018variational}} & - & - & - & - & 35.86 & - & - & - & - & 0.635 & - & - & - & - & 0.58 \\ \midrule
\multicolumn{2}{c|}{None Collaboration} & \multicolumn{5}{c|}{33.81} & \multicolumn{5}{c|}{0.636} & \multicolumn{5}{c}{0.57} \\ \bottomrule
\end{tabular}}
\label{table:other_sampling}
\end{table*}

\subsection{\textcolor{blue}{Communication efficiency}}
This section presents additional ablation studies and discussions on the significance of CoDriving in achieving efficient communication.

\noindent
\textbf{Ablation studies on confidence map and driving request.}
We have provided a comparison between (driving-request+confidence map) and (confidence map) in Figure~\ref{Fig:comm-bandwidth}, here we provide further discussion and ablation studies.
In Table~\ref{table:ablation_cmap_dmap}, we examine the individual effects of the confidence map and the driving-request map across a range of compression rates, from $2^0$ to $2^{12}$, where a compression rate of $2^0$ indicates no compression is applied. We see that: 
1) The proposed driving-request mechanism brings significant improvements to the ultimate closed-loop driving performance. Compared to communication without driving-request (Line 1), the driving-request mechanism (Line 3) significantly improves driving score by ($8.5\%,$ $8.9\%,$ $11.1\%,$ $9.7\%$) under compression rates of ($2^3,2^6,2^9,2^{12}$).
The reason for this improvement is that, the regions near the planned waypoints contain most of the safety-critical objects, which could have a significant impact on the ego vehicle’s planning. As communication volume decreases, the driving-request map assigns higher priority to regions near the planned waypoints during feature transmission. Consequently, the features discarded during compression are more likely to contain object information with minimal impact on vehicle planning. This leads to a smaller decline in driving performance compared to the confidence-map-only approach, which treats feature information within and outside the planning region equally. Note that the driving-request mechanism slightly reduces perception performance. This effect occurs because the driving-request mechanism prioritizes objects in regions near predicted waypoints, thereby strategically abandoning perceptual information in non-safety-critical regions. As a result, detection precision for objects in these areas, which constitute a significant portion, is somewhat compromised. Despite this, the driving-request mechanism achieves superior performance in driving and planning, as not every detected object is critical for planning, and the driving-request mechanism effectively reduces communication volume by selectively ignoring non-critical perceptual features.
2) Compared to using only the driving-request map (Line 2), the combination of the driving-request map and confidence map (Line 3) significantly improves detection, planning, and driving performance. This improvement is due to the CoDriving system’s reliance on accurate detection of foreground objects for effective driving. The confidence map serves as an essential reference for identifying critical foreground object information while filtering out unnecessary background details.

In conclusion, 
the driving-request map enhances the effectiveness of collaborative feature transmission from a planning perspective, while the confidence map improves it from a perception perspective. CoDriving leverages both approaches to precisely select driving-critical foreground features.

\noindent
\textbf{Compare driving-request with other intuitive spatially request mechanism.} We compare the driving-request mechanism with several other request methods, including 1) Random-request, where 64 points are randomly selected within the detection range to replace the predicted waypoints and generate the request map in the same way; 
2) Uncertainty-request, where the request map is defined by the entropy of detection confidence, highlighting regions where detection confidence is around 0.5, indicating that the detector has difficulty distinguishing foreground from background;
3) Near-ego-request, which emphasizes regions near the ego vehicle in the request map, using a Gaussian distribution to prioritize areas close to the ego vehicle;
and 4) Low-confidence-request, where the request map is defined as $1-\mathbf{C}_j^{(t)}$ to target regions with low foreground detection confidence, in other words, highlighting background regions.

For a fair comparison, we apply confidence map in feature sampling for each manner, and apply a training strategy that introduces random communication rates during the training of each manner’s planning module, as done in CoDriving; see detailed training strategy in Section~\ref{subsec:losses}.
Table~\ref{table:other_sampling} shows the driving/planning/detection performance across a range of compression rates, from $2^0$ to $2^{12}$, where a compression rate of $2^0$ indicates no compression is applied.
We see that: 1) the driving-request strategy outperforms other approaches in the driving task, achieving improvements of ($5\%,11\%,11\%,13\%$) under compression rates of ($2^3,2^6,2^9,2^{12}$); 2) The planner trained with the driving-request strategy outperforms other approaches in driving performance by $2\%$ under conditions without compression. Here we provide qualitative analyses to explain the observed results. The random-request method introduces noise rather than task-oriented information, and therefore fails to improve the efficiency of information filtering. The uncertainty-request map mainly highlights the edges of object centers (the boundary between foreground and background), where detector has difficulty distinguishing foreground from
background. These regions largely overlap with the regions highlighted by the confidence map and rarely provide additional information. Additionally, if an object is invisible in the sensor’s view, the detector typically classifies it as background with low uncertainty and low foreground detection confidence, meaning it is unlikely to be highlighted as high-uncertainty in the request map. While the low-confidence-request approach does highlight regions detected as containing no objects, these regions are not sparse enough to significantly narrow the sampling range. Consequently, this approach provides minimal benefit in enhancing the efficiency of information filtering. In contrast, the driving-request strategy effectively selects safety-critical regions. Notably, driving/planning performance with driving-request surpasses other strategies even at a compression rate of $2^0$. This suggests that the driving-request mechanism contributes to a more effective imitation fitting by prioritizing safety-critical regions within the collaborative features during training.

\noindent
\textbf{Compare driving-request with VAE entropy coding compression.} We compare driving-request with two compression methods introduced in ~\cite{balle2018variational}: FactorizedHyperprior and ScaledHyperprior. For these two methods, we define the distortion objective as the reconstruction loss (between the decompressed and uncompressed feature maps). While the compression rate of these two methods varies with the test samples, we approximate the desired compression rate by adjusting the resolution of the compressed features.
In Table~\ref{table:other_sampling}, we see that: 1) Under a compression rate of $2^{12}$, driving-request achieves a superior performance by $81\%$ in driving, $2\%$ in planning, and $5\%$ in detection. The reason is that driving-request selectively transmits the object features in safety-critical regions, while the VAE-based methods fails to explicitly reconstruct the feature map. 2) CoDriving shows advantages in freely adjusting compression rate during inference time, while the VAE-based compression methods are restricted by fixed compression rate, and need to train different models to adapt different bandwidth budgets. The reason is that, CoDriving could adapt to different communication rates by solving the optimization problem Equation~\ref{sec4:problem_formulation}, while the compression rate of VAE-based methods is largely determined by the resolution of the coded feature.

\subsection{\textcolor{blue}{Design rationale of driving-request}}
\label{subsec:rationale}
In this section, we provide a further explanation of why regions near waypoints can be considered safety-critical, which serves as the design motivation for the proposed driving-request mechanism.

\noindent
\textbf{Intuitive analysis.} In driving, the areas near the planned waypoints could be considered more "safety-critical" for several reasons.
1) The areas near the planned waypoints encompass the safety-critical zones around the ego vehicle. Our planner predicts 10 waypoints to be reached within the next 2 seconds, with the area near the first waypoint being particularly close to the ego vehicle. Any changes in this region should be given priority. For example, the ego vehicle must be aware of any pedestrians moving nearby to avoid a collision. Additionally, when the ego vehicle is stopped behind another vehicle at a red light, the driver typically decides when to move forward based on the motion of the vehicle ahead.
2) The areas near the planned waypoints have a higher safety priority compared to regions outside the planned path. It is natural to focus more on obstacles along the planned route rather than on those along other routes, just as human drivers tend to look in the direction aligned with the steering wheel. For example, if the ego vehicle is moving forward, pedestrians ahead pose a greater hazard than those behind. Similarly, if the ego vehicle is turning right at an intersection, obstacles on the right-hand path demand more attention than those on the road ahead.
3) The areas near the planned waypoints are where the ego vehicle is headed, meaning that traffic objects in these regions typically have a higher relative speed to the ego vehicle. This makes these objects more hazardous and these areas more critical for safety.

\noindent
\textbf{Statistical analysis.}
The priority of regions near the planned waypoints can also be validated from a statistical perspective. We use a powerful rule-based expert agent~\cite{InterFuser} to assess which objects pose a hazard to the ego vehicle, and apply a distance threshold $\delta_w$ to determine whether an object is near-waypoints. Note that the expert agent incorporates several rule-based functions to accurately identify hazardous objects by calculating their potential to collide with the ego vehicle. Table~\ref{table:count_actors_2} presents the distribution of objects within the detection range used in CoDriving. This statistical analysis is based on our training set across CARLA town1-town4, encompassing vehicles, cyclists, and pedestrians. We see that \textbf{the near-waypoint regions filter out a small portion of objects while retaining most of the hazardous ones}: 1) Most hazardous objects are located near waypoints under various thresholds $\delta_w$, demonstrating that the regions around waypoints are indeed safety-critical. For example, at a threshold $\delta_w$ of 5 meters, these regions encompass $90\%$ of hazardous objects. 2) As the threshold $\delta_w$ decreases to 2 meters, the near-waypoint area includes only $8\%$ of all objects, yet this subset still accounts for $56\%$ of hazardous objects, indicating that near-waypoint regions effectively capture high-risk objects. These analyses emphasize the critical role of near-waypoint regions in identifying and prioritizing hazardous objects for safer navigation.

\begin{table}[H]
\centering
\caption{\small Object distribution. The regions near the waypoints encompass the majority of hazardous objects.}
\resizebox{1.0\linewidth}{!}{
\begin{tabular}{c|cccc}
\toprule
\multirow{2}{*}{Proportion of Objects} & \multicolumn{4}{c}{threshold $\delta_w$} \\
 & 5m & 4m & 3m & 2m \\ \midrule
$\frac{\text{objects near waypoints}}{\text{total objects}}$ & $33\%$ & $27\%$ & $13\%$ & $8\%$ \\
$\frac{\text{hazard objects near waypoints}}{\text{total hazard objects}}$ & $90\%$ & $81\%$ & $69\%$ & $56\%$ \\ \bottomrule
\end{tabular}}
\label{table:count_actors_2}
\end{table}

\subsection{\textcolor{blue}{Effectiveness of training strategy}}
\label{subsec:training_strategy}
We further validate the effectiveness of the training strategy introduced in Section~\ref{subsec:losses}. Table~\ref{table:training_strategy} compares the performance of CoDriving with and without the training strategy that includes random feature selection rates. In the absence of random feature selection, the communication rate is consistently set to 1. The results indicate that: 1) our training strategy enhances CoDriving’s performance across various communication rates by adapting the model to diverse communication conditions through collaborative feature augmentation; and 2) this strategy improves the planner under both full-communication and limited-communication conditions, suggesting that the driving-request mechanism contributes to more effective imitation learning by prioritizing safety-critical regions within the collaborative features during training.

\begin{table*}[t]
\centering
\renewcommand{\arraystretch}{1.05}
\caption{\small Driving and planning performance on V2Xverse with compression rates ranging from $2^0$ to $2^6$. The training strategy of CoDriving, incorporating random feature selection rates, improves the balance between performance and communication volume.}
\resizebox{1.0\linewidth}{!}{
\begin{tabular}{ccccccc}
\toprule
\multicolumn{7}{c}{\textbf{CoDriving performance}} \\ \midrule
\multicolumn{1}{c|}{\multirow{2}{*}{\begin{tabular}[c]{@{}c@{}}Random communication \\ rates in Training\end{tabular}}} & \multicolumn{3}{c|}{Driving (driving score$\uparrow$)} & \multicolumn{3}{c}{Waypoints Planning (ADE$\downarrow$)} \\ 
\multicolumn{1}{c|}{} & $2^0$ & $2^3$ & \multicolumn{1}{c|}{$2^6$} & $2^0$ & $2^3$ & $2^6$ \\ \midrule
\multicolumn{1}{c|}{$\times$} & 73.67 & 68.45 & \multicolumn{1}{c|}{62.64} & 0.624 & 0.628 & 0.630 \\
\multicolumn{1}{c|}{$\checkmark$} & 77.15\textcolor{gray}{($\uparrow5\%$)} & 73.31\textcolor{gray}{($\uparrow7\%$)} & \multicolumn{1}{c|}{67.08\textcolor{gray}{($\uparrow7\%$)}} & 0.617\textcolor{gray}{($\downarrow1.1\%$)} & 0.618\textcolor{gray}{($\downarrow1.6\%$)} & 0.622\textcolor{gray}{($\downarrow1.2\%$)} \\ \bottomrule
\end{tabular}
}
\label{table:training_strategy}
\end{table*}

\vspace{3mm}
\section{Visualization}

\begin{figure}[t]
\centering
\includegraphics[width=0.90\linewidth]{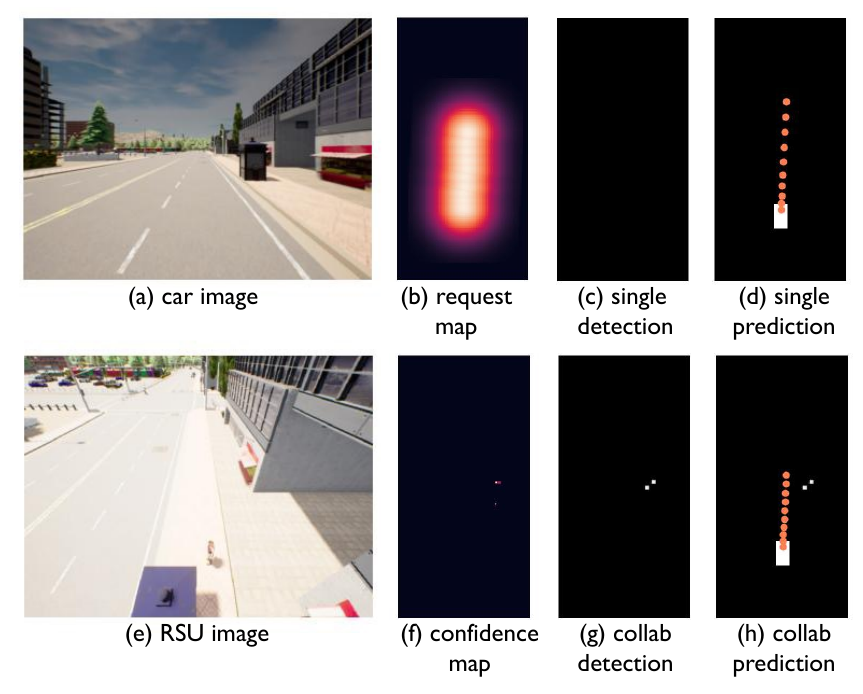}
\vspace{-2mm}
\caption{\small Visualization of collaboration between ego vehicle and RSU, including spatial confidence map, driving request map, and the detection and planning before and after collaboration. In this scenario, a pedestrian is occluded by a telephone booth, and this occluded object can be detected through V2X-communication-enabled information sharing. Compared to single-agent planning, collaborative waypoints planning generates closer waypoints, indicating a slower driving speed.}
\label{fig:Collaboration}
\vspace{-4mm}
\end{figure}
\begin{figure}[t]
\centering
\includegraphics[width=0.90\linewidth]{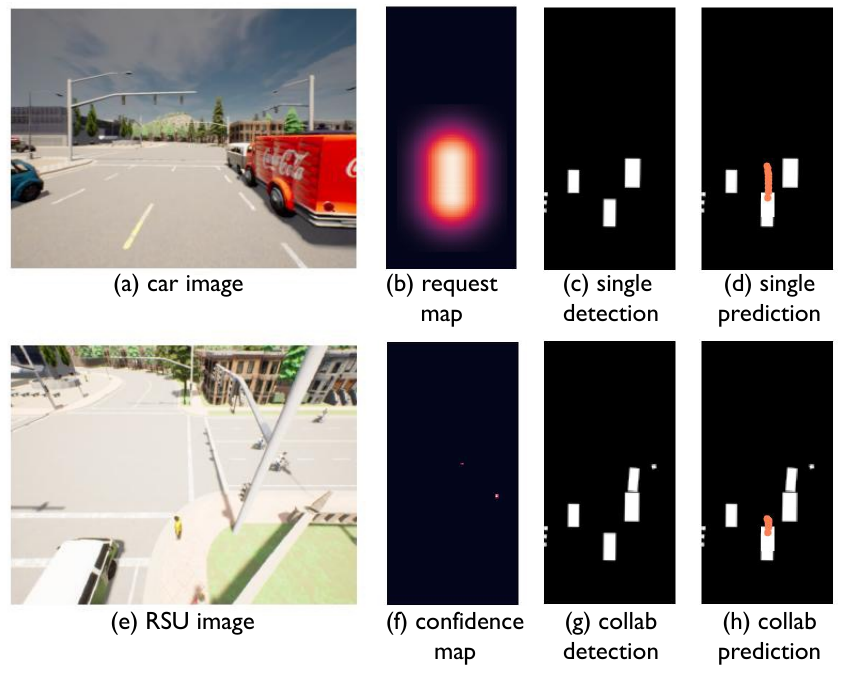}
\vspace{-2mm}
\caption{\small Another visualization of collaboration between ego vehicle and RSU. In this scenario, a pedestrian is occluded by two cars.
Collaborative waypoints planning produces a safer and more cautious driving speed.
}
\label{fig:Collaboration_example2}
\end{figure}

Figure~\ref{fig:Collaboration} and Figure~\ref{fig:Collaboration_example2} provide two examples that illustrate how CoDriving benefits from the proposed driving-request aware collaboration. In these scenarios, a pedestrian is missed in vehicle's view. However, with assistance from the RSU, the ego vehicle is able to detect the missed objects and avoid the catastrophic collision caused by the miss-detection. Figure~\ref{fig:Collaboration}/\ref{fig:Collaboration_example2}(a-d) shows ego vehicle's sensor view, driving-request map, detection results, and predicted waypoints based on its own observation. Figure~\ref{fig:Collaboration}/\ref{fig:Collaboration_example2}(e-f) shows RSU's sensor view, confidence map, and ego vehicle's V2X-communication-aided detection results, and predicted waypoints. Figure~\ref{fig:Collaboration}/\ref{fig:Collaboration_example2}(f) shows the spatial perceptual confidence map, which is sparse, yet highlights the positions of objects. 
Figure~\ref{fig:Collaboration}/\ref{fig:Collaboration_example2}(b) shows the driving request map, which is concentrated on driving-critical areas. 
Figure~\ref{fig:Collaboration}/\ref{fig:Collaboration_example2}(c)/(d) and (g)/(h) compare the detection and planning results before and after the collaboration with RSU. We see that the proposed request map and spatial confidence map contribute to spatially sparse, yet driving-critical message, which effectively helps the ego vehicle to detect occluded objects and achieve safer driving.

\end{document}